\documentclass[final,5p,times,twocolumn,nopreprintline]{elsarticle}

\usepackage{amsmath,amsfonts}
\usepackage{algorithmic}
\usepackage{algorithm}
\usepackage{array}
\usepackage{textcomp}
\usepackage{stfloats}
\usepackage{url}
\usepackage{verbatim}
\usepackage{subcaption}
\usepackage{booktabs} 
\usepackage{multirow}

\usepackage{amssymb}
\usepackage{mathtools}
\usepackage{amsthm}

\usepackage[dvipsnames]{xcolor}

\theoremstyle{plain}
\newtheorem{theorem}{Theorem}[section]

\newtheorem{lemma}[theorem]{Lemma}
\newtheorem{corollary}[theorem]{Corollary}
\theoremstyle{definition}

\newtheorem{assumption}[theorem]{Assumption}
\theoremstyle{remark}

\newcommand{\norm}[1]{\left\lVert#1\right\rVert}

\newcommand{\argmin}[2]{arg\;\underset{#1}{\min}\;#2}

\newcommand{\mycomment}[1]{}

\AtBeginDocument{\mathcode`v=\varv}

\journal{Neurocomputing}

\begin{document}

\begin{frontmatter}


\author{Wendyam Eric Lionel Ilboudo\fnref{label1}}
\ead{ilboudo.wendyam\_eric.in1@is.naist.jp}
\author{Taisuke Kobayashi\fnref{label2}}
\ead{kobayashi@nii.ac.jp}
\author{Takamitsu Matsubara\fnref{label1}}
\ead{takam-m@is.naist.jp}
\affiliation[label1]{organization={Nara Institute of Science and Technology},
           city={Nara},
           country={Japan}}
\affiliation[label2]{organization={National Institute of Informatics/The Graduate University for Advanced Studies, SOKENDAI},
           city={Tokyo},
           country={Japan}}

\title{AdaTerm: Adaptive T-Distribution Estimated Robust Moments for Noise-Robust Stochastic Gradient Optimization}



\begin{abstract}
With the increasing practicality of deep learning applications, practitioners are inevitably faced with datasets corrupted by noise from various sources  such as measurement errors, mislabeling, and estimated surrogate inputs/outputs that can adversely impact the optimization results.
It is a common practice to improve the optimization algorithm's robustness to noise, since this algorithm is ultimately in charge of updating the network parameters. 
Previous studies revealed that the first-order moment used in Adam-like stochastic gradient descent optimizers can be modified based on the Student's t-distribution.
While this modification led to noise-resistant updates, the other associated statistics remained unchanged, resulting in inconsistencies in the assumed models.
In this paper, we propose AdaTerm, a novel approach that incorporates the Student's t-distribution to derive not only the first-order moment but also all the associated statistics.  This provides a unified treatment of the optimization process, offering a comprehensive framework under the statistical model of the t-distribution for the first time.
The proposed approach offers several advantages over previously proposed approaches, including reduced hyperparameters and improved robustness and adaptability. AdaTerm achieves this by considering the interdependence of gradient dimensions.
In particular, upon detection, AdaTerm excludes aberrant gradients from the update process and enhances its robustness for subsequent updates. Conversely, it performs normal parameter updates when the gradients are statistically valid, allowing for flexibility in adapting its robustness. 
This noise-adaptive behavior contributes to AdaTerm's exceptional learning performance, as demonstrated through various optimization problems with different and/or unknown noise ratios.
Furthermore, we introduce a new technique for deriving a theoretical regret bound without relying on AMSGrad, providing a valuable contribution to the field.\end{abstract}



\begin{keyword}
Robust optimization \sep Stochastic gradient descent \sep Deep neural networks \sep Student's t-distribution



\end{keyword}

\end{frontmatter}


\section{Introduction}
Deep learning~\cite{lecun2015deep} has emerged as a highly successful technology in recent years.
A significant factor contributing to its success is the widespread adoption of stochastic gradient descent (SGD) algorithms~\cite{robbins1951stochastic} for optimizing deep neural networks. These algorithms utilize first-order gradients computed through the back-propagation technique to update the network parameters, thereby addressing the optimization problem at hand.
Therefore, in tandem with the advancement of diverse network architectures such as residual networks~\cite{he2016deep}, various SGD-based optimizers have been pursued that can effectively and reliably optimize these networks.
The most representative optimizer is Adam~\cite{kingma2014adam}, and diverse variants with their respective features have been proposed (see the survey for details in~\cite{sun2019survey,schmidt2021descending}).
Among these, RAdam~\cite{liu2019radam} and AdaBelief~\cite{zhuang2020adabelief} have illustrated the state-of-the-art (SOTA) learning performance, to the best of our knowledge.

One notable feature of SGD optimizers is their robustness to gradient noise that generally results from the use of noisy datasets containing sensory and mislabeling errors~\cite{mirylenka2017classifier,suchi2019easylabel}.
Moreover, optimization problems that require the use of estimated inputs and/or outputs, such as long-term dynamics learning~\cite{chen2018neural,kishida2020deep}, reinforcement learning (RL)~\cite{sutton2018reinforcement}, and agents trained using distillation from a trained teacher(s)~\cite{rusu2015policy,gou2021knowledge}, also suffer from the noise caused by the estimation errors.
This feature becomes particularly crucial in robot learning tasks, where the available datasets may be limited in size, augmenting the adverse effects of noise.
Additionally, empirical evidence from ~\cite{simsekli2019tail} and~\cite{zhou2020towards} has demonstrated that both Adam and SGD exhibit heavy-tailed gradients' noise, even in the absence of input/output noise.

To address such noise and robustly conduct efficient gradient updates, previous studies have proposed the detection and exclusion of the aberrant gradients affected by noise.
In particular, recognizing that the conventional exponential moving average (EMA) momentum used in recent Adam-like optimizers is sensitive to noise owing to its association with a Gaussian distribution, \cite{ilboudo2020robust} and \cite{ilboudo2021adaptive} introduced a novel momentum algorithm called \textit{t-momentum} that is derived from the Student's t-distribution.
However, the t-momentum approach lacks a unified derivation of all its algorithm components, as described in section~\ref{sec:prev_work}, leading to certain limitations in its application. 

This paper presents a novel \textit{Adaptive T-distribution estimated robust moments} algorithm, AdaTerm\footnote{Code available at https://github.com/kbys-t/adaterm}, that addresses the limitations of previous approaches by offering a unified derivation of all the distribution parameters based on their gradients for an online maximum likelihood estimation of the Student's t-distribution.
AdaTerm introduces adaptive step sizes, transforming the gradient-based update of the statistics into an interpolation mechanism between past statistics values and the update amounts.
Such adaptive step sizes allow smoothness parameters $\beta$ to be common for all the involved statistics.
This is enabled by AdaTerm's comprehensive integration of the diagonal Student's t-distribution, allowing the scale estimator to involve every entry and avoiding the independent estimation of scale entries for each variable as in the diagonal Gaussian distribution. 
In addition, AdaTerm appropriately approximates the gradient of the degrees-of-freedom in the multi-dimensional case based on the behavior observed in the one-dimensional case (since it has been reported that the degrees-of-freedom was unintuitively small in multiple dimensions; see details in~\cite{ley2012value}).

In order to provide a theoretical basis for the convergence of optimization algorithms, including the proposed AdaTerm, the derivation of a regret bound is required. However, since the introduction of AMSGrad~\cite{reddi2019convergence}, the literature has assumed its usage, even for optimizers that do not practically employ AMSGrad.
To address this theoretical and practical contradiction, this paper introduces a new approach for deriving a regret bound without relying on the usage of AMSGrad. This approach can be applied to other optimizers as well, providing a more consistent framework for analyzing the convergence properties of various algorithms. This allows for a more accurate theoretical understanding of the convergence behavior of optimizers, including the proposed AdaTerm.

To summarize, this study presents four-fold contributions:
\begin{enumerate}
    \item Unified derivation of a novel SGD algorithm that is adaptively robust to noise;
    \item Alleviation of the difficulty of tuning multiple hyper-parameters for each statistic ($\beta_1$, $\beta_2$ and $\beta_z$) through replacement with a common and unique smoothness parameter $\beta$ for all the distribution parameters;
    \item Theoretical proof of the regret bound without incorporating AMSGrad;
    \item Numerical verification of efficacy in major test functions and typical problems such as classification problems with mislabeling, long-term prediction problems, reinforcement learning, and policy distillation.
\end{enumerate}
In the final verification, we compared not only AdaBelief and RAdam as the state-of-the-art algorithms but also t-Adam variants developed in another related study.

%
%
\section{Problem statement}
\label{sec:problem}

\subsection{Optimization problem}
\label{sec:optim_problem}

Here, we briefly define the optimization (or minimization, without loss of generality) problem to be solved using either of the SGD optimizers.
Suppose that certain input data $x$ and output data $y$ are generated according to the problem-specific (stochastic) rule, $p(x, y)$.
Accordingly, the problem-specific minimization target, $\mathcal{L}$, is given as follows:
\begin{align}
    \mathcal{L} = \mathbb{E}_{x,y \sim p(x,y)}[\ell(f(x; \theta), y)]
\end{align}
where $\ell$ denotes the loss function for each data, and $f(x; \theta)$ denotes the mapping function (e.g., from $x$ to $y$) with the parameter set $\theta$ that is optimized through this minimization (e.g., network weights and biases).

The above expectation operation can be approximated by the Monte Carlo method.
In other words, consider a dataset containing $N$ pairs of $(x, y)$, $\mathcal{D} = \{(x_n, y_n)\}_{n=1}^N$, constructed according to the problem-specific rule.
Based on $\mathcal{D}$, the above minimization target is replaced as follows:
\begin{align}
    \mathcal{L}_{\mathcal{D}} = \cfrac{1}{|\mathcal{D}|} \sum_{x_n, y_n \in \mathcal{D}} \ell(f(x_n; \theta), y_n)
\end{align}
where $|\mathcal{D}|$ denotes the size of dataset ($N$ in this case).

\subsection{SGD optimizer}

To solve the above optimization problem, we can compute the gradient with respect to $\theta$, $g = \nabla_\theta \mathcal{L}_{\mathcal{D}}$, and use gradient descent to obtain the (sub)optimal $\theta$ that (locally) minimizes $\mathcal{L}_{\mathcal{D}}$.
However, if $\mathcal{D}$ is large, the above gradient computation would be infeasible due to the limitation of computational resources.
Therefore, the SGD optimizer~\cite{robbins1951stochastic} extracts a subset (also known as mini batch) at each update step $t$, $\mathcal{B}_t \subset \mathcal{D}$, and updates $\theta$ from $\theta_{t-1}$ to $\theta_t$ as follows:
\begin{align}
    g_t &= \nabla_{\theta_{t-1}} \mathcal{L}_{\mathcal{B}_t}
    \\
    \theta_t &= \theta_{t-1} - \alpha \eta(g_t)
\end{align}
where $\alpha > 0$ denotes the learning rate, and $\eta$ represents a function used to modify $g_t$ to improve the learning performance.
In other words, each SGD-based optimizer have their own $\eta(g_t)$.

For example, in the case of Adam~\cite{kingma2014adam}, the most popular optimizer in recent years, $\eta$ has three hyper-parameters, $\beta_1 \in (0, 1)$, $\beta_2 \in (0, 1)$, and $\epsilon \ll 1$, and is defined by:
\begin{align}
    \eta^\mathrm{Adam}(g_t) = \cfrac{m_t (1 - \beta_1^t)^{-1}}{\sqrt{v_t (1 - \beta_2^t)^{-1}} + \epsilon}
    \label{eq:adam}
\end{align}
where
\begin{align}
    m_t &= \beta_1 m_{t-1} + (1 - \beta_1) g_t = m_{t-1} + (1 - \beta_1) (g_t - m_{t-1})
    \label{eq:ema_m}
    \\
    v_t &= \beta_2 v_{t-1} + (1 - \beta_2) g_t^2 = v_{t-1} + (1 - \beta_2) (g_t^2 - v_{t-1})
    \label{eq:ema_v}
\end{align}
where $g_t^2$ is the element-wise square.
A simple interpretation of the aforementioned Adam optimization algorithm is as follows: since $g_t$ fluctuates depending on the sampling method of $\mathcal{B}_t$, the first-order moment $m_t$ smoothens $g_t$ with the past gradients to stabilize the update, and the second-order moment $v_t$ scales $g_t$, according to the problem, to increase the generality without tuning $\alpha$.

If the problem setting described in section~\ref{sec:optim_problem} is satisfied, we can stably acquire one of the local solutions through one of the SGD optimizers, although their respective regret bounds may be different~\cite{reddi2019convergence,alacaoglu2020new}.
However, in real problems, development of datasets that follow the problem-specific rules exactly (i.e., $\mathcal{D}$ is generated from $p(x, y)$) is difficult.
In other words, when dealing with real problems, mostly inaccurate datasets $\mathcal{\tilde{D}} \neq \mathcal{D}$, contaminated by improper estimated values, mislabeling errors, etc., are available.

\section{Previous work}
\label{sec:prev_work}

The ubiquity of imperfect data in practical settings combined with the inherent noisiness of the stochastic gradient has encouraged the propositions of more robust and efficient machine learning algorithms against noisy or heavy-tailed datasets.
All these methods can be divided into two main approaches, ranging from methods that produce robust estimates of the loss function, to methods based on the detection and attenuation of incorrect gradient updates~\cite{gulcehre2017robust, holland2019efficient, prasad2020robust, kim2021hyadamc}.
Each approach has its own pros and cons, as summarized in Table~\ref{tab:road_to_robustness}.
\begin{table*}[tb]
    \caption{Pros/Cons of the two main approaches to regulate robustness
    }
    \label{tab:road_to_robustness}
    \centering
    \begin{tabular}{l cc}
        \hline\hline
        & \multicolumn{2}{c}{Approach}
        \\
        & Robust Loss Estimation & Robust Gradient Descent
        \\
        \hline
        Pros
        & Robustness independent of batch size
        & Widely applicable
        \\
        \hline
        Cons
        & Usually problem specific
        & Robustness dependent of batch size and outliers repartition
        \\

        & Usually requires the use of all the available data
        & Relies on only estimates of true gradient
        \\

        & Can be both unstable and costly in high dimensions
        &
        \\
        \hline\hline
    \end{tabular}
\end{table*}

As mentioned in the Introduction, we proceed with the latter approach in this study, in view of its wide applicability in machine learning applications.
In particular, we extend the concepts pertaining to the use of the Student's t-distribution as a statistical model that was first proposed by~\cite{ilboudo2020robust} for the gradients of the optimization process.
To explicitly highlight the difference between the previous studies (i.e. \cite{ilboudo2020robust} and~\cite{ilboudo2021adaptive}) and our current study, we review the previous methods in this section.

\subsection{The t-momentum}

\subsubsection{Overview}

Given a variable $x_t$ (e.g. $g_t$ or $g_t^2$), the popular EMA-based momentum underlying the most recent optimization algorithms is defined by $m_{t} = \beta m_{t-1} + (1 - \beta) x_t$, as described in eqs.~\eqref{eq:ema_m} and~\eqref{eq:ema_v}.
This arises from the Gaussian distribution by solving the following equation ~\cite{ilboudo2020robust}:
\begin{align}
\frac{\partial}{\partial m} \sum\limits_{i=1}^{n} \ln{\mathcal{N}(x_i \mid m, v)} = 0
\label{eq:gauss_ml}
\end{align}
where $\mathcal{N}(x_i|m, v)$ is the diagonal Gaussian distribution with mean $m$ and variance $v$, and $n$ is a fixed number of samples (corresponding to the $n$ most recent observations) defined by $n = \frac{1}{1-\beta}$.

Based on this observation, the classical EMA-based momentum  (now regarded as the \emph{Gaussian-momentum}), was replaced by the t-momentum, $m_{t} = \frac{W_{t-1}}{W_{t-1} + w_t} m_{t-1} + \frac{w_t}{W_{t-1} + w_t}  x_{t}$, obtained by solving the following equation:
\begin{align}
\frac{\partial}{\partial m} \sum\limits_{i=1}^{n} \ln{\mathcal{T}(x_i \mid m, v, \nu)} = 0
\label{eq:studentt_ml}
\end{align}
where $\mathcal{T}(x_i \mid m, v, \nu)$ is the diagonal Student's t-distribution with mean $m$, diagonal scale $v$, and degrees-of-freedom $\nu$.
Subsequently, based on the requirement of a fixed number of samples and the Gaussian limit (as $\nu \rightarrow \infty$), the sum $W_t = \sum_{i=1}^t w_i$ was replaced by a decaying sum $W_t = \frac{2\beta_{1} - 1}{\beta_{1}} W_{t-1} + w_{t}$, where $w_i = (\nu + d)/(\nu + \sum_{j=1}^d \frac{(x_{i,j} - m_{i-1,j})^2}{v_{i-1,j}+\epsilon})$ with $x_{i,j}$ denoting the $j$-th element of the vector $x_i$.
 Note that both eqs.~\eqref{eq:gauss_ml} and~\eqref{eq:studentt_ml} originate from the maximum log-likelihood (MLL) formulation and provide analytical solutions to the mean estimator.

\subsubsection{Limitations of the t-momentum}

Nonetheless, in the previous studies, the second-order moment $v_{t}$ is still based on the regular EMA, i.e., $v_{t} = \beta_{2} m_{t-1} + (1 - \beta_{2}) s_{t}$, where $s_{t}$ is a function of the squared gradient (e.g. $s_{t} = g_{t}^2$ for Adam~\cite{kingma2014adam} in eq.~\eqref{eq:ema_v} and $s_{t} = (g_{t}-m_{t})^2$ for AdaBelief~\cite{zhuang2020adabelief}).
This is in contrast to its usage in the computation of $w_{i}$, which falsely assumes that $v_{t}$ is also derived from the MLL of the Student's t-distribution, resulting in the unnatural blending of two statistical models.

Although one could also integrate the t-momentum with the second-order moment, this would require the computation of two ``variances'', one for the gradient $g$ and the other for the squared gradient function $s$, increasing the overall complexity of the algorithm.
Another strategy would be to derive the MLL analytical solution corresponding to the scale of the t-distribution, i.e., $v_{t} = \beta v_{t-1} + (1-\beta) w_{t} s_t$, similar to the mean estimator.
However, we found this approach to be unstable in practice, as also mentioned in~\ref{apdx:alt_v}, and it can result in \emph{NaN} errors on tasks such as the long-term prediction task.

Finally, the degrees-of-freedom $\nu$ is treated as a hyper-parameter and kept constant throughout the optimization process.
This is likely because the MLL formulation of the Student's t-distribution provides no analytic solution to the degrees-of-freedom estimator.
This issue can be connected to the high non-linearity of the distribution with respect to $\nu$.
To alleviate this limitation, the subsequent work~\cite{ilboudo2021adaptive} proposed the application of the direct incremental degrees-of-freedom estimation algorithm developed by~\cite{aeschliman2010novel} to automatically update the degrees-of-freedom, as discussed below.

\subsection{The At-momentum}

\subsubsection{Overview}

As stated, in~\cite{ilboudo2021adaptive}, the authors suggested circumventing the difficulties related to the MLL degrees-of-freedom by adopting a different approach and employing an alternative estimation method that do not require the use of the MLL framework.
Indeed, the algorithm developed by Aeschliman et al.~\cite{aeschliman2010novel} and employed in the At-momentum is based on an approximation of $\mathbb{E}[\ln{\norm{g}^2}]$ and $\mathbb{V}[\ln{\norm{g}^2}]$. 
Specifically, given a $d$-dimensional variable $g$ following a Student's t-distribution with scale $v = \xi I$, where $\xi$ is a constant and $I$ the identity matrix, we have the following results:
\begin{align*}
\mathbb{E}[\ln{\norm{g}^2}] &= \ln{\xi} + \ln{\nu} + \psi(\frac{d}{2}) - \psi(\frac{\nu}{2})
\\
\mathbb{V}[\ln{\norm{g}^2}] &= \psi_{1}(\frac{\nu}{2}) + \psi_{1}(\frac{d}{2})
\end{align*}
where $\psi(x)$ and $\psi_{1}(x)$ are the digamma and trigamma functions, respectively, and $\norm{\cdot}$ is the $\ell_2$-norm (or Euclidean norm).
By setting $z_{i} = \ln{\norm{g_i}^2}$, the degrees-of-freedom $\nu$ can be estimated by solving the following equation:
\begin{align*}
\psi_{1}(\frac{\nu}{2}) &= \frac{1}{n} \sum\limits_{i=1}^{n} (z_i - \bar{z})^2 - \psi_{1}(\frac{d}{2}), \;\mathrm{with}\; \bar{z} = \frac{1}{n} \sum\limits_{i=1}^{n} z_i
\end{align*}
Since this equation cannot be solved directly, an approximation of the trigamma function, $\psi_{1}(x) \approx \frac{x+1}{x^2}$, is used on the left side and the estimator for the degrees-of-freedom is expressed as follows:
\begin{align}
b &= \frac{1}{n} \sum\limits_{i=1}^{n} (z_i - \bar{z})^2 - \psi_{1}(\frac{d}{2}) \\
\hat{\nu} &= \frac{1 + \sqrt{1 + 4b}}{b}
\end{align}

In the previous work~\cite{ilboudo2021adaptive}, the incremental version of this estimator obtained by replacing the arithmetic mean and variance of $z$ by EMAs with $\beta_z$, was employed to derive the At-momentum.

\subsubsection{Limitations of At-momentum}

Notably, the direct incremental degrees-of-freedom estimation algorithm was originally developed to utilize specific estimation methods for the location and scale parameters.
This again introduces false assumptions, as the degrees-of-freedom estimator assumes that the mean and scale are based on the principles developed by Aeschliman et al. in~\cite{aeschliman2010novel}, while they are given as the MLL of Student's t-distribution and Gaussian distribution, respectively.
In the experiment section below, we show that this inconsistency causes insufficient adaptability of the robustness to noise.

\subsection{Contributions}

\begin{table*}[tb]
    \caption{Differences summary between AdaTerm and previous studies
    }
    \label{tab:contributions}
    \centering
    \begin{tabular}{c | c | c | c | c}
        \hline\hline
        & & t-momentum & At-momentum & AdaTerm
        \\
        \hline
        \multirow{3}{*}{Parameters' origins}
        & $m$ & Student's t MLL & Student's t MLL & Student's t MLL \\ & $v$ & Gaussian MLL & Gaussian MLL & Student's t MLL \\ & $\nu$ & Fixed & Aeschliman et al.~\cite{aeschliman2010novel} & Student's t MLL
        \\
        \hline
        MLL solution & & Analytic & Analytic & Sequential w/ custom step sizes
        \\
        \hline
        Decay factors & & $\beta_1$, $\beta_2$ & $\beta_1$, $\beta_2$, $\beta_z$ & $\beta$
        \\
        \hline\hline
    \end{tabular}
\end{table*}

As explained in the previous paragraphs, the discrepancy in statistical model between the first- and second-order moments and the introduction of a different framework to estimate the degrees-of-freedom creates a chain of equations with different assumptions pertaining to how their defining parameters are obtained.
This afflicts the optimization algorithm with certain limitations (e.g. limited robustness and limited adaptability).

In this study, we tackle this lack of unified approach by estimating all the parameters through the unified MLL estimation of the t-distribution.
In the previous study, the Student's t-based EMA of the first-order moment was derived by relying on the analytic solution of the MLL (i.e., by setting the gradient of the log-likelihood to zero and solving explicitly).
However, this approach is ill-suited for obtaining the scale $v_{t}$ and the degrees-of-freedom $\nu$ owing to its practical inefficiency and the absence of a closed-form solution, respectively.

To meet the unification requirement, we propose replacing MLL's direct solution of the previous studies with a sequential solution based on a gradient ascent algorithm for all the parameters of the Student's t-model. This new approach has two main advantages: (i) it allows the derivation of an update rule for the degrees-of-freedom that has no close-form solution when one attempts to set its gradient to zero; and
(ii) it allows for the use of customized step-sizes that stabilize the components of the algorithm, avoiding the instability of the analytic solution to the scale $v$ estimator (i.e., apparition of \textit{NaN} values, as mentioned in Appendix~\ref{apdx:alt_v}).
Based on the properties of the t-distribution and the success of the EMA approach for well-behaved datasets, we derive suitable adaptive step sizes to allow for smooth updates.
In addition, additional modifications ensure that the algorithm still recovers the EMA in the Gaussian limit and does not violate the positiveness of both the scale and the degrees-of-freedom parameters.
As another byproduct of this unified approach, we are able to employ one single decay factor $\beta$ for all the parameters, eliminating the requirement for the usual two distinct decay factors ($\beta_1$ and $\beta_2$).

The differences between our algorithm and the previous studies are summarized in Table~\ref{tab:contributions}, and the details of the derivation are given in the next section.

\section{Derivation of AdaTerm}
\label{sec:derivation}

\subsection{The Student's t-distribution statistical model}

\begin{figure}[h]
    \centering
    \includegraphics[keepaspectratio=true,width=0.99\linewidth]{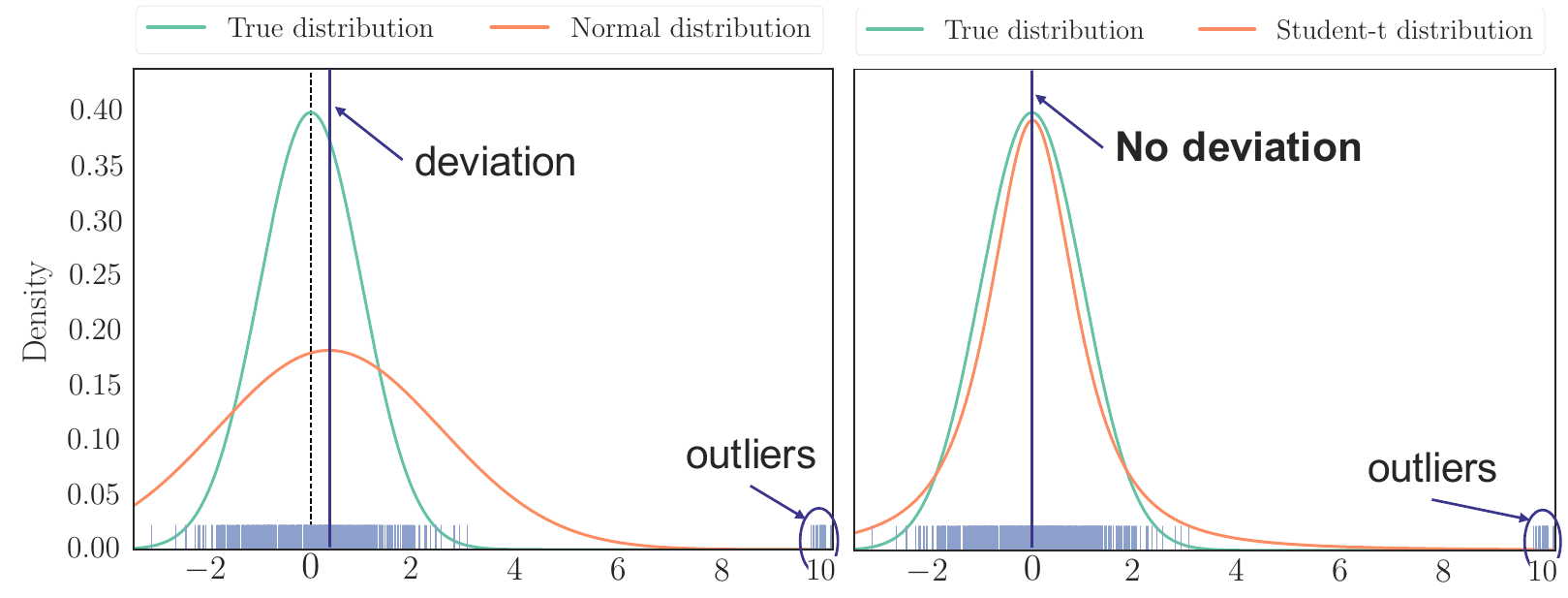}
    \caption{Robustness of Student's t-distribution (right) compared with the Gaussian distribution (left). The green curve represents the true distribution to be captured.}
    \label{fig:robust_tdist}
\end{figure}

To robustly estimate the true average gradient from the disturbed gradients, we model the distribution of these gradients $g$ by the Student's t-distribution, which has a heavy tail and is robust to outliers (as shown in Fig.~\ref{fig:robust_tdist}), referring to the related work~\cite{ilboudo2020robust}.
Notably, as further motivation for this statistical model, empirical evidence in~\cite{simsekli2019tail} has shown that the norm of the gradient noise in SGD has heavy tails and~\cite{ziyin2021strength} revealed that in the continuous-time analysis, the gradient exhibits a stationary distribution following Student's t-like distributions.

Specifically, in the present study, $g$ is assumed to be generated from a $d$-dimensional diagonal Student's t-distribution, characterized by three types of parameters:
a location parameter $m \in \mathbb{R}^d$;
a scale parameter $v \in \mathbb{R}_{> 0}^d$;
and a degrees-of-freedom parameter $\nu \in \mathbb{R}_{> 0}$.
That is,
\begin{align}
    g &\sim \cfrac{\Gamma(\frac{\nu + d}{2})}{\Gamma(\frac{\nu}{2})(\nu \pi)^{\frac{d}{2}} \prod\limits_{i=1}^{d} \sqrt{v_i}}
    \left (1 + \cfrac{1}{\nu}\sum\limits_{i=1}^{d} (g_i - m_i)^2 v_{i}^{-1} \right )^{-\frac{\nu + d}{2}}
    \nonumber \\
    &=: \mathcal{T}(g \mid m, v, \nu)
\end{align}
where $\Gamma$ denotes the gamma function.
Following the related work~\cite{ilboudo2020robust} and PyTorch's implementation~\cite{paszke2017automatic}, AdaTerm is in practice applied to the weight matrix and the bias of each neural network layer separately. Therefore, $d$ here corresponds to the dimension size of each subset of parameters (weight or bias of the layer).


\subsection{Gradients for the maximum log-likelihood}

To derive AdaTerm, let us consider the MLL problem, $\max_{m, v, \nu} \ln \mathcal{T}(g \mid m, v, \nu)$, to estimate the parameters $m$, $v$, and $\nu$ that can adequately model the most recent observed loss function gradients $g$.

To simplify the notation, the following variables are defined.
\begin{align}
    s = (g - m)^2
    , \
    D = \cfrac{1}{d} s^\top v^{-1}
    , \
    \tilde{\nu} = \nu d^{-1}
    , \
    w_{mv} = \cfrac{\tilde{\nu} + 1}{\tilde{\nu} + D}
    \nonumber
\end{align}
When $\nu$ is defined to be proportional to $d$, as in the literature~\cite{ilboudo2020robust}, $\tilde{\nu}$ corresponds to the proportionality coefficient.
Note that it is not necessary to follow the literature~\cite{ilboudo2020robust} for setting the degrees-of-freedom as $\nu = \tilde{\nu}d$. Nonetheless, this procedure is important to provide the same level of robustness to the respective subsets of parameters, regardless of their respective size.

Based on these, the log-likelihood gradients with respect to $m$, $v$, and $\nu$ can respectively be derived as follows:
\begin{align}
    \nabla_{m}\ln \mathcal{T}
    &= - \cfrac{\nu + d}{2} \cfrac{-\nu^{-1} (g - m) v^{-1}}{1 + \nu^{-1} d D}
    \nonumber \\
    &= \cfrac{\nu d^{-1} + 1}{\nu d^{-1} + D} \cfrac{g - m}{2v}
    \nonumber \\
    &= w_{mv} \cfrac{g - m}{2v}
    =: g_m
    \label{eq:grad_m}\\
    \nabla_{v}\ln \mathcal{T}
    &= - \cfrac{1}{2v} - \cfrac{\nu + d}{2} \cfrac{-\nu^{-1} (g - m)^2 v^{-2}}{1 + \nu^{-1} d D}
    \nonumber \\
    &= \cfrac{1}{2v^2}\left ( \cfrac{\tilde{\nu} + 1}{\tilde{\nu} + D} s - v \right )
    \nonumber \\
    &= w_{mv} \cfrac{\tilde{\nu}}{2v^2 (\tilde{\nu} + 1)} \{ (s - v) + (s - Dv) \tilde{\nu}^{-1} \}
    \label{eq:grad_v}\\
    \nabla_{\nu}\ln \mathcal{T}
    &= \cfrac{1}{2}\psi\left( \cfrac{\nu + d}{2} \right) - \cfrac{1}{2}\psi\left( \cfrac{\nu}{2} \right) - \cfrac{d}{2\nu}
    \nonumber \\
     &- \cfrac{1}{2} \ln (1 + \nu^{-1} dD) - \cfrac{\nu + d}{2} \{(\nu + dD)^{-1} - \nu^{-1} \}
     \label{eq:grad_nu}
\end{align}
where the gradients with respect to $m$ and $v$ are transformed so that the response to outliers can be intuitively analyzed by $w_{mv}$.
$\psi$ denotes the digamma function.

In the previous approach, the gradients would be set to zero to find a direct analytical solution.
We instead employ a gradient ascent update rule to solve this MLL problem.
Next, we discuss how the adaptive step sizes and the update rule for each one of the parameters are derived to develop the AdaTerm algorithm.

\subsection{Online maximum likelihood updates with adaptive step sizes}

\begin{figure}[tb]
    \centering
    \includegraphics[keepaspectratio=true,width=0.95\linewidth]{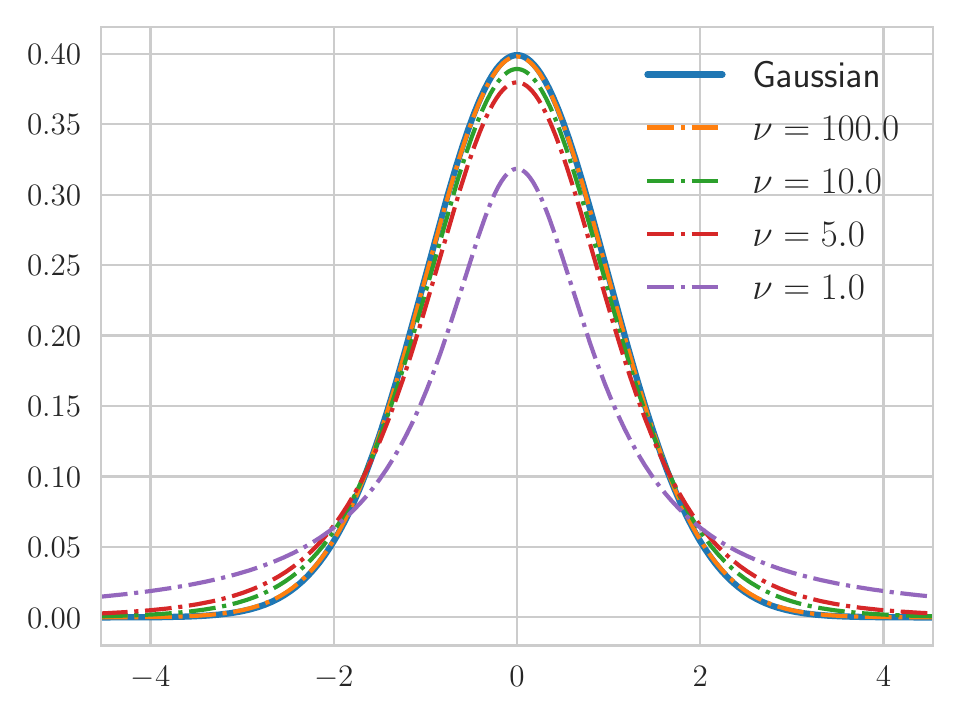}
    \caption{The Student's t-distribution tends toward the Gaussian distribution as $\nu \to \infty$}
    \label{fig:tdistribution}
\end{figure}
 As shown in Fig.~\ref{fig:tdistribution}, the Student's t-distribution exhibits the property that when $\nu \to \infty$, it reverts to a Gaussian distribution.
 Similarly, when $\nu \to \infty$ (i.e. $w_{mv} \to 1$), the AdaTerm optimizer should also yield a non-robust Gaussian optimization algorithm.

 Following the success of the EMA scheme in outlier-free optimization, we specifically choose the adaptive step sizes, such that in the Gaussian limit (i.e., when $\nu \to \infty$), the gradient ascent updates of the mean $m$ and scale $v$ parameters revert to the simple EMA updates given by eq.~\eqref{eq:ema_m} and eq.~\eqref{eq:ema_v}.

\subsubsection{The first-order moment}

Since the location parameter $m$ can take any value in real space, we can perform unconstrained gradient-based updates without any restriction.
The gradient ascent equation is therefore expressed as follows:
\begin{align}
    m_t &= m_{t-1} + \kappa_m g_m
    \nonumber \\
    &= m_{t-1} + \kappa_m w_{mv} \frac{(g_t - m_{t-1})}{2v_{t-1}}
\end{align}
where $\kappa_m$ is the update step size to be defined.
Because we expect an EMA-like update rule similar to eq.~\eqref{eq:ema_m}, $m_t$ can also be written as:
\begin{align}
    m_t &= m_{t-1} + \tau_m (g_{t} - m_{t-1})
    \nonumber
\end{align}
where $\tau_m$ is the interpolation factor (or decay factor) satisfying $\tau_m \in (0, 1)$ and $\tau_m \to (1-\beta)$ for $\tilde{\nu} \to \infty$.
Based on identification with the gradient ascent equation, we can set
\begin{align*}
    \tau_m &= \frac{\kappa_m w_{mv}}{2v_{t-1}}
\end{align*}
Using the second constraint on $\tau_m$, i.e., $\tau_m \to (1-\beta)$, and the fact that $w_{mv} \to 1$ for $\tilde{\nu} \to \infty$, we can write the following proposition:
\begin{align*}
    \tilde{\nu} \to \infty &\implies \tau_m = \frac{\kappa_m w_{mv}}{2v_{t-1}} \to (1-\beta) \implies \kappa_m \to 2 (1-\beta) v_{t-1}
\end{align*}
Let $k(\tilde{\nu})$ be an arbitrary function of $\tilde{\nu}$ such that $\tilde{\nu} \to \infty \implies k(\tilde{\nu}) \to 1$; accordingly, $\kappa_m = 2 (1-\beta) v_{t-1} k(\tilde{\nu})$ satisfies the previous proposition.

To identify a suitable function $k(\tilde{\nu})$ for our optimizer, we can utilize the first constraint on $\tau_m$, i.e., $\tau_m \in (0, 1)$. By replacing the expression for $\kappa_m$ into the expression for $\tau_m$, this constraint can be rewritten as follows:
\begin{align}
0 < (1-\beta) w_{mv} k(\tilde{\nu}) < 1 &\implies 0 < k(\tilde{\nu}) < \frac{1}{(1-\beta) w_{mv}}
\end{align}

This implies that any function satisfying $\tilde{\nu} \to \infty \implies k(\tilde{\nu}) \to 1$ and $0 < k(\tilde{\nu}) < \frac{1}{(1-\beta) w_{mv}}$ will result in a valid step size $\kappa_m$. In this paper, we consider the simplest version among these functions for which $\tau_m$ is not a constant, i.e., $k(\tilde{\nu}) = \frac{1}{\overline{w}_{mv}}$, where $\overline{w}_{mv} = (\tilde{\nu} + 1)\tilde{\nu}^{-1} \geq w_{mv} \geq (1-\beta) w_{mv}$ (since $\beta \in (0, 1)$ by definition).

Based on this, the adaptive step size $\kappa_m$ is expressed as follows:
\begin{align}
    \kappa_m &= 2v_{t-1} \frac{1 - \beta}{\overline{w}_{mv}}
\end{align}
and the update rule for the first-order moment $m$ becomes
\begin{align}
    m_t &= m_{t-1} + \kappa_m g_m = m_{t-1} + (1 - \beta) \frac{w_{mv}}{\overline{w}_{mv}} (g_t - m_{t-1})
    \nonumber \\
    &= (1 - \tau_m) m_{t-1} + \tau_m g_t
\end{align}
where $\tau_m = (1 - \beta) \frac{w_{mv}}{\overline{w}_{mv}}$.

\mycomment{
Using the first constraint on $\tau_m$, i.e. $0 < \tau_m < 1$, we can derive the following:
\begin{align*}
    0 < \tau_m < 1 &\implies 0 < \kappa_m < \frac{2v_{t-1}}{w_{mv}}
    \implies \kappa_m = \frac{2kv_{t-1}}{w_{mv}}\\
    &\implies \tau_m = k, \;\; k \in (0, 1)
\end{align*}

Furthermore, to recover a Gaussian model (i.e., a classical EMA) in the limit as $\tilde{\nu} \to \infty$, we require $\tau_m \to (1 - \beta)$ with $\beta \in (0, 1)$, the common smoothness parameter, implying $k \to (1 - \beta)$.
Since the interpolation ratio $\tau_m$ should be adaptive to outliers, i.e, it should ultimately involve $w_{mv}$; since $k \in (0, 1)$, we set $k = (1 - \beta) w_{mv}\overline{w}_{mv}^{-1}$ in this study, where $w_{mv} \leq \overline{w}_{mv} = (\tilde{\nu} + 1)\tilde{\nu}^{-1}$.}

\subsubsection{Central second-order moment}

Similarly, the gradient ascent update rule for the scale parameter $v$ is expressed as follows:
\begin{align}
    v_t &= v_{t-1} + \kappa_v g_v
    \nonumber \\
    &= v_{t-1} + \kappa_v \frac{w_{mv} \tilde{\nu}}{2v_{t-1}^{2} (\tilde{\nu} + 1)} \left[ (s_{t} + \Delta s) - v_{t-1} \right] \nonumber \\
    &= v_{t-1} + \kappa_v \frac{w_{mv}}{2v_{t-1}^{2} \overline{w}_{mv}} \left[ (s_{t} + \Delta s) - v_{t-1} \right]
\end{align}

where $\kappa_v$ is the update step size, and $\Delta s = (s - Dv) \tilde{\nu}^{-1}$.
Again, we restrict ourselves to an EMA-like update rule similar to eq.~\eqref{eq:ema_v}:
\begin{align}
    v_t &= v_{t-1} + \tau_v \left[ (s_t + \Delta s) - v_{t-1} \right]
    \nonumber
\end{align}
where $\tau_v$ must satisfy the same constraints as $\tau_m$, i.e., $\tau_v \in (0, 1)$ and $\tau_v \to (1-\beta)$ for $\tilde{\nu} \to \infty$.
The simplicity of this second constraint despite the presence of $\Delta s$ is because we already have $\tilde{\nu} \to \infty \implies \Delta s \to 0$. This leaves only $(1-\beta)(s_t - v_{t-1})$, as found in the Gaussian-based EMA for the variance.
 Therefore, the derivations for $\kappa_v$ and $\tau_v$ follow almost exactly the same procedure as performed above for $\kappa_m$ and $\tau_m$:
\begin{align*}
    \tilde{\nu} \to \infty &\implies \tau_v = \frac{\kappa_v w_{mv}}{2v_{t-1}^{2} \overline{w}_{mv}} \to (1-\beta) \implies \kappa_v \to 2 (1-\beta) v_{t-1}^{2} \\
    &\implies \kappa_v = 2 (1-\beta) v_{t-1}^{2} k(\tilde{\nu})\;,\; (\tilde{\nu} \to \infty \implies k(\tilde{\nu}) \to 1) \\
    0 < \tau_v < 1 &\implies 0 < k(\tilde{\nu}) < \frac{\overline{w}_{mv}}{(1-\beta) w_{mv}}
\end{align*}
Again, we can use the lowest bound for $\frac{1}{(1-\beta) w_{mv}}$, i.e., $\frac{1}{\overline{w}_{mv}}$, and set $k(\tilde{\nu}) = \frac{\overline{w}_{mv}}{\overline{w}_{mv}} = 1$. This results in the following step size:
\begin{align}
    \kappa_v &= 2v_{t-1}^2 (1 - \beta)
\end{align}
and the corresponding update rule:
\begin{align}
    v_t &= v_{t-1} + \kappa_v g_v
    \nonumber \\
    &= v_{t-1} + (1 - \beta) \frac{w_{mv}}{\overline{w}_{mv}} \left[ (s_t + \Delta s) - v_{t-1} \right]
    \nonumber \\
    &= (1 - \tau_v) v_{t-1} + \tau_v (s_t + \Delta s)
\end{align}
where $\tau_v = \tau_m = (1 - \beta) \frac{w_{mv}}{\overline{w}_{mv}}$.

\mycomment{
Therefore, 
\begin{align}
    \tau_v &= \frac{\kappa_v w_{mv} \tilde{\nu}}{2v_{t-1}^{2} (\tilde{\nu} + 1)} \implies 0 < \kappa_v < \frac{2v_{t-1}^{2} (\tilde{\nu} + 1)}{w_{mv} \tilde{\nu}}
    \nonumber \\
    &\implies \kappa_v = \frac{2kv_{t-1}^{2} (\tilde{\nu} + 1)}{w_{mv} \tilde{\nu}},\; k \in (0, 1) \implies \tau_v = k
\end{align}

Here again, $k$ must satisfy the same conditions as in the first-moment derivation.
This implies that we can again set $k = (1 - \beta) w_{mv}\overline{w}_{mv}^{-1}$
}

A closer look at this update rule reveals that although for $\tilde{\nu} \to \infty$, $\Delta s = (s - Dv) \tilde{\nu}^{-1} \to 0$, the value of $\Delta s$ can still be negative for $\tilde{\nu} \ll \infty$.
%
This implies that the update shown above can potentially lead the scale parameter $v$ into a negative region during the learning process. This is undesirable since $v$ is only defined in the positive real space.
There are many ways to satisfy this positive constraint, e.g., reparameterization or the mirror descent (MD) strategies~\cite{beck2003mirror}.
However, to preserve the EMA-like interpolation directly on the parameter $v$, we simply employ the projected gradient method.
Specifically, we use a simple gradient clipping method~\cite{gorbunov2020stochastic}, where we place a lower bound on $\Delta s$
\begin{align}
    \Delta s =: \max(\epsilon^2, (s - Dv) \tilde{\nu}^{-1})
\end{align}
where $\epsilon \ll 1$.
Such a gradient upper bound is disadvantageous in that it can cause the algorithm to overestimate the scale parameter $v$.
In this case, when $v$ is larger than the exact value, $D$ and $w_{mv}$ becomes smaller and larger, respectively; in other words, the robustness to noise may be impaired.
However, this drawback is attenuated by the fact that the scaled learning rate (similar to that of Adam in eq.~\eqref{eq:adam}) is inversely proportional to $v$, rendering the effect of noise insignificant.
We also argue that the slow decrease of $v$ permitted by the gradient clipping strategy is advantageous in preventing excessive robustness, compared with the case where $v$ is analytically estimated (see details in~\ref{apdx:alt_v}).

Finally, since $\tau_m = \tau_v = (1 - \beta) w_{mv}\overline{w}_{mv}^{-1}$, we adopt one common notation for both as $\tau_{mv} = (1 - \beta) w_{mv}\overline{w}_{mv}^{-1}$.
Note that the past work uses the discounted sum of $w_{mv}$, $W_t$, by storing it in memory; however, in AdaTerm, $\overline{w}_{mv}$ is used instead, saving memory.

\subsubsection{Degrees-of-freedom}

\paragraph{Simplifying the gradient}
Compared with $m$ and $v$, the update rule of $\nu$ is considerably more delicate to handle.
In particular, when the positive constraint on $\nu$ is violated by a simple gradient ascent update cannot be identified.
This is mainly due to the digamma function $\psi$. Hence, the first step is to eliminate it from the problem.
In this study, we use the upper and lower bounds of $\psi$ to find the upper bound of the gradient with respect to $\nu$.
Specifically, the upper and lower bounds of $\psi$ are expressed as follows:
\begin{align*}
    \ln x - \cfrac{1}{x} \leq \psi(x) \leq \ln x - \cfrac{1}{2x}
\end{align*}
implying that we can define an upper-bound for $\psi(\frac{\nu+d}{2}) - \psi(\frac{\nu}{2})$ as $\ln (\frac{\nu+d}{2}) - \frac{1}{\nu+d} - \ln (\frac{\nu}{2}) + \frac{2}{\nu}$.

\begin{figure}[tb]
    \centering
    \includegraphics[keepaspectratio=true,width=0.95\linewidth]{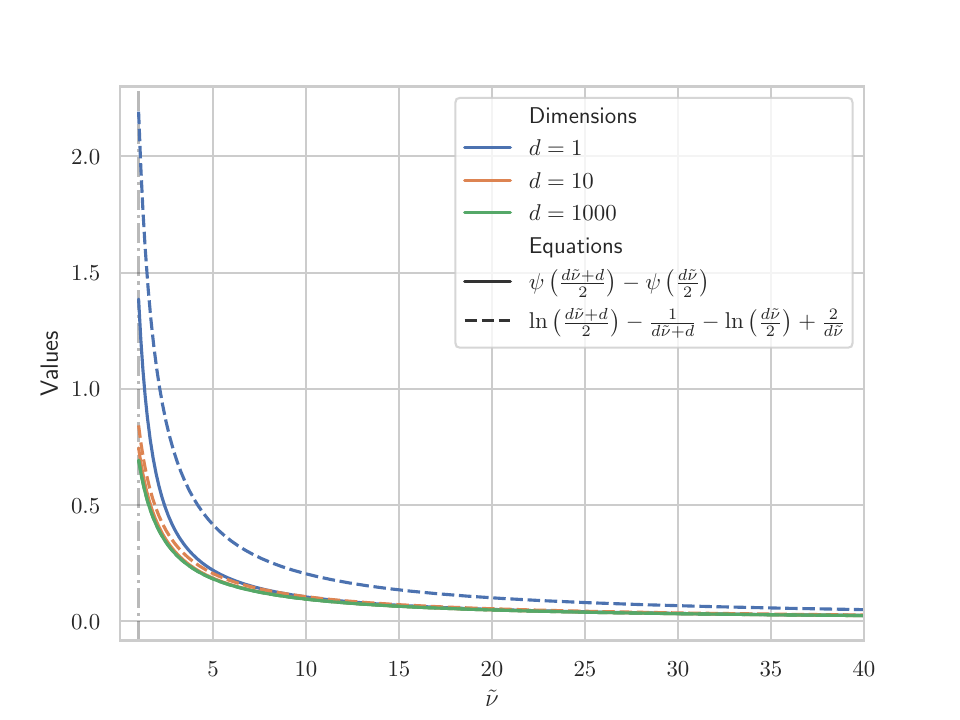}
    \caption{The tightness of the upper bound employed for eq.~\eqref{eq:grad_nu_m1}. For a large $d$, the upper bound's curve and the original equation's curve are practically indistinguishable.}
    \label{fig:digamma_bound}
\end{figure}

Fig.~\ref{fig:digamma_bound} shows the curves traced by the two equations as a function of $\tilde{\nu}$, for different values of $d$ (given that for our application $\nu = \tilde{\nu}d$ with typically $\tilde{\nu} \geq 1$).
As can be seen, for large values of $d$, which is generally the case for neural networks (NN), the defined upper-bound tightens. For $d=1$ (which can occur for the NN biases), the gap evaluated in the range $\tilde{\nu}\in[1,\infty)$, is at most $\approx 0.80685$ and occurs when $\tilde{\nu}=1$.

Although the use of an upper bound can impair the noise robustness in theory (as in the case of $v$), the relative small gap shown in Fig.~\ref{fig:digamma_bound} hardly impairs this in practice.
Therefore, the following upper bound for eq.~\eqref{eq:grad_nu} can be defined:
\begin{align}
    \nabla_{\nu}\ln \mathcal{T}
    &\leq \cfrac{1}{2}\Biggl \{ \ln(\nu + d) - \ln 2 - \cfrac{1}{\nu + d} - \ln \nu + \ln 2 + \cfrac{2}{\nu}
    \nonumber \\
    &- \cfrac{d}{\nu} - \ln(\nu + dD) + \ln \nu - \cfrac{\nu + d}{\nu + dD} + 1 + \cfrac{d}{\nu} \Biggr \}
    \nonumber \\
    &= \cfrac{1}{2}\Biggl \{ - w_{mv} + \ln w_{mv} + 1 + \cfrac{\tilde{\nu} + 2}{\tilde{\nu} + 1}\cfrac{1}{\nu} \Biggl \}
    \nonumber \\
    &= \cfrac{1}{2}\Biggl \{ - w_{\tilde{\nu}} + 1 + \cfrac{\tilde{\nu} + 2}{\tilde{\nu} + 1}\cfrac{1}{\nu} \Biggl \}
    \label{eq:grad_nu_m1}
\end{align}
where $w_{\tilde{\nu}} = w_{mv} - \ln w_{mv}$.
\begin{figure}[tb]
    \centering
    \includegraphics[keepaspectratio=true,width=0.95\linewidth]{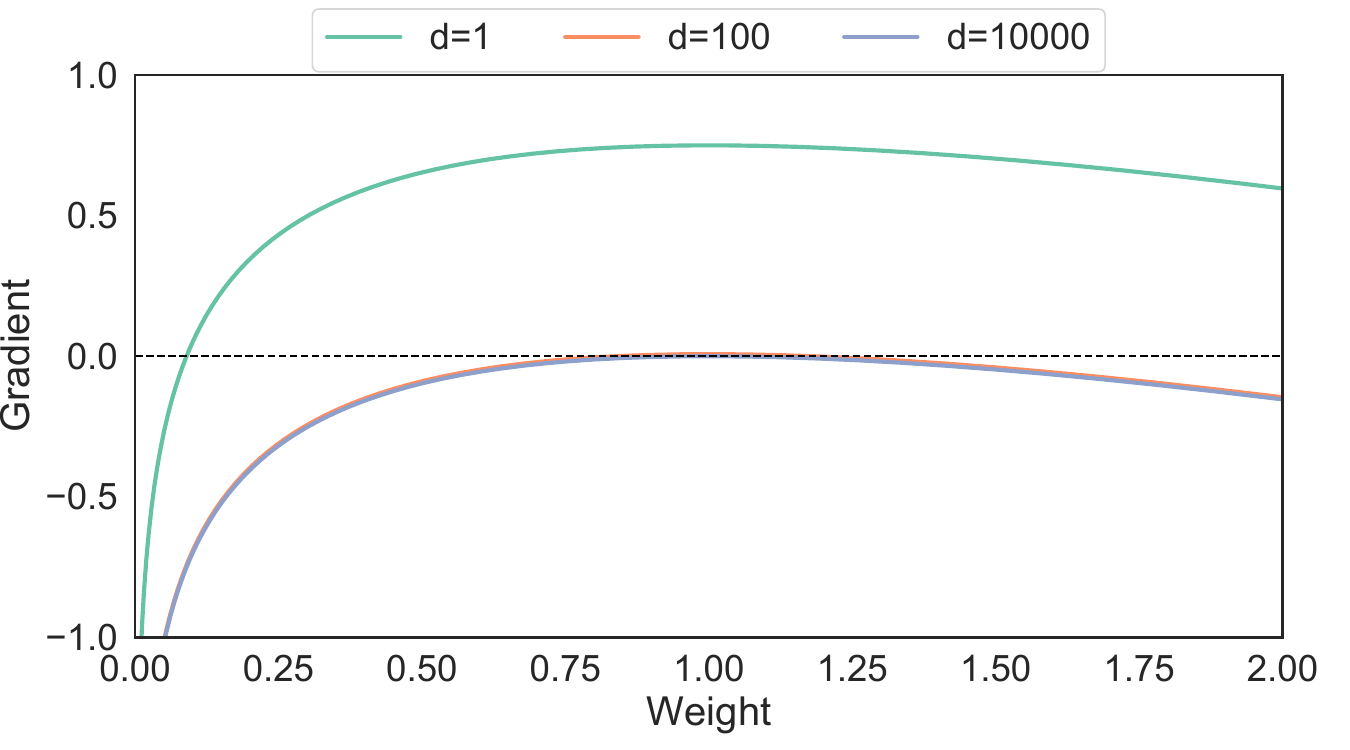}
    \caption{$w_{mv}$ vs. the surrogated gradient in eq.~\eqref{eq:grad_nu_m1} with $\nu = d$ according to several dimension sizes}
    \label{fig:grad_nu}
\end{figure}

\paragraph{Tackling the curse of dimensionality}
The overestimated gradient of the previous paragraph is simple and its behavior is easy to analyze.
Notably, the behavior of the overestimated gradient is considerably affected by $d$, as shown in Fig.~\ref{fig:grad_nu} with $\nu = d$ (i.e. $\tilde{\nu} = 1$).
As can be seen, when $d=1$, the gradient with respect to $\nu$ is negative only when $w_{mv}$ is exceedingly small, corresponding to an extremely large deviation $D$.
In this case, the negative gradient effectively decreases $\nu$ to exclude the noisy gradients of the network's parameters $g$.
Conversely, for small deviations $D$, the gradient in~\eqref{eq:grad_nu_m1} becomes positive and $\nu$ is increased to suppress the robustness and easily update the parameters.

Unfavorably, with $d \sim = 10,000$, which is common for a NN's weights matrix, the gradient of eq.~\eqref{eq:grad_nu_m1} is always negative and almost zero even when $w_{mv}=1$, implying that $\nu$ can only decrease.
The same phenomenon occurs even for $d \sim = 100$, whose curve is practically indistinguishable from the curve of $d \sim = 10,000$.
Such a pessimistic behavior is consistent with the non-intuitive behavior of the multivariate Student's t-distribution reported in the literature~\cite{ley2012value}.
This problem has also been empirically confirmed in~\cite{ilboudo2021adaptive}, where the excessive robustness is forcibly suppressed by correcting the obtained (approximated) estimate by multiplying it with $d$.

To alleviate this problem, we further consider an upper bound, where $\nu$ is replaced by $\tilde{\nu}$. 
This choice is further supported by the fact that $\nu$ does not appear directly in any of the other MLL gradients, implying that directly dealing with the gradient with respect to $\tilde{\nu}$ is more natural.
The above modifications are summarized below.
\begin{align}
    \nabla_{\tilde{\nu}}\ln \mathcal{T} &= d \nabla_{\nu}\ln \mathcal{T}
    \nonumber \\
    &\leq \cfrac{d}{2}\Biggl \{ - w_{\tilde{\nu}} + 1 + \cfrac{\tilde{\nu} + 2}{\tilde{\nu} + 1}\cfrac{1}{\tilde{\nu}} \Biggl \}
    \nonumber \\
    &= w_{\tilde{\nu}} \cfrac{d}{2}\Biggl \{ -1 + \Biggl (\cfrac{\tilde{\nu} + 2}{\tilde{\nu} + 1} + \tilde{\nu} \Biggr )\cfrac{1}{\tilde{\nu} w_{\tilde{\nu}}} \Biggl \}
    =: g_{\tilde{\nu}}
    \label{eq:grad_nu_m2}
\end{align}

Where the second line is obtained by using $\frac{1}{\nu} \leq \frac{1}{\tilde{\nu}}$.
This new upper bound can be interpreted as an approximation of the multivariate distribution by a univariate distribution and allows the degrees-of-freedom to increase even for large dimensions $d$ when $w_{mv}$ is small (i.e., $D$ is large).
Although it is possible to model the distribution as a univariate distribution from the beginning of the derivation~\cite{kobayashi2021t}, the robustness may be degraded, since it would capture the average without focusing on the amount of deviation along each axis.

\paragraph{Update rule}
Based on this upper bound, we now proceed to the derivation of a suitable update rule for the robustness parameter $\tilde{\nu}$.
We draw similarities with the update rules obtained previously for $m$ and $v$ and set our goal to an EMA-like equation with an adaptive smoothness parameter $\tau_{\tilde{\nu}} = (1 - \beta) w_{\tilde{\nu}}\overline{w}_{\tilde{\nu}}^{-1}$, where $w_{\tilde{\nu}} \leq \overline{w}_{\tilde{\nu}}$.
We aim for the following update rule:
\begin{align}
    \tilde{\nu}_{t} &= (1 - \tau_{\tilde{\nu}}) \tilde{\nu}_{t-1} + \tau_{\tilde{\nu}} \lambda_{t} = \tilde{\nu}_{t-1} + \tau_{\tilde{\nu}} (\lambda_{t} - \tilde{\nu}_{t-1})
    \label{eq:dof_ema}
\end{align}
where $\lambda_{t}$ is some function of $w_{mv}$.
To derive such an equation, we focus first on $\tau_{\tilde{\nu}}$ and particularly on determining the maximum value of $w_{\tilde{\nu}}$, $\overline{w}_{\tilde{\nu}}$.

We begin by noticing that $w_{\tilde{\nu}}$ is a convex function over $w_{mv}$ with a minimum value at $w_{mv}=1$.
The maximum value is therefore determined when $w_{mv}$ is the largest ($w_{mv} \gg 1$) or smallest ($w_{mv} \ll 1$).
The largest $w_{mv}$ value has already been derived as $\overline{w}_{mv}$; however, the smallest value cannot exactly be given since $w_{mv} \to 0$ when $D \to \infty$.
Therefore, instead of the exact minimum value, we employ the tiny value of float $\epsilon_{\mathrm{float}}$ that is the closest to zero numerically (in the case of float32, $w_{\tilde{\nu}}(w_{mv} = \epsilon_{\mathrm{float}}) \simeq 87.3365$).
In summary, the maximum $w_{\tilde{\nu}}$, $\overline{w}_{\tilde{\nu}}$, can be defined as follows:
\begin{align}
    \overline{w}_{\tilde{\nu}} = \max(\overline{w}_{mv} - \ln(\overline{w}_{mv}), \epsilon_{\mathrm{float}} - \ln(\epsilon_{\mathrm{float}}))
\end{align}

If we design the step size with $\overline{w}_{\tilde{\nu}}$, we can obtain an interpolation update rule for $\tilde{\nu}$ similar to $m$ and $v$.
However, although the minimum value of $\tilde{\nu}$ is expected to be positive, an excessive robustness could inhibit the learning process.
Thus, it is desirable that the user retain a certain level of control on the extent of robustness the algorithm is allowed to achieve.
Therefore, our final step is to transform $\tilde{\nu} = \underline{\tilde{\nu}} + \Delta \tilde{\nu}$ with a user-specified minimum value, $\underline{\tilde{\nu}} > 0$, and the deviation, $\Delta \tilde{\nu} > 0$, automatically controlled by the algorithm.
With this transformation, the appropriate step size $\kappa_{\Delta \tilde{\nu}}$ satisfies the following:
\begin{align}
	\Delta \tilde{\nu}_t &= \Delta \tilde{\nu}_{t-1} + \kappa_{\Delta \tilde{\nu}} g_{\tilde{\nu}} = \Delta \tilde{\nu}_{t-1} + \tau_{\tilde{\nu}} (\lambda'_{t} - \Delta \tilde{\nu}_{t-1})
\end{align}
where the right side of the second equality is obtained from eq.~\eqref{eq:dof_ema} by replacing $\tilde{\nu}$ with $\Delta \tilde{\nu}$ and using $\lambda'_{t} \neq \lambda_{t}$. Substituting $g_{\tilde{\nu}}$ by its expression from eq.~\eqref{eq:grad_nu_m2} and using $\tau_{\tilde{\nu}} = (1 - \beta) w_{\tilde{\nu}}\overline{w}_{\tilde{\nu}}^{-1}$, the following equalities are obtained:
\begin{align}
	&\underbrace{ \kappa_{\Delta \tilde{\nu}} \left[ w_{\tilde{\nu}} \cfrac{d}{2}\Biggl \{ \cfrac{1}{\tilde{\nu}_{t-1} w_{\tilde{\nu}}} \Biggl (\cfrac{\tilde{\nu}_{t-1} + 2}{\tilde{\nu}_{t-1} + 1} + \tilde{\nu}_{t-1} \Biggr ) - 1 \Biggr \} \right] }_{\text{A}} = (1 - \beta) \frac{w_{\tilde{\nu}}}{\overline{w}_{\tilde{\nu}}} (\lambda'_{t} - \Delta \tilde{\nu}_{t-1}) \nonumber \\
	&A = \underbrace{ \kappa_{\Delta \tilde{\nu}} \left[ \frac{d w_{\tilde{\nu}}}{2 \Delta \tilde{\nu}_{t-1}} \right] }_{(1 - \beta) \frac{w_{\tilde{\nu}}}{\overline{w}_{\tilde{\nu}}}} \left( \underbrace{ \cfrac{\Delta \tilde{\nu}_{t-1}}{\tilde{\nu}_{t-1} w_{\tilde{\nu}}} \Biggl (\cfrac{\tilde{\nu}_{t-1} + 2}{\tilde{\nu}_{t-1} + 1} + \tilde{\nu}_{t-1} \Biggr ) }_{\lambda'_{t}} - \Delta \tilde{\nu}_{t-1} \right)
\end{align}
From this, we can derive
\begin{align}
    \kappa_{\Delta \tilde{\nu}} &= 2 \Delta \tilde{\nu}_{t-1} \cfrac{1 - \beta}{d \overline{w}_{\tilde{\nu}}} \\
    \lambda'_{t} &= \Biggl ( \cfrac{\tilde{\nu}_{t-1} + 2}{\tilde{\nu}_{t-1} + 1} + \tilde{\nu}_{t-1} \Biggr )
    \cfrac{\tilde{\nu}_{t-1} - \underline{\tilde{\nu}}}{\tilde{\nu}_{t-1} w_{\tilde{\nu}}}
\end{align}
along with the update rule for $\Delta \tilde{\nu}$:
\begin{align}
    \Delta \tilde{\nu}_t &= \Delta \tilde{\nu}_{t-1} + \kappa_{\Delta \tilde{\nu}} g_{\tilde{\nu}}
    \nonumber \\
    &= (1 - \tau_{\tilde{\nu}}) \Delta \tilde{\nu}_{t-1}
    + \tau_{\tilde{\nu}} \Biggl ( \cfrac{\tilde{\nu}_{t-1} + 2}{\tilde{\nu}_{t-1} + 1} + \tilde{\nu}_{t-1} \Biggr )
    \cfrac{\tilde{\nu}_{t-1} - \underline{\tilde{\nu}}}{\tilde{\nu}_{t-1} w_{\tilde{\nu}}}
    \nonumber
\end{align}
Finally, by adding $\underline{\tilde{\nu}} + \epsilon$ to both sides, we can directly obtain the update rule for $\tilde{\nu}$.
\begin{align}
    \tau_{\tilde{\nu}} &= (1 - \beta) w_{\tilde{\nu}}\overline{w}_{\tilde{\nu}}^{-1} \\
    \lambda_{t} &= \lambda'_{t} + \underline{\tilde{\nu}} + \epsilon
    \\
    \tilde{\nu}_{t} &= (1 - \tau_{\tilde{\nu}}) \tilde{\nu}_{t-1} + \tau_{\tilde{\nu}} \lambda_{t}
\end{align}
Note that the minimum value of $\Delta \tilde{\nu} = \tilde{\nu} - \underline{\tilde{\nu}}$ is given by $\epsilon$, such that $\Delta \tilde{\nu} > 0$ is always satisfied.
This process also prevents $\lambda$ from becoming $0$ and stopping the update of $\tilde{\nu}$.

\subsection{Algorithm}

Finally, the update amount of the optimization parameters $\theta$ for AdaTerm is expressed as follows:
\begin{align}
    \eta^\mathrm{AdaTerm}(g_t) = \cfrac{m_t (1 - \beta^t)^{-1}}{\sqrt{v_t (1 - \beta^t)^{-1}}}
    \label{eq:adaterm_update_rule}
\end{align}
Unlike Adam~\cite{kingma2014adam}, the small amount usually added to the denominator, $\epsilon$, is removed, since $\sqrt{v} \geq \epsilon$ in our implementation.

The pseudo-code of AdaTerm\footnote{AdaTerm's t-momentum requires scale estimation and would incur a higher cost if integrated into the pure SGD algorithm. Therefore, we omit it in this study and only consider a variant of the Adam algorithm.} is summarized in Alg.~\ref{alg:adaterm}.
The regret bound is also analyzed based on a novel approach different from the literature~\cite{reddi2019convergence} in~\ref{apdx:proofs}, through combination of the approach proposed by \cite{alacaoglu2020new}, with a novel method based on the Lemma~\ref{lem:pp_ineq}, to eliminate the AMSGrad assumption from the regret analysis.

\begin{algorithm}[tb]
    \caption{AdaTerm optimizer}
    \label{alg:adaterm}
    \begin{algorithmic}[1]
        \STATE{Set $\alpha > 0$ ($10^{-3}$ is the default value)}
        \STATE{Set $\beta \in (0, 1)$ ($0.9$ is the default value)}
        \STATE{Set $\epsilon \ll 1$ ($10^{-5}$ is the default value)}
        \STATE{Set $\underline{\tilde{\nu}} > 0$ ($1$ is the default value)}
        \STATE{Set $d$ as the dimension size of each subset of parameters}
        \STATE{Initialize $\mathcal{\tilde{D}}$, $\theta_1$, $m_0 \gets 0$, $v_0 \gets \epsilon^2$, $\tilde{\nu}_0 \gets \underline{\tilde{\nu}} + \epsilon$, $t \gets 0$}
        \WHILE{$\theta_t$ not converged}
            \STATE{// Compute gradient}
            \STATE{$t \gets t + 1$}
            \STATE{$g_t = \nabla_{\theta_{t}} \mathcal{L}_{\mathcal{B}_t}$, $\mathcal{B}_t \sim \mathcal{\tilde{D}}$}
            \STATE{// Compute index of outlier}
            \STATE{$s = (g_t - m_{t-1})^2$}
            \STATE{$D = d^{-1} s^\top v_{t-1}^{-1}$}
            \STATE{// Compute adaptive step sizes}
            \STATE{$w_{mv} = (\tilde{\nu}_{t-1} + 1)(\tilde{\nu}_{t-1} + D)^{-1}$}
            \STATE{$\overline{w}_{mv} = (\tilde{\nu}_{t-1} + 1)\tilde{\nu}_{t-1}^{-1}$}
            \STATE{$w_{\tilde{\nu}} = w_{mv} - \ln(w_{mv})$}
            \STATE{$\overline{w}_{\tilde{\nu}} = \max(\overline{w}_{mv} - \ln(\overline{w}_{mv}), \epsilon_{\mathrm{float}} - \ln(\epsilon_{\mathrm{float}}))$}
            \STATE{$\tau_{mv} = (1 - \beta) \cfrac{w_{mv}}{\overline{w}_{mv}}$}
            , {$\tau_{\tilde{\nu}} = (1 - \beta) \cfrac{w_{\tilde{\nu}}}{\overline{w}_{\tilde{\nu}}}$}
            \STATE{// Compute update amounts}
            \STATE{$\Delta s = \max(\epsilon^2, (s - D v_{t-1})\tilde{\nu}_{t-1}^{-1})$}
            \STATE{$\lambda = \left ( \cfrac{\tilde{\nu}_{t-1} + 2}{\tilde{\nu}_{t-1} + 1} + \tilde{\nu}_{t-1} \right ) \cfrac{\tilde{\nu}_{t-1} - \underline{\tilde{\nu}}}{\tilde{\nu}_{t-1} w_{\tilde{\nu}}} + \underline{\tilde{\nu}} + \epsilon$}
            \STATE{// Update parameters}
            \STATE{$m_t = (1 - \tau_{mv}) m_{t-1} + \tau_{mv} g_t$}
            \STATE{$v_t = (1 - \tau_{mv}) v_{t-1} + \tau_{mv} (s + \Delta s)$}
            \STATE{$\tilde{\nu}_t = (1 - \tau_{\tilde{\nu}}) \tilde{\nu}_{t-1} + \tau_{\tilde{\nu}} \lambda$}
            \STATE{$\theta_{t+1} = \theta_{t} - \alpha \cfrac{m_t (1 - \beta^t)^{-1}}{\sqrt{v_t (1 - \beta^t)^{-1}}}$}
        \ENDWHILE
    \end{algorithmic}
\end{algorithm}

\subsection{Behavior analysis}

\subsubsection{Convergence analysis}

Our convergence proof adopts the approach highlighted in~\cite{alacaoglu2020new}.
As such, we start by enunciating the same assumptions:
\begin{assumption} \label{asmp:necessary_assumptions}
Necessary assumptions:
\begin{enumerate}
\item $\mathcal{F} \subset \mathbb{R}^d$ is a compact convex set
\item $\mathcal{L}_{\mathcal{B}_t}: \mathcal{F} \to \mathbb{R}$ is a convex lower semicontinuous (lsc) function
\item $\mathcal{F}$ has a bounded diameter, i.e. $D = \underset{x, y \in \mathcal{F}}{\max} \norm{x - y}_{\infty}$, and $G = \underset{t \in [T]}{\max} \norm{g_{t}}_{\infty}$
\end{enumerate}
\end{assumption}

Following the convergence result, the proof of which can be found in~\ref{apdx:proofs}, the following holds true:
\begin{theorem} \label{th:convergence_bound}
Let $\tau_{t}$ be the value of $\tau_{mv}$ at time step $t$ and let $\underline{\tau} \leq \tau_{t}$, $\forall t$.
Under Assumption~\ref{asmp:necessary_assumptions}, and with $\beta < 1$, $\alpha_{t} = \alpha/\sqrt{t}$, AdaTerm achieves a regret $R_{T} = \sum\limits_{t=1}^{T} \mathcal{L}_{\mathcal{B}_t}(\theta_t) - \mathcal{L}_{\mathcal{B}_t}(\theta^*)$, such that
\begin{align}
\begin{split}
    &R_{T} \leq \left. \frac{D^2\sqrt{T}}{4\tau_{T}\alpha} \sum\limits_{i=1}^{d} v_{T,i}^{1/2} + \left[ \frac{\underline{\tau}^2 + 1 - (\beta + \underline{\tau})}{2\underline{\tau}^2} \right] \sum\limits_{t=1}^{T-1} \frac{D^2}{\alpha_{t}} \sum\limits_{i=1}^{d} v_{t,i}^{1/2} \right. \\
    &\left.\; + \left[ \frac{(1-\beta)^2 \alpha}{\epsilon\underline{\tau}^2\sqrt{T}}  \right] \sum\limits_{t=1}^{T-1} \sum\limits_{i=1}^{d} (1 - \underline{\tau})^{T-k} g^2_{k,i} \right. \\
    &\;+ \left[ \frac{\underline{\tau} (1 - \underline{\tau}) + (1-\beta)}{2\underline{\tau}^2} \right]\left[ \frac{(1-\beta)^2 \alpha}{\epsilon\underline{\tau}^2} \right] \sqrt{1 + \ln (T-1)} \sum\limits_{i=1}^d \norm{g_{1:T-1,i}^2}_2
\end{split}
\end{align}
\end{theorem}

\begin{corollary}[Non-robust case regret bound] \label{corol:non_robust_convergence_bound}
If $\underline{\tilde{\nu}} \to \infty$, then $\underline{\tau} \to \mathrm{constant} = (1-\beta)$. Then, the regret becomes:
\begin{align}
\begin{split}
    R_{T} &\leq \left. \frac{D^2\sqrt{T}}{4(1-\beta)\alpha} \sum\limits_{i=1}^{d} v_{T,i}^{1/2} + \frac{1}{2} \sum\limits_{t=1}^{T-1} \frac{D^2}{\alpha_{t}} \sum\limits_{i=1}^{d} v_{t,i}^{1/2} \right. \\
    &\left.\; + \frac{\alpha}{\epsilon\sqrt{T}} \sum\limits_{t=1}^{T-1} \sum\limits_{i=1}^{d} \beta^{T-k} g^2_{k,i} \right. \\
    &\;+ \frac{(1+\beta)\alpha\sqrt{1 + \ln (T-1)}}{2(1-\beta)\epsilon} \sum\limits_{i=1}^d \norm{g_{1:T-1,i}^2}_2
\end{split}
\end{align}
\end{corollary}

Note the similarity between this regret bound and the one derived by~\cite{reddi2019convergence} and by~\cite{zhuang2020adabelief} using AMSGrad. In particular, this regret can be bounded by $\mathcal{O}(G\sqrt{T})$, and thus, the regret is upper-bounded by a minimum of $\mathcal{O}(G\sqrt{T})$. This leads to the worst-case dependence of the regret on $T$ to remain $\mathcal{O}(\sqrt{T})$, although AMSGrad is not used.

We emphasize that this approach to the regret bound is not specific to AdaTerm but can be used to bound the regret of other moment-based optimizers, including Adam and AdaBelief.

\subsubsection{Qualitative robustness behavior analysis} \label{ssec:intuitive_behavior}

In AdaTerm, the robustness to noise is qualitatively expressed differently from the previous studies~\cite{ilboudo2020robust,ilboudo2021adaptive}, owing to the following two factors.

First, $v$ is now updated robustly depending on $w_{mv}$, ensuring that the observation of aberrant gradients $g$ would not cause a sudden increase in the scale parameter and inadvertently loosen the threshold for other aberrant update directions in the next and subsequent steps.
In addition, the update of the scale $v$ is expected to self-coordinate among the axes, by virtue of the inter-dependency between all of the parameters when estimating the scale.

Indeed, in a diagonal Gaussian distribution, the scale for each axis can and is estimated independently; however, in a diagonal Student's t-distribution, each entry of the scale also depends on the previous value of the other entries, leading to a coordinated update of the scales.
This inter-dependency exists thanks to the Mahalanobis distance metric $D$ and its intervention in the weight computation $w_{mv}$.

\begin{figure}[tb]
    \centering
    \includegraphics[keepaspectratio=true,width=0.95\linewidth]{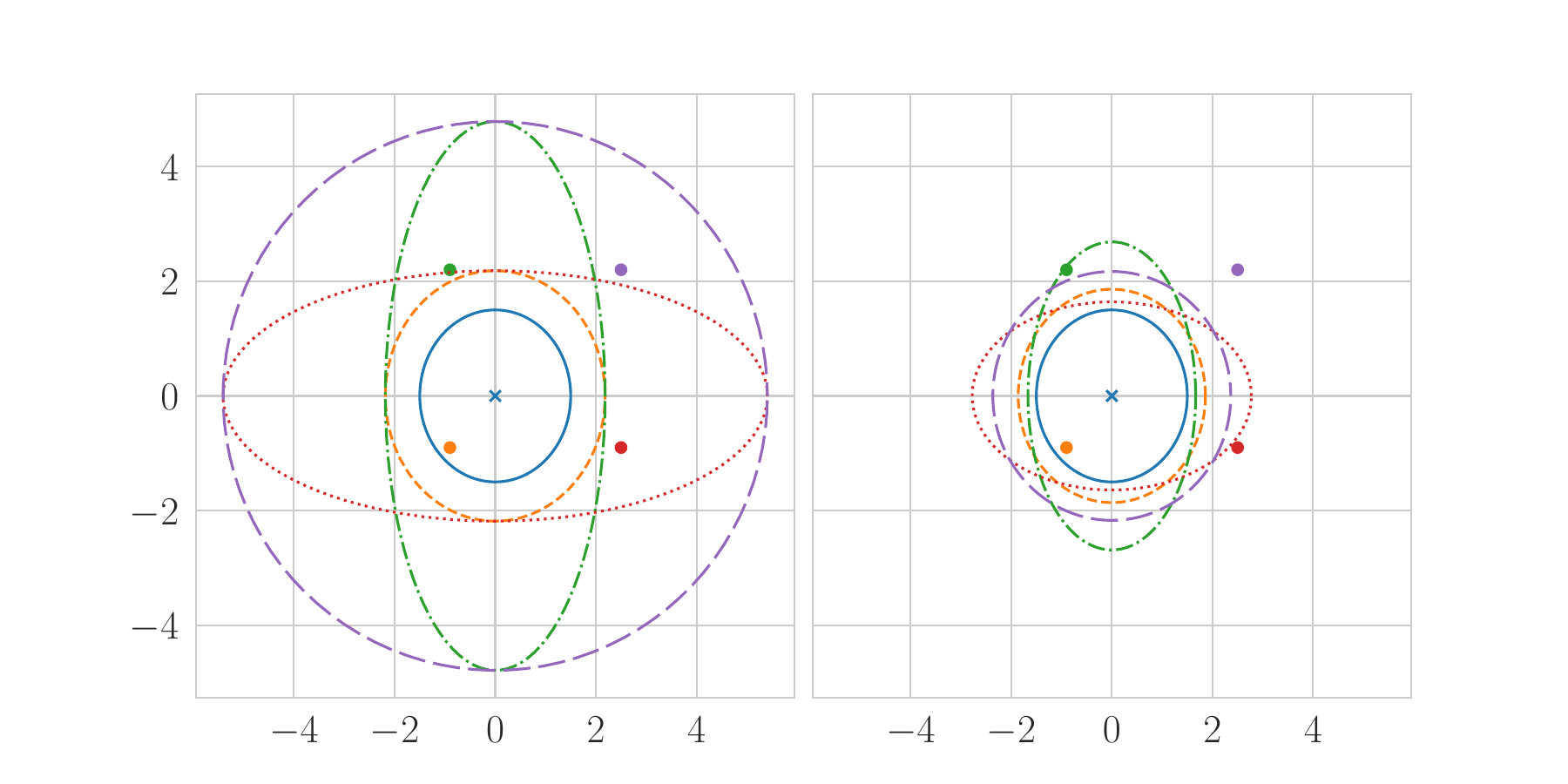}
    \caption{Interdependence of the scale entries. The ellipses represent $3\sigma$ and the blue color is the initial $\sigma$ value. The other colors correspond to how the $\sigma$ would change under the algorithm when the associated point with the same color is observed.}
    \label{fig:adaterm_scale_viz}
\end{figure}

Specifically, as illustrated in Fig.~\ref{fig:adaterm_scale_viz}, if an anomaly is detected only along a particular axis (e.g., points red and green in Fig.~\ref{fig:adaterm_scale_viz}), the corresponding entry of the vector $\Delta s$ becomes larger, causing the associated entry of $v$ to increase.
This in turn would mitigate the anomaly detection of the next step (i.e., a gradient far from the current location estimate $m$ would be attributed a higher weight $w_{mv}$, since the squared Mahalanobis distance $D \propto s^\top v^{-1}$ will be overall smaller).
Conversely, if anomalies are detected on most of the axes (e.g., violet point in Fig.~\ref{fig:adaterm_scale_viz}), $\Delta s$ will become smaller ($\Delta s \simeq \epsilon^2$) and either prevent any further increase in $v$ or cause it to decrease.
This would prompt the algorithm to treat a larger number of the further gradients as anomalies ($D$ becomes overall larger, implying $w_{mv}$ is smaller) as described above.
Such adaptive and coordinated behavior would yield stable updates even if $\beta$ is smaller than the conventional $\beta_2$ (set to be $= 0.999$ in most optimizers).

\begin{figure}[tb]
    \centering
    \includegraphics[keepaspectratio=true,width=0.95\linewidth]{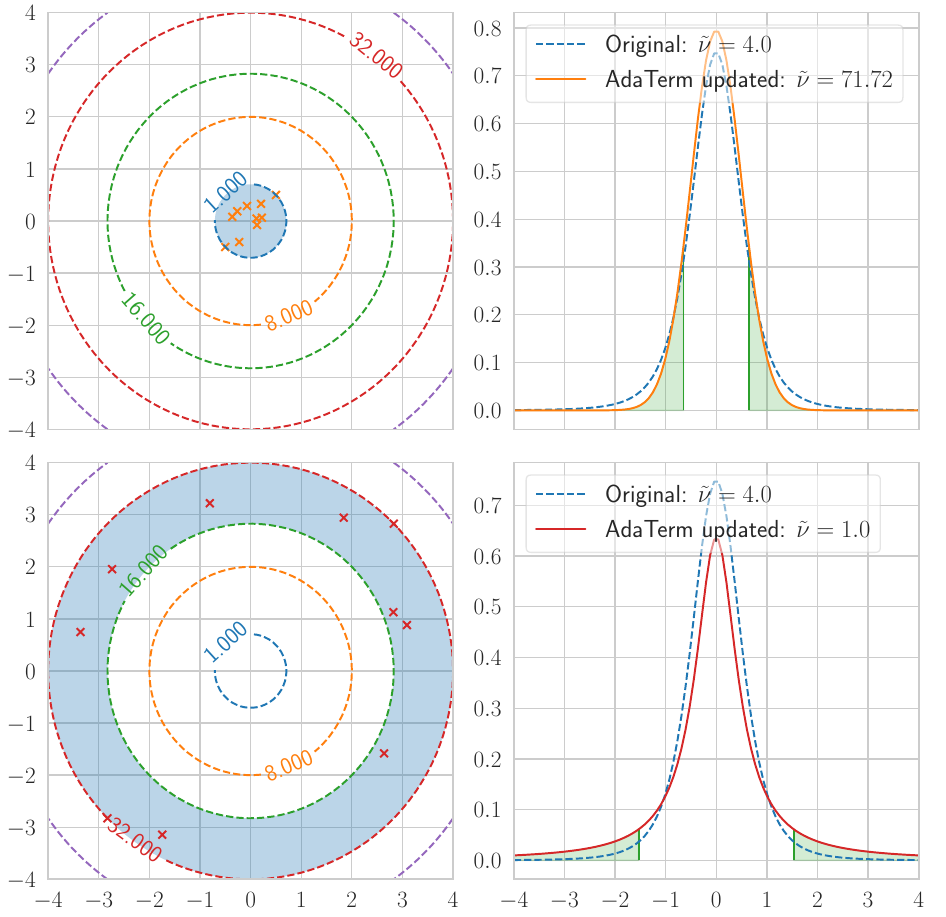}
    \caption{Change in the robustness parameter $\nu$ based on the Mahalanobis distance $D$ (represented by the contour plot) of the observations. The green vertical lines represent the 90\% percentile. As illustrated, when $D$ is small, the degree of freedom increases. When $D$ is large, it decreases, and the tail of the distribution becomes heavier to account for the outliers.}
    \label{fig:adaterm_dof_viz}
\end{figure}

Secondly, as illustrated in Fig.~\ref{fig:adaterm_dof_viz}, the robustness indicator $\tilde{\nu}$ is increased when the deviation $D$ is small (i.e., when the observed gradient is not an aberrant value).
Indeed, in this case, $\lambda'_{t}$ becomes larger and $\kappa_{\Delta \tilde{\nu}} g_{\tilde{\nu}}$ becomes positive.
However, the increased speed will be limited by $\overline{w}_{\tilde{\nu}}$, owing to its inclusion in the step size $\kappa_{\Delta \tilde{\nu}}$.
In contrast, if an aberrant value is observed, $w_{\tilde{\nu}} \gg 1$ and $\lambda'_{t}$ decreases, such that $\kappa_{\Delta \tilde{\nu}} g_{\tilde{\nu}}$ becomes negative.

An intuitive visualization of the described behavior obtained by the AdaTerm equations for $\tau_{mv}$ and $\kappa_{\Delta \tilde{\nu}} g_{\tilde{\nu}}$ can also be found in~\ref{apdx:adaterm_algo_vis}.

\begin{figure}[tb]
    \centering
    \includegraphics[keepaspectratio=true,width=0.97\linewidth]{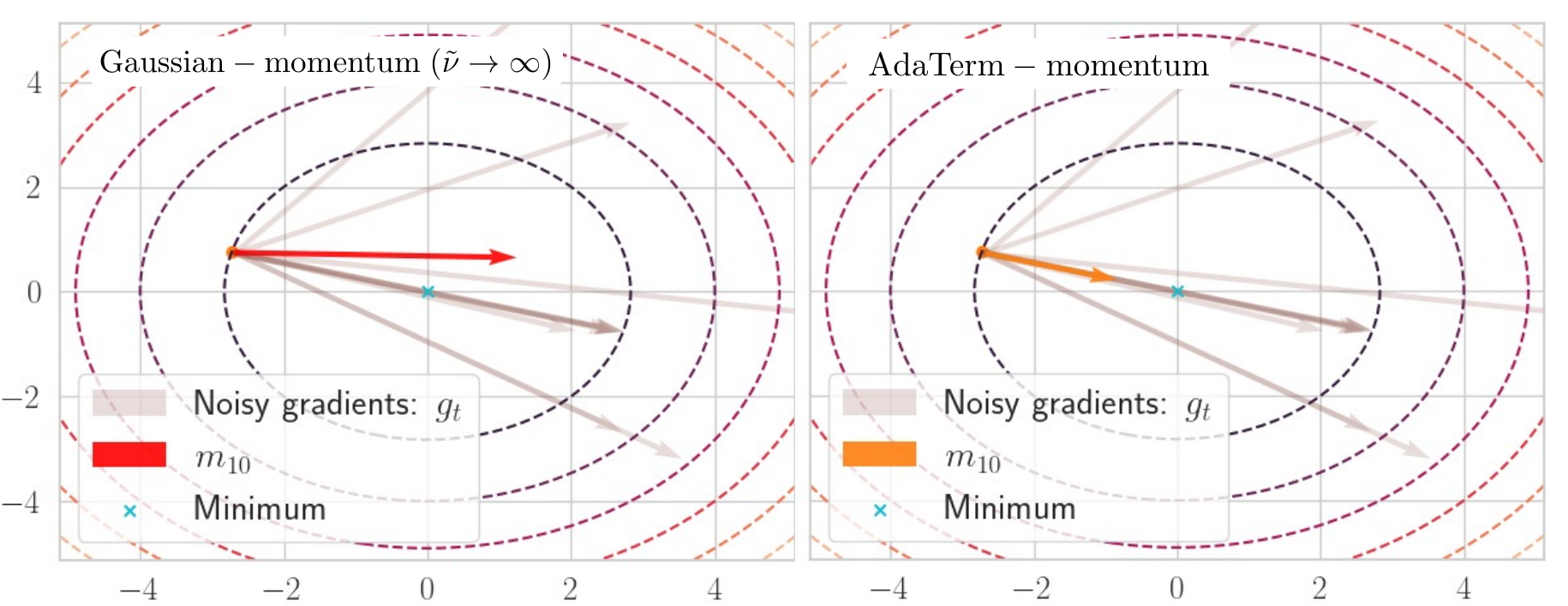}
    \caption{Illustration of the first-order moment acquired by AdaTerm and Gaussian ($\tilde{\nu} \to \infty$) strategies after 10 steps (with $\beta=0.9$) based on step-by-step noisy gradients observed on the sphere function $f(x, y) = x^2 + y^2$.}
    \label{fig:ag_m}
\end{figure}

Overall, although the robustness-tuning mechanism replaces the MLL gradient by the upper bounds, it still behaves conservatively and can be expected to retain its excellent robustness to noise, as illustrated in Fig.~\ref{fig:ag_m}.

As a final remark, if $\underline{\tilde{\nu}} \to \infty$, the robustness is lost by design, and AdaTerm essentially matches a slightly different version of AdaBelief exhibiting performance difference as a result of the simplification of the hyper-parameters ($\beta_1=\beta_2=\beta$) and the variance computation mechanism (AdaBelief estimates $\mathbb{E}[(g_t - m_t)^2]$, whereas AdaTerm estimates $\mathbb{E}[(g_t - m_{t-1})^2]$, which is the usual variance estimator).
Therefore, in problems where AdaBelief would be effective, AdaTerm would perform effectively as well.

\section{Simulation benchmarks}
\label{sec:simulations}

Before solving more practical problems, we analyze the behavior of AdaTerm through minimization of typical test functions, focusing mainly on the adaptiveness of the robustness parameter and the convergence profile.

To evaluate its adaptiveness, in all simulations and experiments, the initial $\nu$ in At-Adam is set to be the same value used in t-Adam. Furthermore, although many SGD-based optimization algorithms are available, in this study, we focus on comparing AdaTerm against the two related works (t-momentum-based Adam - t-Adam ~\cite{ilboudo2020robust} - and At-momentum-based Adam - At-Adam~\cite{ilboudo2021adaptive} -)\footnote{Employed implementation can be found at https://github.com/Mahoumaru/t-momentum} and include only the results from Adam~\cite{kingma2014adam}, AdaBelief~\cite{zhuang2020adabelief}, and RAdam~\cite{gulcehre2017robust} as references for non-robust optimization.

\subsection{Analysis based on test functions}

\begin{figure}[tb]
    \centering
    \includegraphics[keepaspectratio=true,width=0.95\linewidth]{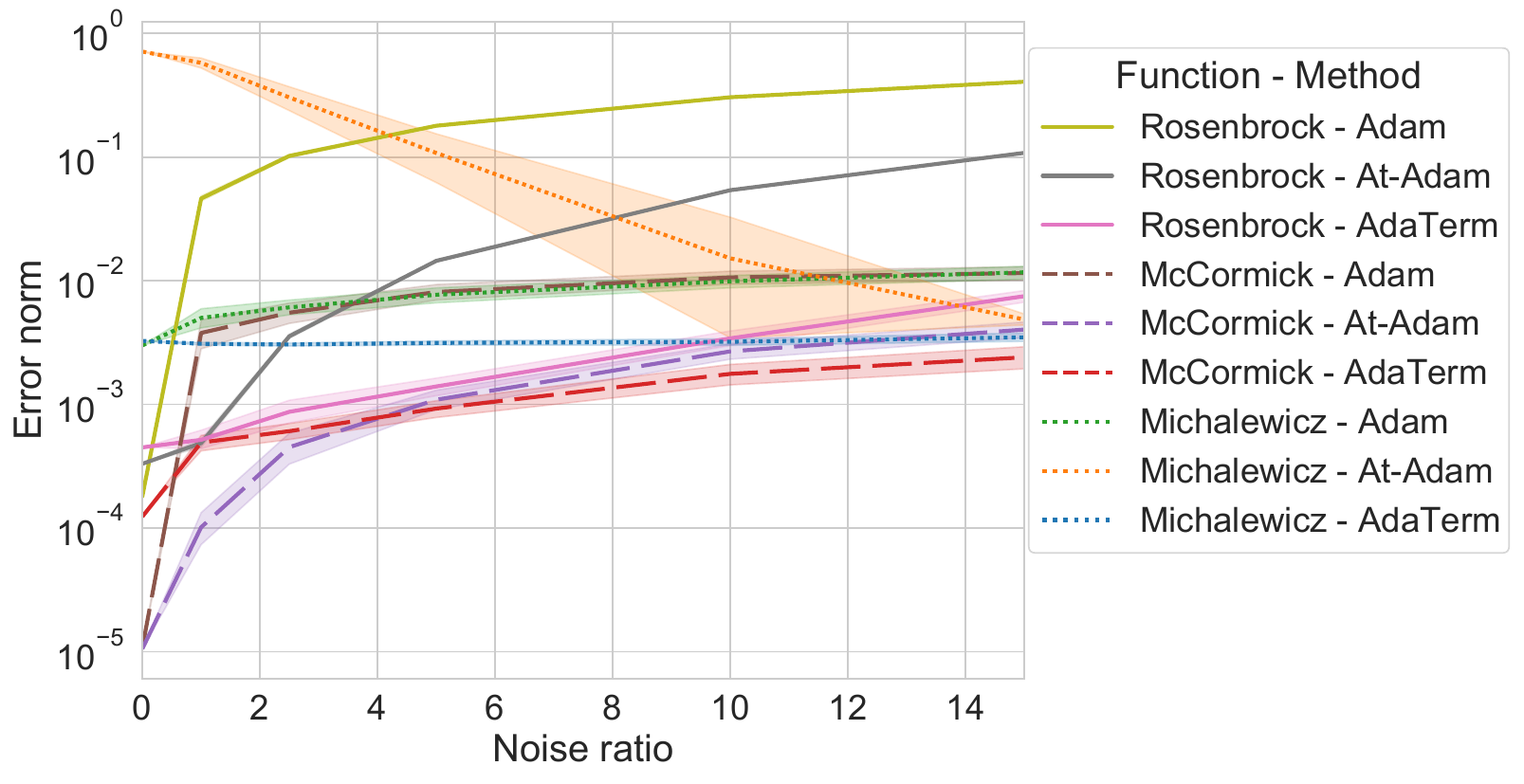}
    \caption{Noise ratio vs. L2 norm between the converged point and the analytically-optimal point}
    \label{fig:comp_norm}
\end{figure}

\begin{figure}[tb]
    \centering
    \includegraphics[keepaspectratio=true,width=0.95\linewidth]{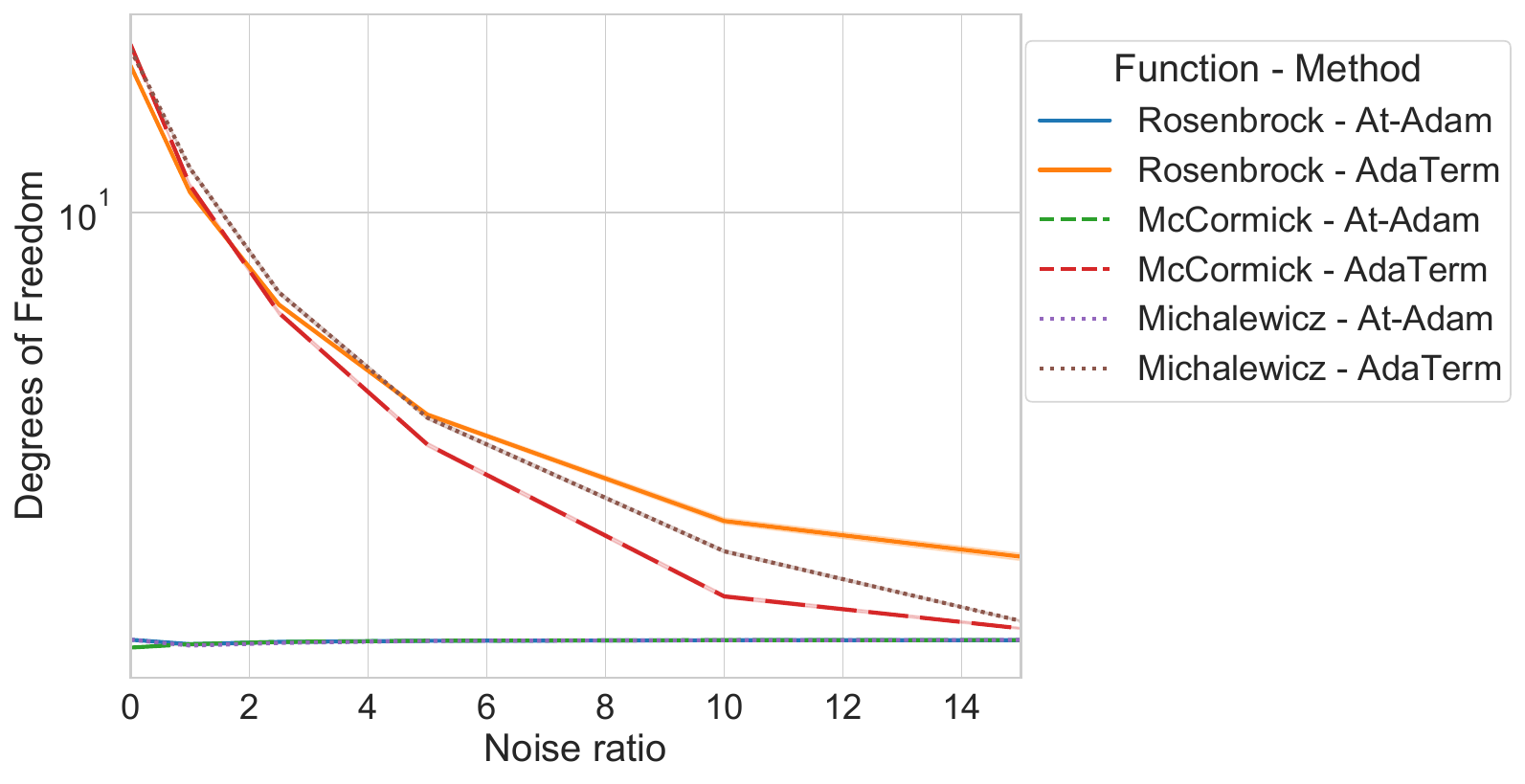}
    \caption{Noise ratio vs. degrees-of-freedom $\nu$}
    \label{fig:comp_dof}
\end{figure}

In this task, our aim is to determine the minimum point of a two-dimensional potential field (benchmark function) by relying on the field gradients.
To analyze the robustness of the optimization algorithms, a uniformly distributed noise ($\in (-0.1, 0.1)$) is added at a specified ratio to the point coordinates. 
The results are summarized in Figs.~\ref{fig:comp_norm} and~\ref{fig:comp_dof} (details are in~\ref{apdx:test}).

Fig.~\ref{fig:comp_norm} shows the error norm from the analytically-optimal point, revealing that AdaTerm's convergence accuracy on the McCormick function was not effective in the lower noise ratio, probably due to the large learning rate.
However, on the other fields, AdaTerm was able to maintain the standard update performance while mitigating the adverse effects of the added noise.

The performance of Adam~\cite{kingma2014adam} significantly degraded when noise was introduced, indicating its sensitivity to noise.
On the Michalewicz function, At-Adam~\cite{ilboudo2021adaptive} attributes the steep gradients near the optimal value to noise and tends to exclude them; thus, the optimal solution is not obtained.
This result implies that the automatic mechanism for tuning $\nu$ employed by At-Adam has insufficient robustness adaptability to noise.

Indeed, the final $\nu$ plotted in Fig.~\ref{fig:comp_dof} shows that the robustness parameter $\nu$ converges to a nearly constant value with At-Adam, independently of the noise ratio.
In contrast, in AdaTerm, the degrees-of-freedom $\nu$ is inversely proportional to the noise ratio; thus, the typical behavior of increase in $\nu$ under minimal noise and decrease under the condition of a high noise ratio is observed.

\subsection{Robustness on regression task}

\paragraph{Problem settings}

Following the same process as in~\cite{ilboudo2020robust}, we consider the problem of fitting a ground truth function $f(x) = x^2 + \ln(x+1) + \sin(2\pi x) \cos(2\pi x)$ given noisy observations $y = f(x) + \zeta$ with $\zeta$ expressed as:
\begin{align}
    \zeta &\sim \mathcal{T}(1, 0, 0.05) \mathrm{Bern} \left ( \frac{p}{100} \right ), \ p = 0, 10, 20, \ldots, 100
\end{align}
where $\mathcal{T}(1, 0, 0.05)$ designates a Student's t-distribution with degrees-of-freedom $\nu_{\zeta} = 1$, location $\mu_{\zeta} = 0$, and scale $v_{\zeta} = 0.05$; $\mathrm{Bern}(p/100)$ is a Bernoulli distribution with the ratio $p$ as its parameter.
$50$ trials with different random seeds are conducted for each noise ratio $p$ and each optimization method, using $40000$ $(x, y)$ pairs sampled as observations and split into batches of size $10$. This batch size is chosen arbitrarily as being neither exceedingly large nor exceedingly small with respect to the full dataset size.

The model used is a fully connected neural network with $5$ linear layers, each composed of $50$ neurons, equipped with an ReLU activation function~\cite{relu} for all the hidden layers.
The training and the test loss functions are the mean squared error (MSE) applied on $(\hat{y}, y)$ and $(\hat{y}, f(x))$, respectively, where $\hat{y}$ is the network's estimate given $x$.

\paragraph{Result}

\begin{figure}[tb]
    \centering
    \includegraphics[keepaspectratio=true,width=0.95\linewidth]{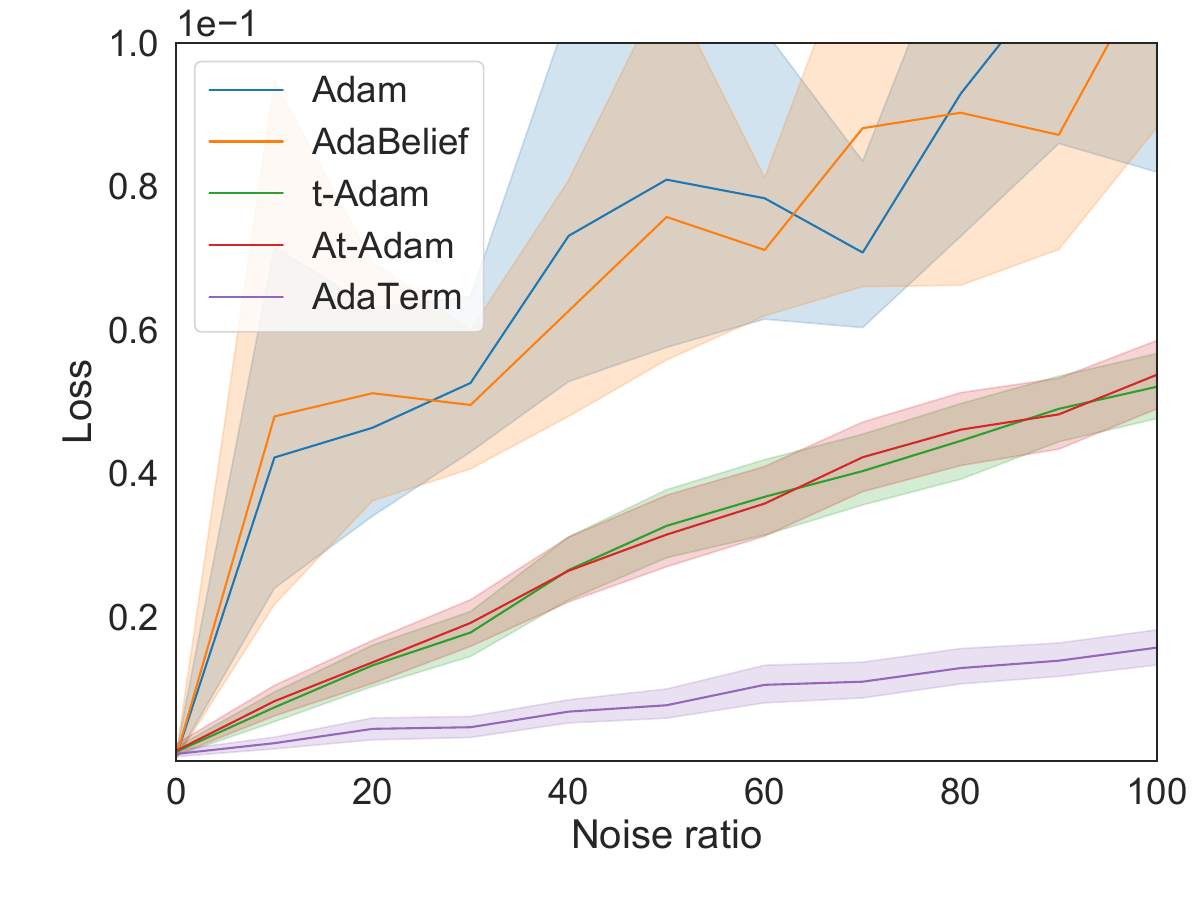}
    \caption{Test loss on the Regression task against the noise ratio $p$}
    \label{fig:result_regr}
\end{figure}

The test loss against the noise ratio $p$ is plotted in Fig.~\ref{fig:result_regr}.
As can be observed, AdaTerm performed even more robustly than t-Adam and At-Adam and was able to maintain its prediction loss almost constant across all values of noise ratio.
The effect of batch size on the learning performance is also studied and can be found in~\ref{apdx:batch_size_effect}.

\section{Experiments}
\label{sec:experiments}

\subsection{Configurations of practical problems}

After verifying the optimization efficiency and robustness on certain test functions and a regression task with artificial noise, we now investigate four practical benchmark problems to evaluate the performance of the proposed algorithm, AdaTerm.
Details of the setups (e.g., network architectures) can be found in~\ref{apdx:learning}.

For each method, 24 trials are conducted with different random seeds, and the mean and standard deviation of the scores for the respective benchmarks are evaluated.
All the test results after learning can be found in Table~\ref{tab:exp_result}.
An ablation test was also conducted for AdaTerm, as summarized in~\ref{apdx:ablation}.

\subsubsection{Classification of mislabeled CIFAR-100}

The first problem is an image classification problem with the CIFAR-100 dataset.
As artificial noise, a proportion of the labels (0~\% and 10~\%) of the training data was randomized to differ from the true labels.
As a simple data augmentation during the training process, random horizontal flip is introduced along with four paddings.

The loss function is the cross-entropy.
To stabilize the learning performance, we utilize the label smoothing technique~\cite{szegedy2016rethinking}, a regularization tool
for deep neural networks. It involves the generation of ``soft'' labels by
applying a weighted average between a uniform distribution
and the ``hard'' labels.
In our experiments, the degree of smoothing is set to 20\%, in reference to the literature~\cite{lukasik2020does}.

\subsubsection{Long-term prediction of robot locomotion}

The second problem is a motion-prediction problem.
The dataset used was sourced from the literature~\cite{kobayashi2020q}. It contains 150 trajectories as training data, 25 trajectories as validation data, and 25 trajectories as test data, all collected from a hexapod locomotion task with 18 observed joint angles and 18 reference joint angles.
An agent continually predicts the states within the given time interval (1 and 30 steps) from the initial true state and the subsequent predicted states.
Therefore, the predicted states used as inputs deviate from the true states and become noise in the learning process.

For the loss function, instead of the commonly used mean squared error (MSE), we employ the negative log likelihood (NLL) of the predictive distribution, allowing the scale difference of each state to be considered internally.
The NLL is computed at each prediction step, and their sum is used as the loss function.
Because of the high cost of back-propagation through time (BPTT) over the entire trajectory, truncated BPTT (30 steps on average)~\cite{puskorius1994truncated,tallec2017unbiasing} is employed.

\subsubsection{Reinforcement learning on Pybullet simulator}

The third problem is RL performed on the Pybullet simulation engine using OpenAI Gym~\cite{brockman2016openai,coumans2016pybullet} environments.
The tasks to be solved are Hopper and Ant, both of which require an agent to move as straight as possible on a flat terrain.
As mentioned before, RL relies on estimated values in behalf of the absence of true target signals. This can easily introduce noise in the training.

The implemented RL algorithm is an actor-critic algorithm based on the literature~\cite{kobayashi2023proximal}.
The agent only learns after each episode using an experience replay.
This experience replay samples 128 batches after each episode, and its buffer size is set to be sufficiently small (at 10,000) to reveal the influence of noise.

\subsubsection{Policy distillation by behavioral cloning}

The fourth problem is policy distillation~\cite{rusu2015policy}; a method used to replicate the policy of a given RL agent by training a smaller and computationally more efficient network to perform at the expert level.
Three policies for Ant properly trained on the RL problem above were considered as experts, and 10 trajectories of state-action pairs were collected for each policy.
We also collected, as non-expert (or amateur) demonstrations, three trajectories from one policy that fell into a local solution.
In the dataset constructed with these trajectories, not only the amateur trajectories but also the expert trajectories could be non-optimal in certain situations, thereby leading to noise.

The loss function is the negative log likelihood of the policy according to behavioral cloning~\cite{bain1995framework}, which is the simplest imitation learning method.
In addition, a weight decay is added to prevent the small networks (i.e., the distillation target) from over-fitting particular behaviors.

\subsection{Analysis of results of practical problems}
\label{sec:exp_analysis}

\begin{table*}[tb]
    \caption{Experimental results:
        the best in all the results for each benchmark is written in bold;
        The numbers in parentheses denote standard deviations. We can see that AdaTerm relaxes the requirement of manual tuning of the robustness of the algorithm for every single task and interpolates between or prevails over the performance of non-robust and robust optimizers when facing practical problems (see section~\ref{sec:exp_analysis} for the detailed analysis).
    }
    \label{tab:exp_result}
    \centering
    \begin{tabular}{l | cc cc | cc cc}
        \hline\hline
         & \multicolumn{2}{c}{Classification}
         & \multicolumn{2}{c|}{Distillation}
         & \multicolumn{2}{c}{Prediction}
         & \multicolumn{2}{c}{RL}
         \\
         & \multicolumn{2}{c}{Accuracy}
         & \multicolumn{2}{c|}{The sum of rewards}
         & \multicolumn{2}{c}{MSE at final prediction}
         & \multicolumn{2}{c}{The sum of rewards}
         \\
        Methods & 0~\% & 10~\% & w/o amateur & w/ amateur & 1 step & 30 steps & Hopper & Ant
        \\
        \hline\hline
        Adam
        & 0.7205
        & 0.6811
        & 1686.94
        & 1401.01
        & 0.0321
        & 1.1591
        & 1539.06
        & 863.31
        \\

        & (4.33e-3)
        & (4.93e-3)
        & (2.34e+2)
        & (2.39e+2)
        & (4.48e-4)
        & (1.04e-1)
        & (5.64e+2)
        & (4.01e+2)
        \\
        \hline
        AdaBelief
        & 0.7227
        & 0.6799
        & 1731.51
        & 1344.23
        & 0.0320
        & 1.2410
        & 1152.10
        & 692.68
        \\

        & (5.56e-3)
        & (3.94e-3)
        & (1.88e+2)
        & (2.75e+2)
        & (5.67e-4)
        & (1.34e-1)
        & (5.48e+2)
        & (1.17e+2)
        \\
        \hline
        RAdam
        & 0.7192
        & 0.6797
        & 1574.13
        & 1258.10
        & 0.0328
        & 1.1218
        & 1373.85
        & 871.23
        \\

        & (4.28e-3)
        & (3.41e-3)
        & (2.86e+2)
        & (2.70e+2)
        & (3.55e-4)
        & (1.36e-1)
        & (7.55e+2)
        & (3.57e+2)
        \\
        \hline\hline
        t-Adam
        & 0.7174
        & 0.6803
        & 1586.99
        & 1347.29
        & 0.0320
        & 1.1048
        & \textcolor{orange}{\textbf{1634.20}}
        & \textcolor{orange}{\textbf{2272.58}}
        \\

        & (5.67e-3)
        & (3.10e-3)
        & (2.06e+2)
        & (2.11e+2)
        & (4.09e-4)
        & (2.66e-1)
        & (4.52e+2)
        & (3.20e+2)
        \\
        \hline
        At-Adam
        & 0.7178
        & 0.6811
        & 1611.23
        & 1347.51
        & \textcolor{orange}{\textbf{0.0319}}
        & 1.1075
        & 1523.37
        & 2213.75
        \\

        & (4.21e-3)
        & (4.03e-3)
        & (2.25e+2)
        & (2.67e+2)
        & (3.14e-4)
        & (1.97e-1)
        & (5.48e+2)
        & (2.31e+2)
        \\
        \hline
        AdaTerm
        & \textcolor{orange}{\textbf{0.7315}}
        & \textcolor{orange}{\textbf{0.6815}}
        & \textcolor{orange}{\textbf{1770.17}}
        & \textcolor{orange}{\textbf{1411.02}}
        & 0.0335
        & \textcolor{orange}{\textbf{1.0016}}
        & 1550.25
        & 2021.37
        \\
        (Ours)
        & (3.66e-3)
        & (4.46e-3)
        & (2.17e+2)
        & (1.92e+2)
        & (3.09e-4)
        & (2.31e-1)
        & (5.88e+2)
        & (3.87e+2)
        \\
        \hline\hline
    \end{tabular}
\end{table*}

To analyze the experimental results, we divide the four tasks into two cases depending on the composition of the dataset and the origin of the inaccuracies:
\begin{itemize}
\item In the first class of problems, the corrupted dataset is such that it can be partitioned into two subsets of accurate and noisy observations, i.e. $\mathcal{\tilde{D}} = \mathcal{D} \cup \mathcal{E}$ where $\mathcal{E} \nsubseteq \mathcal{D}$ is the noisy subset.
\item In the second class, the corruption extends to the entire dataset, either due to a stationary noise distribution or a dynamic noise distribution.
\end{itemize}

\subsubsection{Case 1 Tasks: $\mathcal{\tilde{D}} = \mathcal{D} \cup \mathcal{E}$}

 This first case comprises of the classification and distillation tasks. In the classification task, the subset data (0\% or 10\%) with randomized labels constitutes the corrupted subset $\mathcal{E}$, and in the distillation task, $\mathcal{E}$ corresponds to the amateur demonstrations. This subdivision holds even though what we consider to be the uncorrupted data may contain noise (CIFAR-100 contains general images that are prone to imperfections, and similarly, the RL expert policies are stochastic) because these corruptions are not as problematic as the ones mentioned previously.

 In these types of tasks, adaptiveness is primordial because the proportion of $\mathcal{E}$ is typically unknown in advance.
 As can be seen in the left part of Table~\ref{tab:exp_result}, the non-robust optimizers (Adam, AdaBelief and RAdam) outperform the fixed-robustness t-Adam and the low-adaptive-capability At-Adam when the proportion of $\mathcal{E}$ is small (0\% and w/o amateur). Furthermore, although At-Adam exhibits a low level of adaptiveness, the adaptiveness nevertheless results in a better performance compared to t-Adam.
 In contrast, when the proportion of $\mathcal{E}$ increases (10\% and w/ amateur), the robust optimizers lead in terms of average performance. Notably, At-Adam still outperforms t-Adam, which uses the most robust version of the algorithm with its degrees-of-freedom fixed at $1$.
 Finally, we observe that AdaTerm's adaptive ability allows it to outperform both the other non-robust and robust optimizers under all dataset conditions considered. Its better performance over non-robust optimizers even without corruption can be attributed to the natural imperfections mentioned earlier, implying that a small level of robustness ($\nu \ll \infty$) is still desirable during the optimization process even in the absence of notable corruption.

\subsubsection{Case 2 Tasks: $\mathcal{\tilde{D}} = \mathcal{E}$}

The prediction and RL tasks constitute the second case. Indeed, the estimators and approximators used in these types of tasks are a constant source of noise and corruptions. Although such noise can be suspected to be stronger in the earlier stages of learning, their prevalence may necessitate an optimizer that maintains its a constant robustness.

This preference can be observed on the right side of Table~\ref{tab:exp_result}. Indeed, the robust optimizers lead by performance, achieving the lowest prediction errors in the 1-step and 30-step prediction and the highest scores in the Hopper and Ant environments.

Although the single-step prediction task is a relatively simple supervised learning approach that does not necessitate a high robustness to noise, the performance of AdaTerm is the worst among all algorithms. We postulate that, similar to the error norm of the McCormick function in Fig.~\ref{fig:comp_norm}, the exceedingly high learning rate caused by $\beta < \beta_2$ contributes to the performance degradation. In contrast, in the 30-step prediction task, only AdaTerm succeeds in achieving a MSE of approximately $1$.
As a consequence of the inaccurate estimated inputs, particularly in the earlier stages of learning, this task challenges the ability of the optimization algorithm to ignore aberrant gradient update directions.
However, as learning progresses, the accuracy of the estimated inputs improves, and the noise robustness gradually becomes less critical.

For the RL tasks, although AdaTerm did not exhibit the best performance, we can confirm its usefulness from two facts: (i) it demonstrated the second-best performance in Hopper task, and (ii) Ant task was only successful with noise-robust optimizers.
This particular task highlights a common drawback of adaptive methods compared with methods based on the right assumption (whether or not robustness will be needed in the considered problem). The inferior performance of the non-robust optimizers indicate the prevalence of noise in RL tasks. Therefore, t-Adam unsurprisingly exhibits the best performance in such situation where robustness is critical. t-Adam was followed closely by At-Adam in the Ant task (clearly, from the results of the non-robust algorithms, the most challenging in terms of noisiness) thanks to its low adaptiveness (similarly to Fig.~\ref{fig:comp_dof}), enabling it to maintain its robustness level close to that of At-Adam during the learning process. Regardless, it was outperformed by AdaTerm (and Adam) on the Hopper.

In summary, the experiments show that AdaTerm interpolates between or prevails over the performance of non-robust and robust optimizers when encountering practical and noisy problems. 

\section{Discussion}
\label{sec:discussion}

The above experiments and simulations demonstrated the robustness and adaptability improvement of AdaTerm compared with the related works.
However, we discuss its limitations below.

\subsection{Performance and drawbacks of AdaTerm}

 As can be seen in Fig.~\ref{fig:comp_norm} and on the right part of Table~\ref{tab:exp_result}, there is a drawback of using AdaTerm on a noise-free optimization problems when compared with a non-robust optimizer, such as Adam, and similarly on full-noise problems compared with a fixed-robust optimizer, such as t-Adam.
These results show that in the absence of noise (respectively in the presence of abundant noise), employing an optimization method that assumes the absence of aberrant data points (respectively the presence of worst-case noisy data) from the beginning can afford better results on noise-free (respectively on extremely noisy) problems.
This is to be expected from an adaptive method such as AdaTerm, which inevitably requires a non-zero adaptation time to converge to a non-robust behavior or may relax its robustness during the learning process before hitting an extremely noisy set of gradients necessitating a fixed high robustness.

Nevertheless, the overall results indicate that the drawback or penalty incurred from using the AdaTerm algorithm instead of Adam or AdaBelief in the case of noise-free applications or t-Adam in the case of high-noise applications is not a significant problem when considering its ability to adapt to different unknown noise ratios. This is particularly why algorithms are equipped with adaptive parameters.

When encountering a potentially noise-free problem, AdaTerm allows setting a large initial value for the degrees-of-freedom and during the course of the training process, the algorithm will decide if the dataset is indeed noise-free and adapt in consequence. Similarly, if there is uncertainty pertaining to the noisiness of the task at hand, the practitioner can directly execute the algorithm and let it automatically adapt to the problem.
Such freedom is only possible thanks to the adaptive capability of AdaTerm as clearly displayed in Fig.~\ref{fig:comp_dof}, and the resulting performance is guaranteed to be sub-optimal with respect to the true nature of the problem.

\subsection{Gap between theoretical and experimental convergence analysis}

 Although Theorem~\ref{th:convergence_bound} provides a theoretical upper bound on the regret achieved by AdaTerm; as a typical approach to convergence analysis, it completely eludes the robustness factor introduced by AdaTerm.
 In particular, Corollary~\ref{corol:non_robust_convergence_bound} provides a more effective bound compared with Theorem~\ref{th:convergence_bound} that is in contrast to the practical application (as displayed in the ablation study of~\ref{apdx:ablation}).
 This implies a gap between the theoretical and experimental analyses and is anchored on the fact that the theoretical bound relies on $\underline{\tau}$.

 As a remedy to this shortcoming of the regret analysis, we have employed (in section~\ref{ssec:intuitive_behavior}) an intuitive analysis of the robustness factor based on the behavior of the different components of our algorithm.
 Although this qualitative analysis possesses a weak theoretical convergence value, the experimental analysis against different noise settings shows that AdaTerm is not only robust but also efficient as an optimization algorithm.
 Nevertheless, we acknowledge the requirement for a stronger theoretical analysis that considers the noisiness of the gradients, while allowing for a theoretical comparison between the robustness and efficiency of different optimizers.

\subsection{Normalization of gradient}

By considering the gradients to be generated from Student's t-distribution, AdaTerm normalizes the gradients (more precisely, its first-order moment) using the estimated scale, instead of the second-order moment.
This is similar to the normalization of AdaBelief, as mentioned before.
However, as shown in ~\ref{apdx:variants}, the normalization with the second-order moment can exhibit better performances in certain circumstances.

Since the second-order moment is larger than the scale, the normalization by the second-order moment renders the update conservative, while the one by the scale is expected to break through the stagnation of updates.
While both characteristics are important, the answer to the question ``which one is desirable?" remains situation-dependent.
Therefore, we need to pursue the theory concerning this point and introduce a mechanism to adaptively switch the use of both normalization approach according to the required characteristics.

\section{Conclusion}
\label{sec:conclusion}

In this study, we presented AdaTerm, an optimizer adaptively robust to noise and outliers for deep learning.
AdaTerm models the stochastic gradients generated during training using a multivariate diagonal Student's t-distribution. Accordingly, the distribution parameters are optimized through the surrogated maximum log-likelihood estimation.
Optimization of test functions reveals that AdaTerm can adjust its robustness to noise in accordance with the impact of noise.
Through the four typical benchmarks, we confirmed that the robustness to noise and the learning performance of AdaTerm are comparable to or better than those of conventional optimizers.
In addition, we derived the new regret bound for the Adam-like optimizers without incorporating AMSGrad.

This study focused on computing the moments of the t-distribution; however, in recent years, the importance of integration with raw SGD (i.e., decay of scaling in Adam-like optimizers) has been confirmed~\cite{luo2019adaptive,zhou2020towards}.
Therefore, we will investigate a natural integration of AdaTerm and the raw SGD by reviewing $\Delta s$ that may enable the normalization to be constant.
In addition, as mentioned in the discussion, a new framework for analyzing optimization algorithms both in terms of robustness and efficiency (such that they can be compared)is required.
One such analysis was performed by~\cite{scaman2020robustness} on SGD; however, its extension to momentum-based optimizers and its ability to allow theoretical comparison across different algorithms remain limited.
Therefore, we will seek a similar but better approach in the future, e.g. via trajectory analysis~\cite{sandler2023training}, that can be applied to analyze the robustness and efficiency of different optimization algorithms, including AdaTerm.
Finally, latent feature analysis methods~\cite{luo2020position} have attracted extensive attention in recent years. The use of AdaTerm in such approaches is therefore an interesting field of study that we will investigate in the future. 

\section*{Acknowledgements}

This work was supported by JSPS KAKENHI, Grant-in-Aid for Scientific Research (B), Grant Number JP20H04265.

\bibliographystyle{elsarticle-num}
\bibliography{biblio}


\clearpage
\appendix
\onecolumn


\clearpage
\section{Proof of regret bounds}
\label{apdx:proofs}

%
Below is the proof and the detailed analysis of the AdaTerm algorithm.
Following the literature, we ignore the bias correction term.
We also consider a scheduled learning rate $\alpha_{t} = \alpha (\sqrt{t})^{-1}$ and the non-expansive weighted projection operator $\Pi_{\mathcal{F}, V}(\theta) = \argmin{\theta' \in \mathcal{F}}{ \norm{\theta' - \theta}_{V} }$.
The use of the projection operator is not necessary for deriving the upper bound; in fact, no projection is used in Algorithm~\ref{alg:adaterm}.
Regardless, it provides a more general setting, and therefore, following the literature, we consider its potential use in our proof.

\subsection{Preliminary}

 We start by stating intermediary results that we shall use later to bound the regret.
\begin{lemma}[Peter?Paul inequality or Young's inequality with $\zeta$ and exponent 2] \label{lem:pp_ineq}
$\forall \zeta > 0$ and $\forall (x, y) \in \mathbb{R}^2$ we have that $xy \leq \frac{\zeta}{2} x^2 + \frac{1}{2\zeta} y^2$.
\end{lemma}

From which we obtain the following results:
\begin{corollary}[Partial bound of $\langle m_{t}, \theta_{t} - \theta^{*} \rangle$] \label{corol:partial_bound_mt_D}
Let $(m_t)_{t\geq 0}$, $(v_t)_{t\geq 0}$ and $(\theta_{t})_{t\geq 0}$ be the sequence of moment, variance and iterates produced by AdaTerm with a scheduled learning rate $\alpha_{t} > 0, \forall t$. Then, we have:
\begin{align}
\langle m_{t}, \theta_{t} - \theta^{*} \rangle &\leq \frac{D^{2}}{2\alpha_{t}} \norm{v_{t}^{1/4}}^{2} + \frac{1}{2} \alpha_{t} \norm{m_{t}}^{2}_{v_{t}^{1/2}}
\end{align}
\end{corollary}
\begin{proof}
This follows from the application of Lemma~\ref{lem:pp_ineq} with the identification:
\begin{align*}
\zeta &= \frac{v_{t, i}^{1/2}}{\alpha_{t}},\; x_{i} = (\theta_{t, i} - \theta^{*}_{i}),\; y_{i} = m_{t, i}
\end{align*}
Combined with the fact that $\forall i\in [d]$, $(\theta_{t, i} - \theta^{*}_{i}) \leq D$.
\end{proof}

\subsection{Introduction}

Let $m_t$ and $v_t$ be the first- and second-order moments produced by AdaTerm, then:
\begin{align}
    m_t &= (1-\tau_t) m_{t-1} + \tau_t g_t\;,\;\;\;\; v_t = (1-\tau_t) m_{t-1} + \tau_t \left( (g_t - m_{t-1})^2 + \Delta s \right)
\end{align}

From the definition of $m_{t}$, we can derive the following:
\begin{align}
    g_t = \frac{1}{\tau_t} m_t - \frac{1-\tau_t}{\tau_t} m_{t-1}
\end{align}
which leads to:
\begin{align}
    \langle g_t, \theta_t - \theta^{*} \rangle &= \frac{1}{\tau_t} \langle m_t, \theta_t - \theta^{*} \rangle - \frac{1-\tau_t}{\tau_t} \langle m_{t-1}, \theta_t - \theta^{*} \rangle = \frac{1}{\tau_t} \langle m_t, \theta_t - \theta^{*} \rangle - \frac{1-\tau_t}{\tau_t} \langle m_{t-1}, \theta_{t-1} - \theta^{*} \rangle - \frac{1-\tau_t}{\tau_t} \langle m_{t-1}, \theta_t - \theta_{t-1} \rangle \\
    &= \frac{1}{\tau_t} \left( \langle m_t, \theta_t - \theta^{*} \rangle - \langle m_{t-1}, \theta_{t-1} - \theta^{*} \rangle \right) + \langle m_{t-1}, \theta_{t-1} - \theta^{*} \rangle - \frac{1-\tau_t}{\tau_t} \langle m_{t-1}, \theta_t - \theta_{t-1} \rangle
\end{align}

We therefore have that:
\begin{align}
    \sum\limits_{t=1}^{T} \langle g_t, \theta_t - \theta^{*} \rangle &= R_{1,T} + R_{2,T} + R_{3,T} \\
    R_{1,T} &= \sum\limits_{t=1}^{T} \langle m_{t-1}, \theta_{t-1} - \theta^{*} \rangle = \sum\limits_{t=1}^{T-1} \langle m_{t}, \theta_{t} - \theta^{*} \rangle \;\;\;\mathrm{(since}\; m_0 = 0 \mathrm{)} \\
    R_{2,T} &= - \sum\limits_{t=1}^{T} \frac{1-\tau_t}{\tau_t} \langle m_{t-1}, \theta_t - \theta_{t-1} \rangle = \sum\limits_{t=1}^{T} \frac{1-\tau_t}{\tau_t} \langle m_{t-1}, \theta_{t-1} - \theta_{t} \rangle \\
    R_{3,T} &= \sum\limits_{t=1}^{T} \frac{1}{\tau_t} \left( \langle m_t, \theta_t - \theta^{*} \rangle - \langle m_{t-1}, \theta_{t-1} - \theta^{*} \rangle \right)
\end{align}

\subsection{Bound of $R_{1,T}$}

Using Corollary~\ref{corol:partial_bound_mt_D}, we have that:
\begin{align}
  R_{1,T} &= \sum\limits_{t=1}^{T-1} \langle m_{t}, \theta_{t} - \theta^{*} \rangle \leq \sum\limits_{t=1}^{T-1} \frac{D^{2}}{2\alpha_{t}} \norm{v_{t}^{1/4}}^{2} + \frac{1}{2} \sum\limits_{t=1}^{T-1} \alpha_{t} \norm{m_{t}}^{2}_{v_{t}^{1/2}}
\end{align}

\subsection{Bound of $R_{2,T}$}

With $\tilde{m}_{t-1} = \frac{1-\tau_t}{\tau_t} m_{t-1}$ and by using $m_0 = 0$, we have:
\begin{align*}
    R_{2,T} &= \sum\limits_{t=1}^{T} \langle \tilde{m}_{t-1}, \theta_{t-1} - \theta_{t} \rangle = \sum\limits_{t=2}^{T} \langle \tilde{m}_{t-1}, \theta_{t-1} - \theta_{t} \rangle = \sum\limits_{t=1}^{T-1} \langle \tilde{m}_{t}, \theta_{t} - \theta_{t+1} \rangle
\end{align*}
Then, by applying Holder's inequality, followed by the use of the projection operator where $\theta_t = \Pi_{\mathcal{F}, \hat{v}_{t}^{1/2}}(\theta_{t})$, since $\theta_t \in \mathcal{F}$, we obtain:
\begin{align*}
    R_{2,T} &\leq \sum\limits_{t=1}^{T-1} \norm{ \tilde{m}_{t} }_{\hat{v}_{t}^{-1/2}}  \norm{\theta_{t} - \theta_{t+1}}_{\hat{v}_t^{1/2}} = \sum\limits_{t=1}^{T-1} \norm{ \tilde{m}_{t} }_{\hat{v}_{t}^{-1/2}}  \norm{\Pi_{\mathcal{F}, \hat{v}_{t}^{1/2}}(\theta_{t}) - \Pi_{\mathcal{F}, \hat{v}_{t}^{1/2}}(\theta_t - \alpha_t \hat{v}_{t}^{-1/2} m_t)}_{\hat{v}_t^{1/2}}
\end{align*}

The non-expansive property of the operator allows the following:
\begin{align*}
    R_{2,T} &\leq \sum\limits_{t=1}^{T-1} \norm{ \tilde{m}_{t} }_{\hat{v}_{t}^{-1/2}}  \norm{\theta_{t} - \theta_{t+1}}_{\hat{v}_t^{1/2}} \leq \sum\limits_{t=1}^{T-1} \norm{ \tilde{m}_{t} }_{\hat{v}_{t}^{-1/2}}  \norm{\theta_{t} - (\theta_t - \alpha_t \hat{v}_{t}^{-1/2} m_t)}_{\hat{v}_t^{1/2}} = \sum\limits_{t=1}^{T-1} \norm{ \tilde{m}_{t} }_{\hat{v}_{t}^{-1/2}}  \norm{\alpha_t \hat{v}_{t}^{-1/2} m_t}_{\hat{v}_t^{1/2}} = \sum\limits_{t=1}^{T-1} \alpha_t \norm{ \tilde{m}_{t} }_{\hat{v}_{t}^{-1/2}} \norm{ m_{t} }_{\hat{v}_{t}^{-1/2}}
\end{align*}

Finally, by restoring $\tilde{m}_{t} = \frac{1-\tau_{t+1}}{\tau_{t+1}} m_{t}$ and using $\frac{1-\tau_{t+1}}{\tau_{t+1}} \leq \frac{1-\underline{\tau}}{\underline{\tau}}$, we have:
\begin{align}
    R_{2,T} &\leq \sum\limits_{t=1}^{T-1} \alpha_t \frac{1-\tau_{t+1}}{\tau_{t+1}} \norm{ m_{t} }_{\hat{v}_{t}^{-1/2}}^2 \leq \frac{1-\underline{\tau}}{\underline{\tau}} \sum\limits_{t=1}^{T-1} \alpha_t \norm{ m_{t} }_{\hat{v}_{t}^{-1/2}}^2
\end{align}

\subsection{Bound of $R_{3,T}$}

\begin{align*}
    R_{3,T} &= \sum\limits_{t=1}^{T} \frac{1}{\tau_t} \left( \langle m_t, \theta_t - \theta^{*} \rangle - \langle m_{t-1}, \theta_{t-1} - \theta^{*} \rangle \right) = \sum\limits_{t=1}^{T} A_{t}\\
    A _{t} &= \frac{1}{\tau_t} \langle m_t, \theta_t - \theta^{*} \rangle - \frac{1}{\tau_{t-1}} \langle m_{t-1}, \theta_{t-1} - \theta^{*} \rangle + \left( \frac{\tau_{t} - \tau_{t-1}}{\tau_{t} \tau_{t-1}} \right) \langle m_{t-1}, \theta_{t-1} - \theta^{*} \rangle \\
    &\leq \frac{1}{\tau_t} \langle m_t, \theta_t - \theta^{*} \rangle - \frac{1}{\tau_{t-1}} \langle m_{t-1}, \theta_{t-1} - \theta^{*} \rangle + \left( \frac{|\tau_{t} - \tau_{t-1} |}{2\tau_{t} \tau_{t-1}} \right) \left\{ \frac{D^{2}}{\alpha_{t-1}} \norm{v_{t-1}^{1/4}}^{2} + \alpha_{t-1} \norm{m_{t-1}}^{2}_{v_{t-1}^{1/2}} \right\}
\end{align*}
where the last line is obtained by applying Corollary~\ref{corol:partial_bound_mt_D} on time step t-1.
Finally, using the telescoping sum over the first two terms in $A_{t}$ with the particularity of $m_{0} = 0$, combined with $|\tau_{t} - \tau_{t-1}| \leq (1 - \beta) - \underline{\tau} = \left[1 - (\beta + \underline{\tau})\right]$, we arrive at the following relation:
\begin{align}
    \sum\limits_{t=1}^{T} A_{t} &\leq \frac{1}{\tau_T} \langle m_T, \theta_T - \theta^{*} \rangle + \left( \frac{1 - (\beta + \underline{\tau})}{2\underline{\tau}^2} \right) \left\{ \sum\limits_{t=1}^{T} \frac{D^{2}}{\alpha_{t-1}} \norm{v_{t-1}^{1/4}}^{2} + \sum\limits_{t=1}^{T} \alpha_{t-1} \norm{m_{t-1}}^{2}_{v_{t-1}^{1/2}} \right\} \nonumber \\
    R_{3,T} &\leq \frac{1}{\tau_T} \langle m_T, \theta_T - \theta^{*} \rangle + \left( \frac{1 - (\beta + \underline{\tau})}{2\underline{\tau}^2} \right) \left\{ \sum\limits_{t=1}^{T-1} \frac{D^{2}}{\alpha_{t}} \norm{v_{t}^{1/4}}^{2} + \sum\limits_{t=1}^{T-1} \alpha_{t} \norm{m_{t}}^{2}_{v_{t}^{1/2}} \right\}
\end{align}

\subsection{Bound of $\frac{1}{\tau_T} \langle m_T, \theta_T - \theta^{*} \rangle$}

 Applying Lemma~\ref{lem:pp_ineq} with $\zeta = \frac{v_{T,i}^{1/2}}{2\alpha_{T}}$ and $x_{i} = (\theta_{T,i} - \theta^{*}_{i})$, $y_{i} = m_{T, i}$, we get:
\begin{align}
\frac{1}{\tau_T} \langle m_T, \theta_T - \theta^{*} \rangle \leq \frac{1}{\tau_T} \left[ \frac{D^{2}}{4\alpha_{T}} \norm{v_{T}^{1/4}}^{2} + \alpha_{T} \norm{m_{T}}^{2}_{v_{T}^{1/2}} \right] \leq \frac{1}{\tau_T} \frac{D^{2}}{4\alpha_{T}} \norm{v_{T}^{1/4}}^{2} + \frac{(1-\beta)^2 \alpha_{T}}{\epsilon \underline{\tau}^2} \sum\limits_{i=1}^d \sum\limits_{k=1}^{T} (1 - \underline{\tau})^{T-k} g^2_{k,i}
\end{align}

\subsection{Bound of $\sum\limits_{t=1}^{T-1} \alpha_t \norm{ m_{t} }_{\hat{v}_{t}^{-1/2}}^2$}

With $\alpha_{t} = \frac{\alpha}{\sqrt{t}}$, we have:
\begin{align*}
    \sum\limits_{t=1}^{T-1} \alpha_t \norm{ m_{t} }_{\hat{v}_{t}^{-1/2}}^2 &= \sum\limits_{t=1}^{T-1} \alpha_t \norm{ \hat{v}_{t}^{-1/4} m_{t} }^2_2 \leq \sum\limits_{t=1}^{T-1} \frac{\alpha}{\epsilon\sqrt{t}} \norm{ m_{t} }^2 = \sum\limits_{t=1}^{T-1} \frac{\alpha}{\epsilon\sqrt{t}} \sum\limits_{i=1}^d \left( \sum\limits_{k=1}^{t} \tau_{k} g_{k,i} \prod\limits_{j=1}^{t-k} (1-\tau_{k + j}) \right)^2 \\
    &\leq \sum\limits_{t=1}^{T-1} \frac{(1-\beta)^{2} \alpha}{\epsilon\sqrt{t}} \sum\limits_{i=1}^d \left( \sum\limits_{k=1}^{t} |g_{k,i}| (1 - \underline{\tau})^{t-k} \right)^2
    \leq \sum\limits_{t=1}^{T-1} \frac{(1-\beta)^{2} \alpha}{\epsilon\sqrt{t}} \sum\limits_{i=1}^d \left( \sum\limits_{k=1}^{t} (1 - \underline{\tau})^{t-k} \right) \left( \sum\limits_{k=1}^{t} (1 - \underline{\tau})^{t-k} g_{k,i}^2 \right) \\
    &\leq \frac{(1-\beta)^{2} \alpha}{\epsilon} \sum\limits_{t=1}^{T-1} \left( \frac{1 - (1 - \underline{\tau})^{t}}{\underline{\tau}} \right) \sum\limits_{i=1}^d \sum\limits_{k=1}^{t} (1 - \underline{\tau})^{t-k} g_{k,i}^2 \frac{1}{\sqrt{t}}
    \leq \frac{(1-\beta)^{2} \alpha}{\epsilon\underline{\tau}} \sum\limits_{t=1}^{T-1} \sum\limits_{i=1}^d \sum\limits_{k=1}^{t} (1 - \underline{\tau})^{t-k} g_{k,i}^2 \frac{1}{\sqrt{t}} \\
    &= \frac{(1-\beta)^{2} \alpha}{\epsilon\underline{\tau}} \sum\limits_{i=1}^d \sum\limits_{t=1}^{T-1} g_{t,i}^2 \sum\limits_{k=t}^{T-1} (1 - \underline{\tau})^{k-t} \frac{1}{\sqrt{k}}
    \leq \frac{(1-\beta)^{2} \alpha}{\epsilon\underline{\tau}} \sum\limits_{i=1}^d \sum\limits_{t=1}^{T-1} g_{t,i}^2 \sum\limits_{k=t}^{T-1} (1 - \underline{\tau})^{k-t} \frac{1}{\sqrt{t}} \\
    &= \frac{(1-\beta)^{2} \alpha}{\epsilon\underline{\tau}} \sum\limits_{i=1}^d \sum\limits_{t=1}^{T-1} g_{t,i}^2 \frac{1}{\sqrt{t}} \left( \frac{1 - (1 - \underline{\tau})^{T-t}}{\underline{\tau}} \right)
    \leq \frac{(1-\beta)^{2} \alpha}{\epsilon\underline{\tau}^2} \sum\limits_{i=1}^d \sum\limits_{t=1}^{T-1} g_{t,i}^2 \frac{1}{\sqrt{t}}  \;\;\;\mathrm{(Since}\;(1 - (1 - \underline{\tau})^{T-t}) < 1 \mathrm{)}
\end{align*}

Finally, using Cauchy-Schwartz's inequality (without the squares) with $u_t = g_{t,i}^2$ and $v_t = \frac{1}{\sqrt{t}}$, and with $\sum\limits_{t=1}^{T-1} \frac{1}{t} \leq 1 + \ln (T-1)$, we get:
\begin{align}
    \sum\limits_{t=1}^{T-1} \alpha_t \norm{ m_{t} }_{\hat{v}_{t}^{-1/2}}^2 &\leq \frac{\alpha (1-\beta)^2}{\epsilon \underline{\tau}^2 } \sum\limits_{i=1}^d \norm{g_{1:T-1,i}^2}_2 \sqrt{\sum\limits_{t=1}^{T-1} \frac{1}{t}} \leq \frac{\alpha (1-\beta)^2}{\epsilon \underline{\tau}^2} \sqrt{1 + \ln (T-1)} \sum\limits_{i=1}^d \norm{g_{1:T-1,i}^2}_2
\end{align}

where $\norm{g_{1:T-1,i}^2}_2 = \sqrt{\sum\limits_{t=1}^{T-1} g_{t, i}^{4}} \leq G_{\infty} \sqrt{T-1}$ where $G_{\infty} = \underset{t \in [T-1]}{\max} \norm{g_{t}}_{\infty}$.

\subsection{Bound of the regret}

With
\begin{align*}
    R_{T} &= \sum\limits_{t=1}^{T} \mathcal{L}_{\mathcal{B}_t}(\theta_t) - \mathcal{L}_{\mathcal{B}_t}(\theta^*) \leq \sum\limits_{t=1}^{T} \langle g_t, \theta_t - \theta^{*} \rangle = R_{1,T} + R_{2,T} + R_{3,T}
\end{align*}
the bound of the regret is obtained by combining the upper bounds of $R_{1,T}$, $R_{2,T}$ and $R_{3,T}$ and through a straightforward factorization:
\begin{align}
\begin{split}
    R_{T} &\leq \left. \frac{D^2\sqrt{T}}{4\tau_{T}\alpha} \sum\limits_{i=1}^{d} v_{T,i}^{1/2} + \left[ \frac{\underline{\tau}^2 + 1 - (\beta + \underline{\tau})}{2\underline{\tau}^2} \right] \sum\limits_{t=1}^{T-1} \frac{D^2}{\alpha_{t}} \sum\limits_{i=1}^{d} v_{t,i}^{1/2} \right. \\
    &\left.\; + \left[ \frac{(1-\beta)^2 \alpha}{\epsilon\underline{\tau}^2\sqrt{T}}  \right] \sum\limits_{t=1}^{T-1} \sum\limits_{i=1}^{d} (1 - \underline{\tau})^{T-k} g^2_{k,i} \right. \\
    &\;+ \left[ \frac{1}{2} + \frac{1 - \underline{\tau}}{\underline{\tau}} + \frac{1 - (\beta + \underline{\tau})}{2\underline{\tau}^2} \right]\left[ \frac{(1-\beta)^2 \alpha}{\epsilon\underline{\tau}^2} \right] \sqrt{1 + \ln (T-1)} \sum\limits_{i=1}^d \norm{g_{1:T-1,i}^2}_2
\end{split}
\end{align}

\section{Visualization of the key elements of AdaTerm} \label{apdx:adaterm_algo_vis}

\begin{figure}[ht]
    \centering
    \includegraphics[keepaspectratio=true,width=0.95\linewidth]{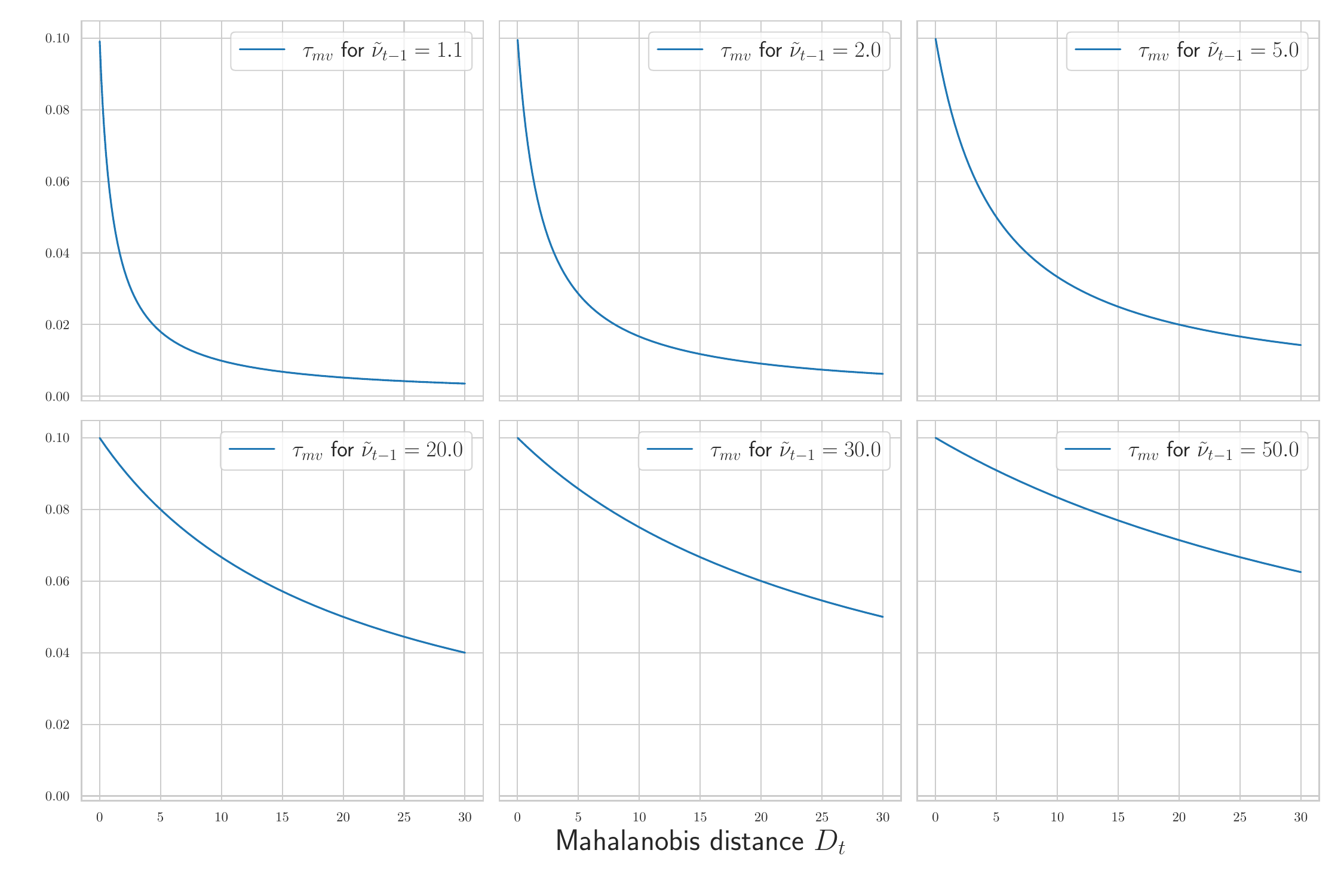}
    \caption{Visualization of $\tau_{mv}$ against $\tilde{\nu}_{t-1}$ and $D_{t}$ values (with $\beta = 0.9$)}
    \label{fig:adaterm_algo_vis}
\end{figure}

\begin{figure}[ht]
    \centering
    \includegraphics[keepaspectratio=true,width=0.95\linewidth]{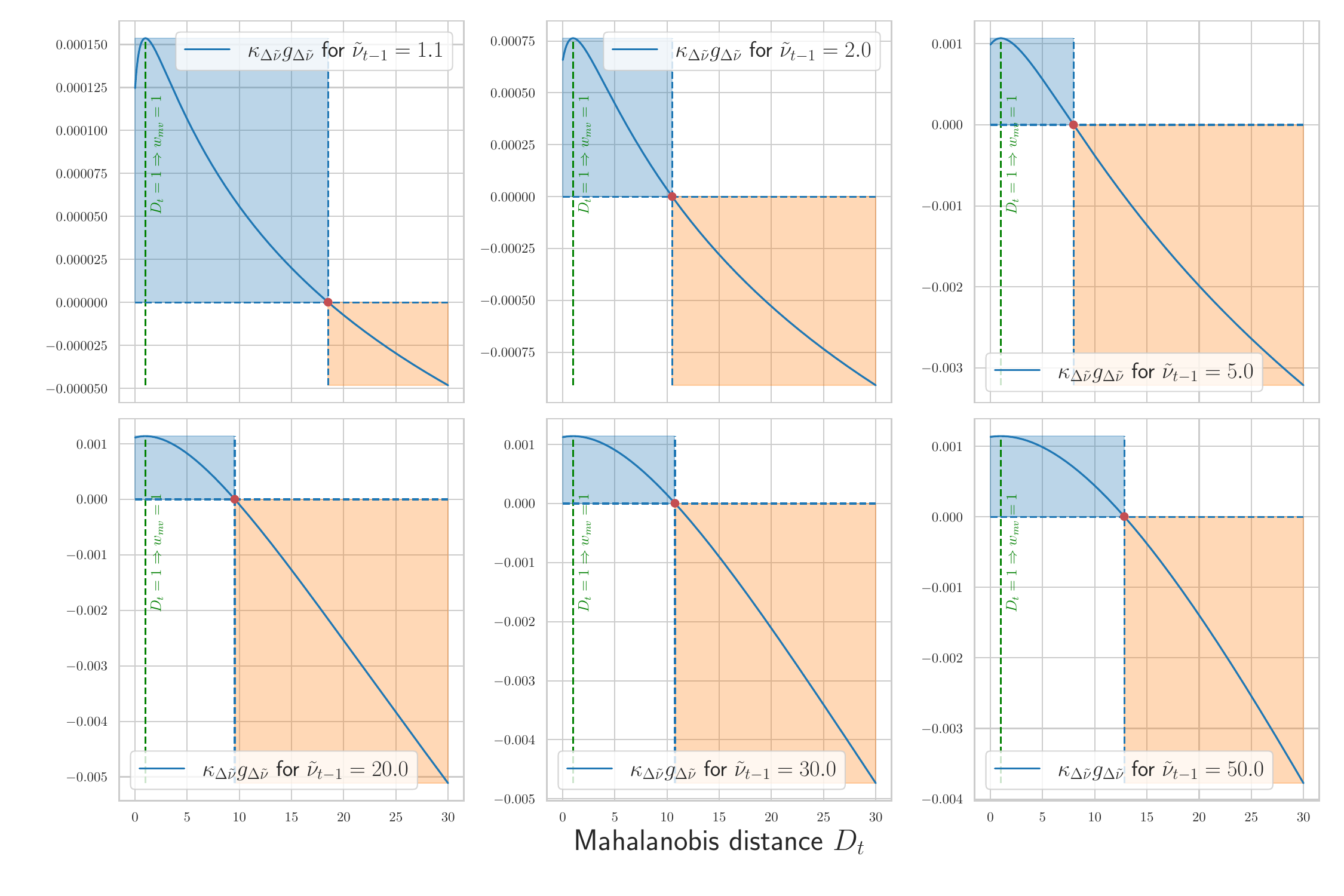}
    \caption{Visualization of $\kappa_{\Delta \tilde{\nu}} g_{\tilde{\nu}}$ against $\tilde{\nu}_{t-1}$ and $D_{t}$ values (with $\beta = 0.9$)}
    \label{fig:adaterm_algo_vis_dofIncrease}
\end{figure}

Figures~\ref{fig:adaterm_algo_vis} and~\ref{fig:adaterm_algo_vis_dofIncrease} show the visual description of the interpolation factor $\tau_{mv}$ and of the gradient ascent increment $\kappa_{\Delta \tilde{\nu}} g_{\tilde{\nu}}$ for the robustness parameter $\tilde{\nu}$, respectively.
As can be seen in Fig.~\ref{fig:adaterm_algo_vis}, for small deviations $D_t$, $\tau_{mv} \to (1 - \beta)$, and conversely, when $D_t$ is large, $\tau_{mv} \to 0$ to alleviate the influence of the aberrant gradient on both $m_t$ and $v_t$.
Similarly, Fig.~\ref{fig:adaterm_algo_vis_dofIncrease} shows how the degrees-of-freedom is incremented based on the value of $D_t$.
The blue region corresponds to a region where $\lambda_t$ is larger than the current robustness value $\tilde{\nu}_{t-1}$, leading to an increase in $\tilde{\nu}_{t}$.
Conversely, the orange region, corresponding to large $D_t$, prompts a decrease in the value of $\tilde{\nu}_{t}$ because $\kappa_{\Delta \tilde{\nu}} g_{\tilde{\nu}}$ is negative.

\section{Details of test functions}
\label{apdx:test}


The following test functions were tested: $f(x,y)$ were Rosenbrock with a valley, McCormick with a plate, and Michalewicz with steep drops, as defined below.
\begin{align}
    f(x,y) = \begin{cases}
        100(y - x^2)^2 + (x - 1)^2
        & \mathrm{Rosenbrock} \\
        \sin(x + y) + (x - y)^2 - 1.5 x + 2.5 y + 1
        & \mathrm{McCormick} \\
        - \sin(x)\sin^{20}(x^2/\pi)
        - \sin(y)\sin^{20}(2y^2/\pi)
        & \mathrm{Michalewicz}
    \end{cases}
\end{align}
Using the gradients of $f(x,y)$, $(x, y)$ can be optimized to minimize $f(x,y)$.
The initial values of $(x, y)$ were fixed to $(-2, 2)$, $(4, -3)$, and $(1, 1)$, respectively.
The noise added to $(x, y)$ for computing the gradients was sampled from the uniform distribution with range $(-0.1, 0.1)$.
The probability of noise was set to six conditions: $\{0, 1, 2.5, 5, 10, 15\}$\%.

The following optimizers were tested: Adam, At-Adam with adaptive $\nu$, and the proposed AdaTerm.
AdaBelief~\cite{zhuang2020adabelief}, similar to AdaTerm with $\nu \to \infty$, was also tested but excluded from Fig.~\ref{fig:comp_norm} and ~\ref{fig:comp_dof} owing to its similarity to Adam.
The learning rate was set to $\alpha=0.01$, which is higher than that for typical network updates; however, other hyper-parameters were left at their default settings.
The update was performed 15000 times, and the error norm between the final $(x, y)$ and the analytically-optimal point was computed.
To evaluate the statistics, 100 trials were run using different random seeds.

\begin{figure*}[ht]
    \centering
    \begin{subfigure}[b]{0.225\linewidth}
        \centering
        \includegraphics[keepaspectratio=true,width=\linewidth]{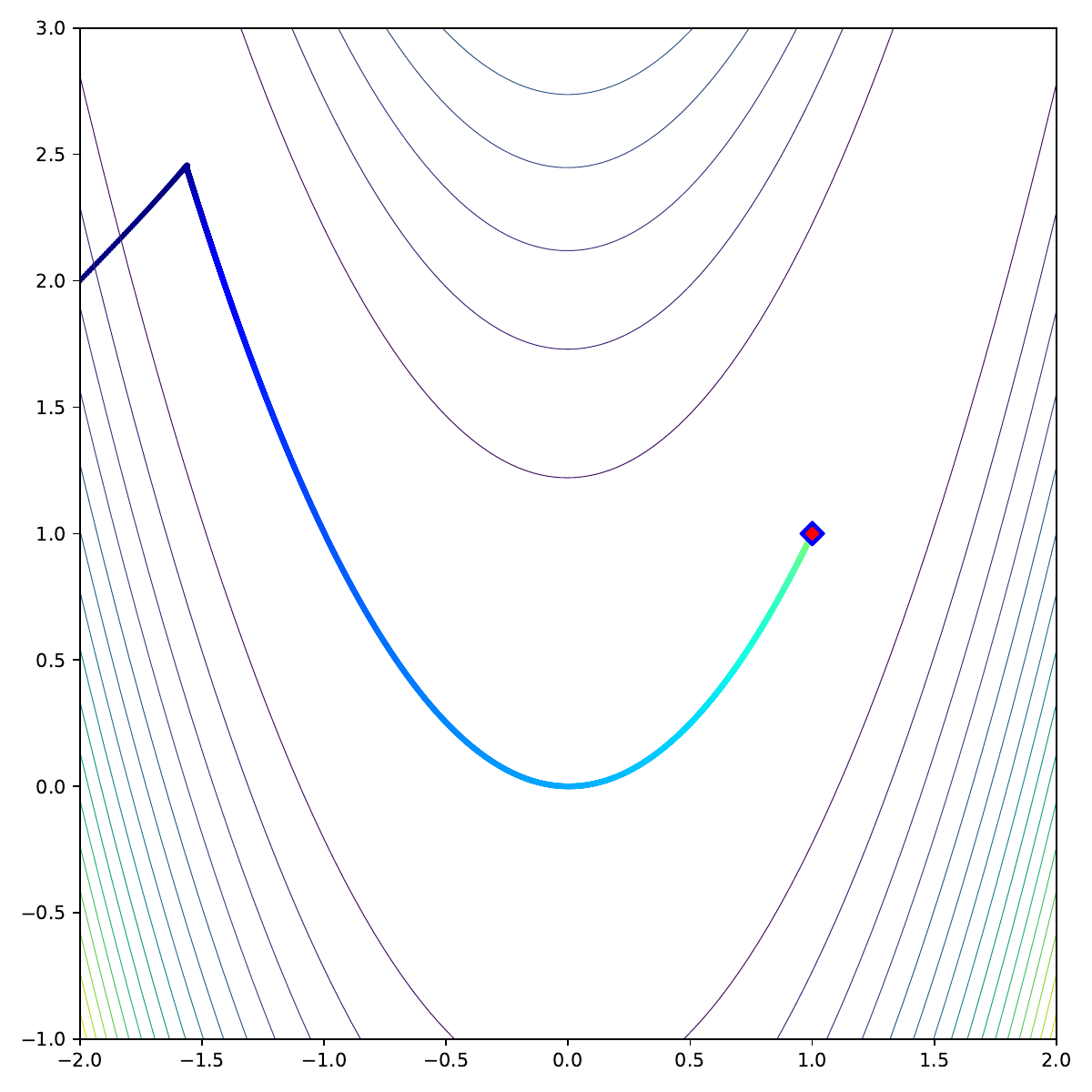}
        \subcaption{Adam at 0\% noise}
        \label{fig:trajectory_rosenbrock_0_adam}
    \end{subfigure}
    \begin{subfigure}[b]{0.225\linewidth}
        \centering
        \includegraphics[keepaspectratio=true,width=\linewidth]{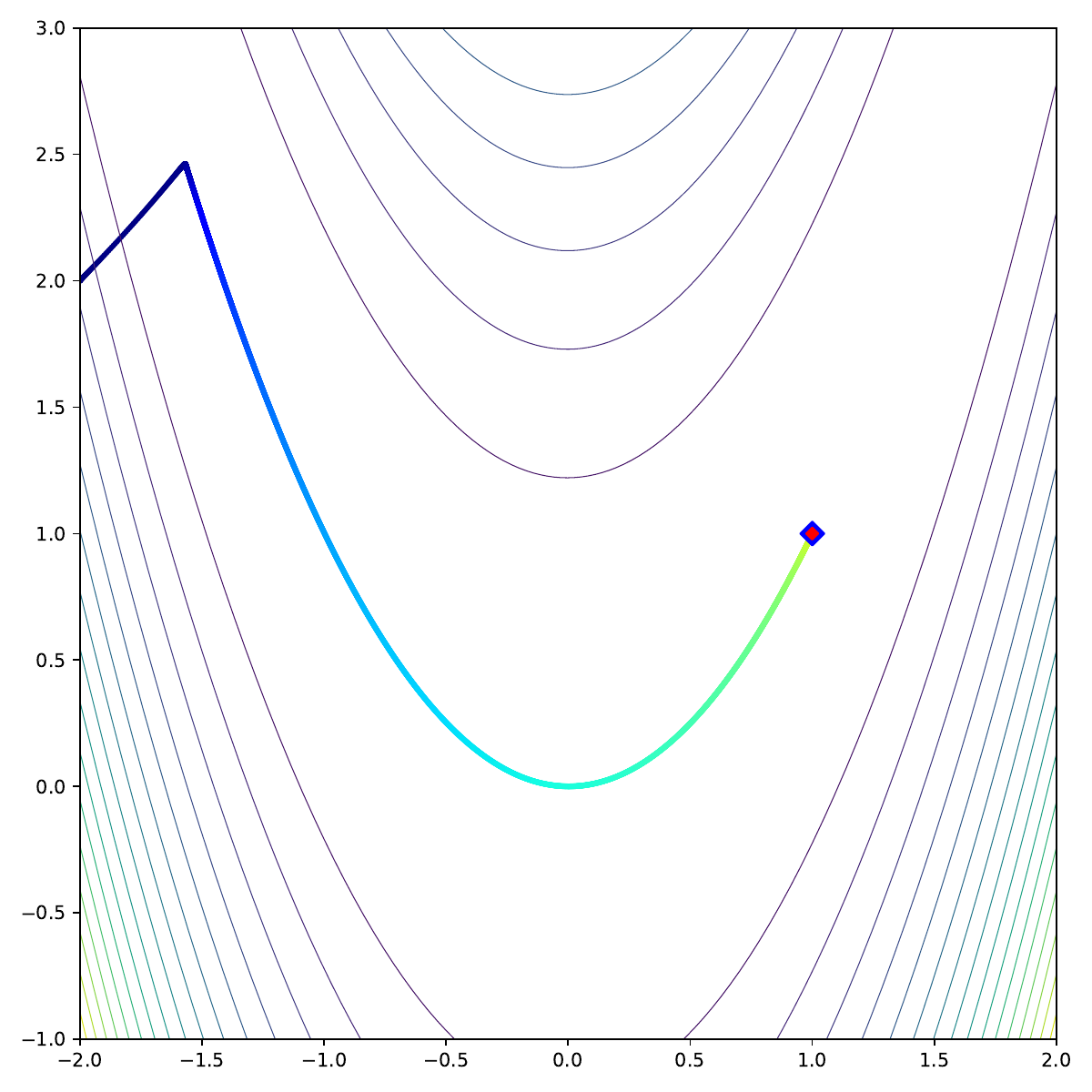}
        \subcaption{At-Adam at 0\% noise}
        \label{fig:trajectory_rosenbrock_0_atadam}
    \end{subfigure}
    \begin{subfigure}[b]{0.225\linewidth}
        \centering
        \includegraphics[keepaspectratio=true,width=\linewidth]{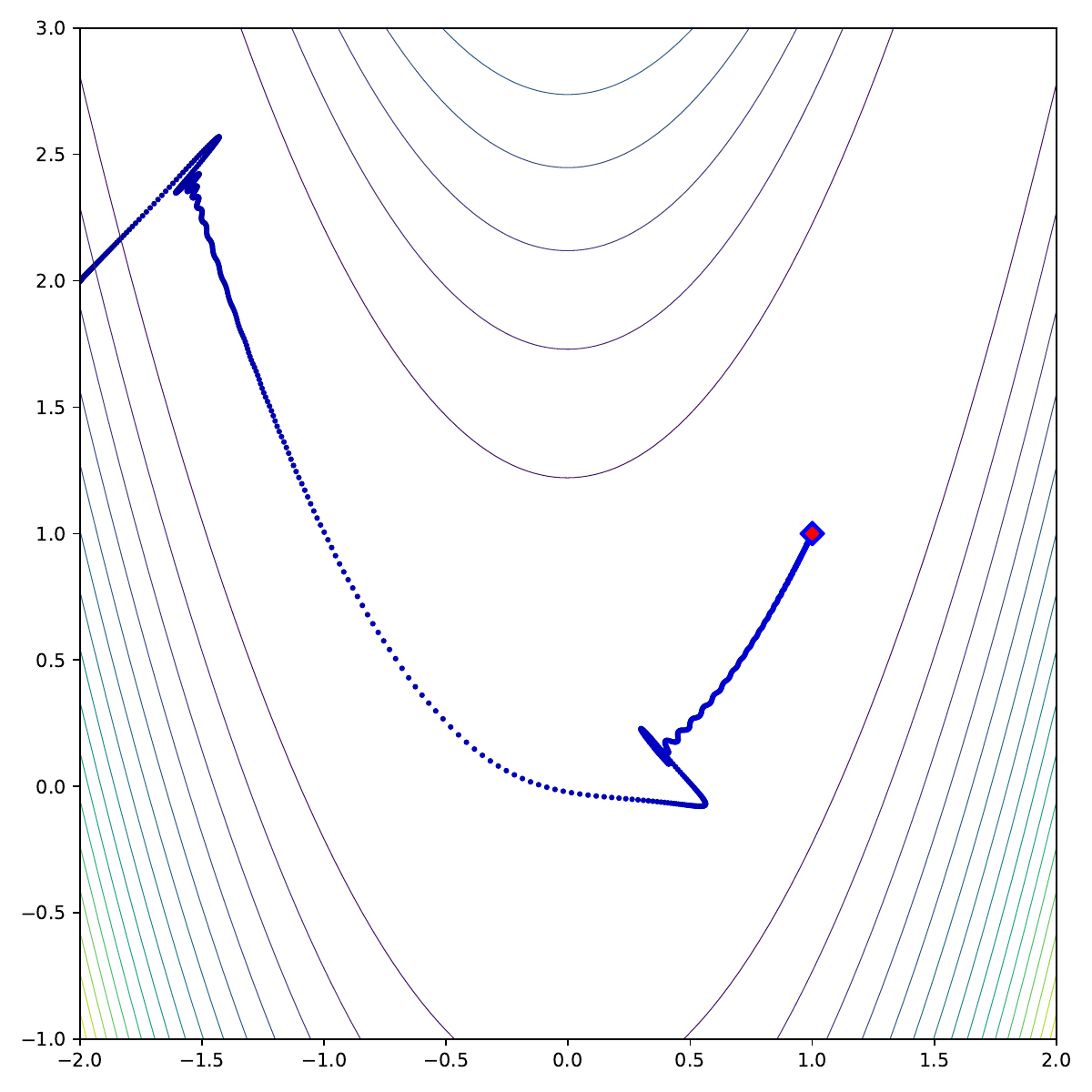}
        \subcaption{AdaTerm at 0\% noise}
        \label{fig:trajectory_rosenbrock_0_adaterm}
    \end{subfigure}
    \begin{subfigure}[b]{0.225\linewidth}
        \centering
        \includegraphics[keepaspectratio=true,width=\linewidth]{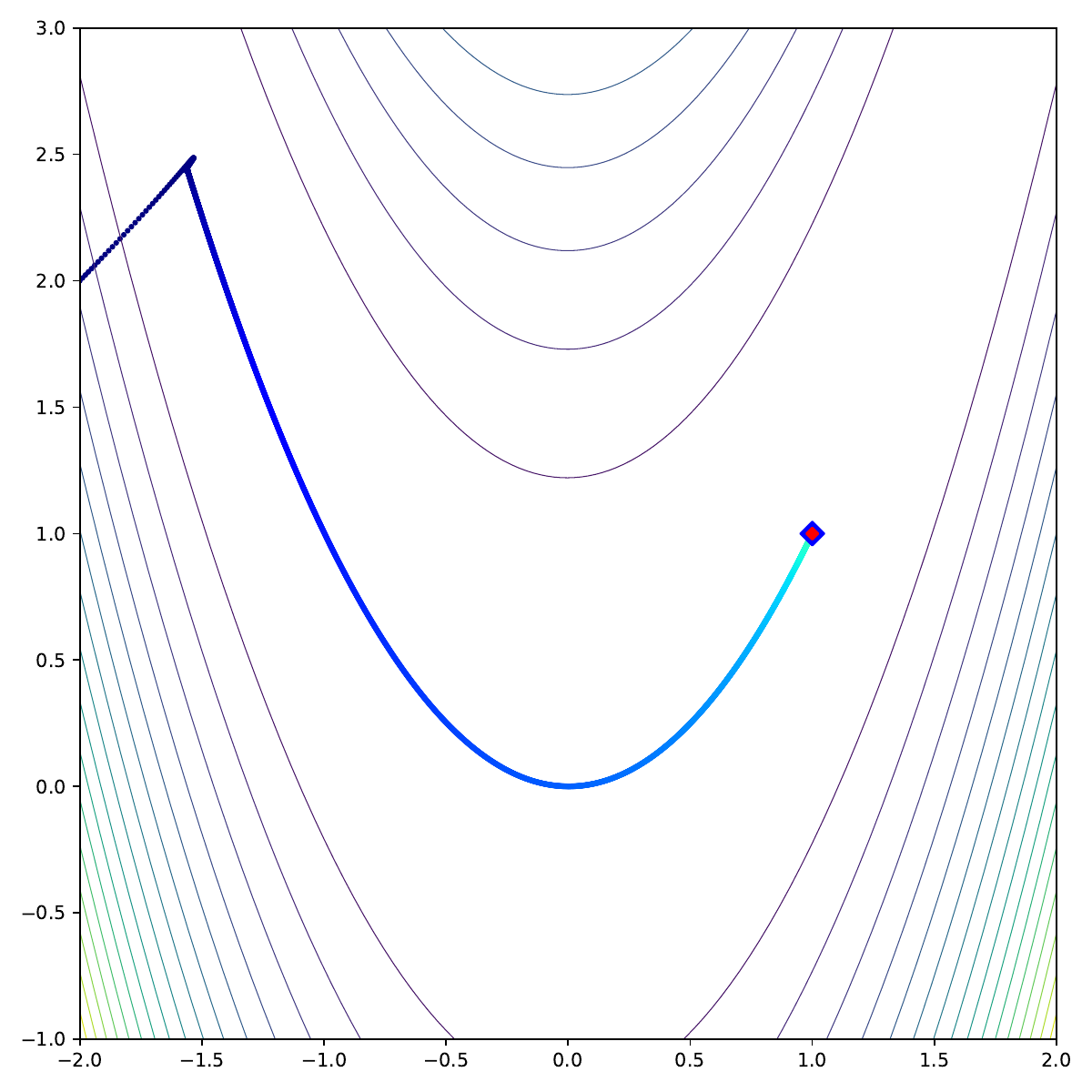}
        \subcaption{AdaBelief at 0\% noise}
        \label{fig:trajectory_rosenbrock_0_adabelief}
    \end{subfigure}
    \\
    \begin{subfigure}[b]{0.225\linewidth}
        \centering
        \includegraphics[keepaspectratio=true,width=\linewidth]{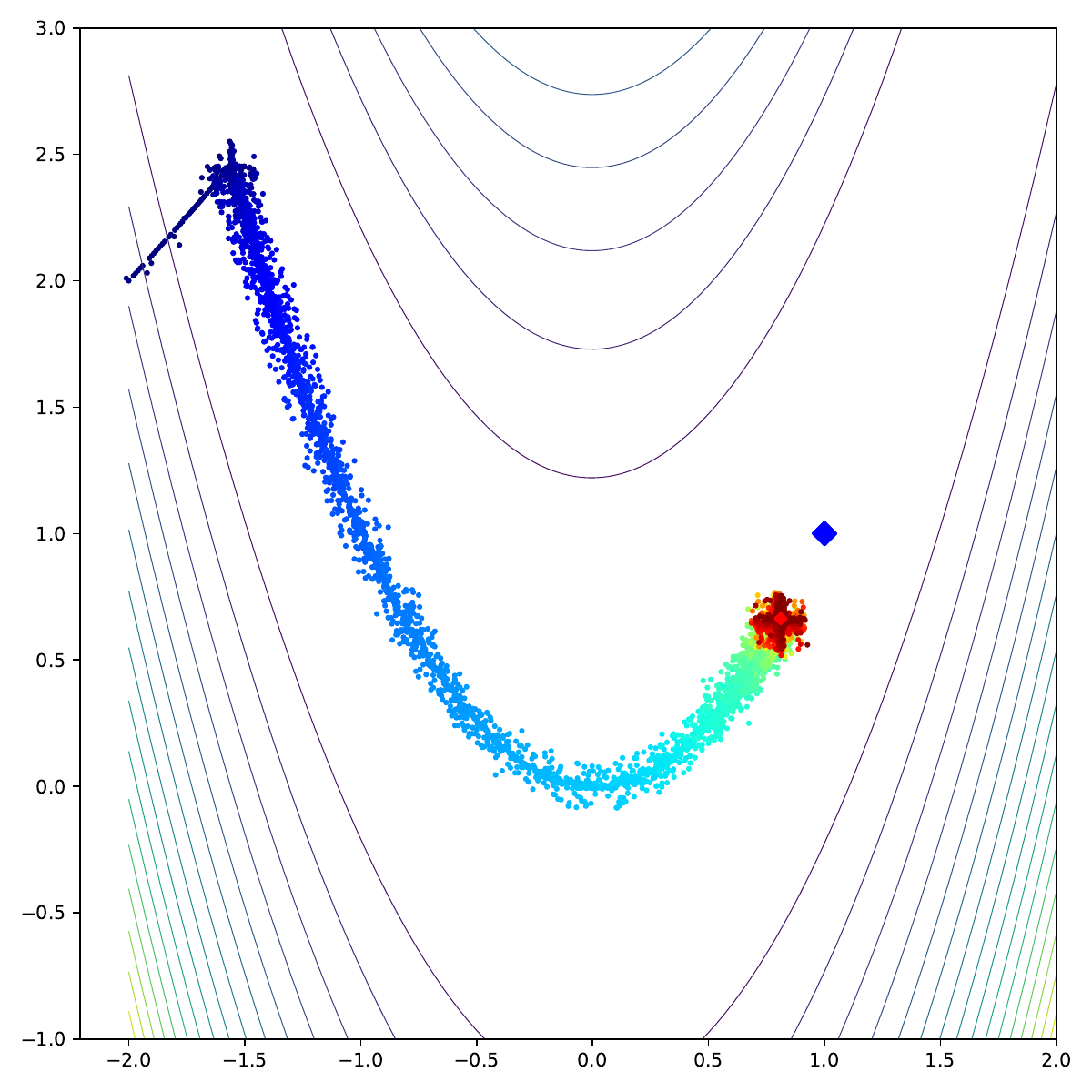}
        \subcaption{Adam at 15\% noise}
        \label{fig:trajectory_rosenbrock_15_adam}
    \end{subfigure}
    \begin{subfigure}[b]{0.225\linewidth}
        \centering
        \includegraphics[keepaspectratio=true,width=\linewidth]{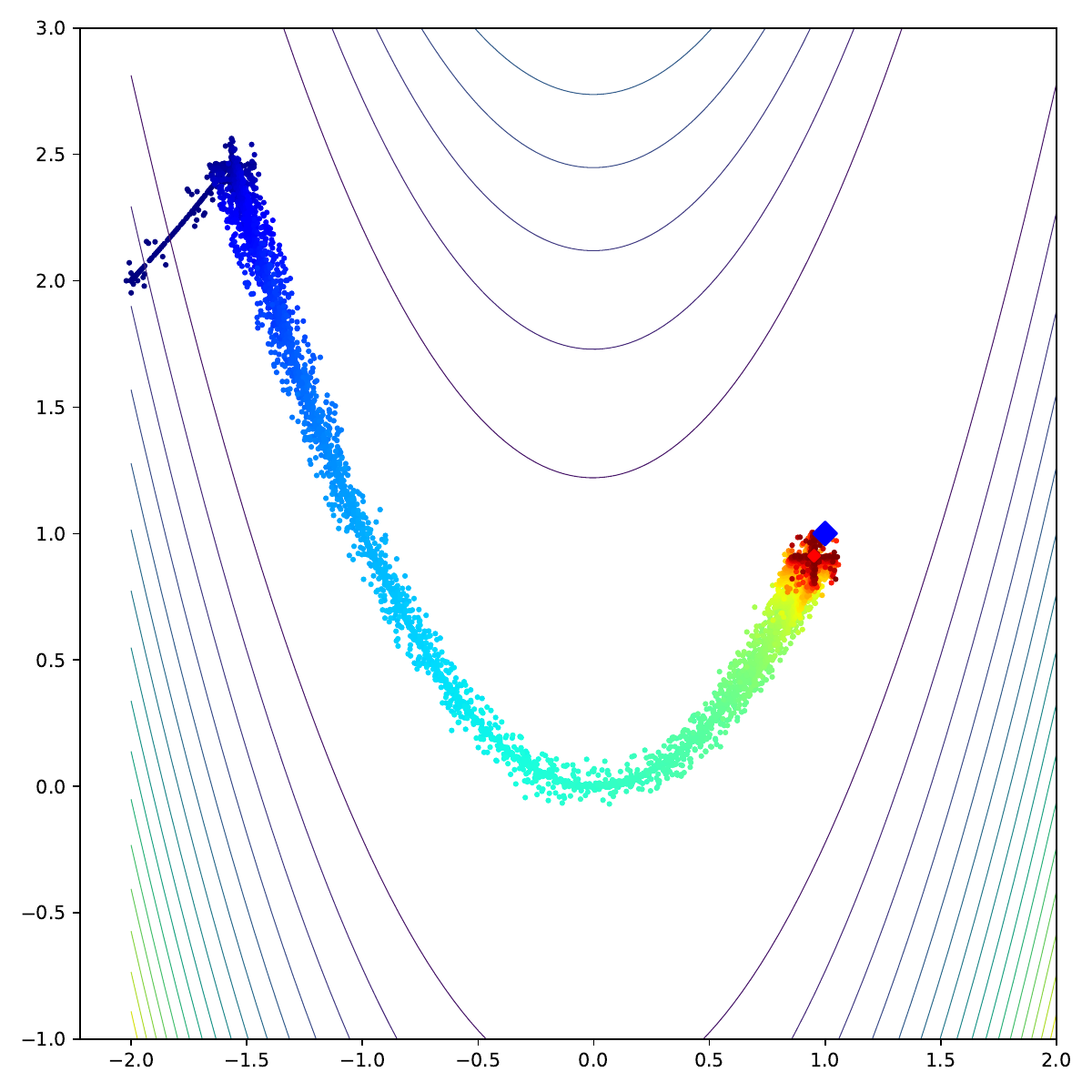}
        \subcaption{At-Adam at 15\% noise}
        \label{fig:trajectory_rosenbrock_15_atadam}
    \end{subfigure}
    \begin{subfigure}[b]{0.225\linewidth}
        \centering
        \includegraphics[keepaspectratio=true,width=\linewidth]{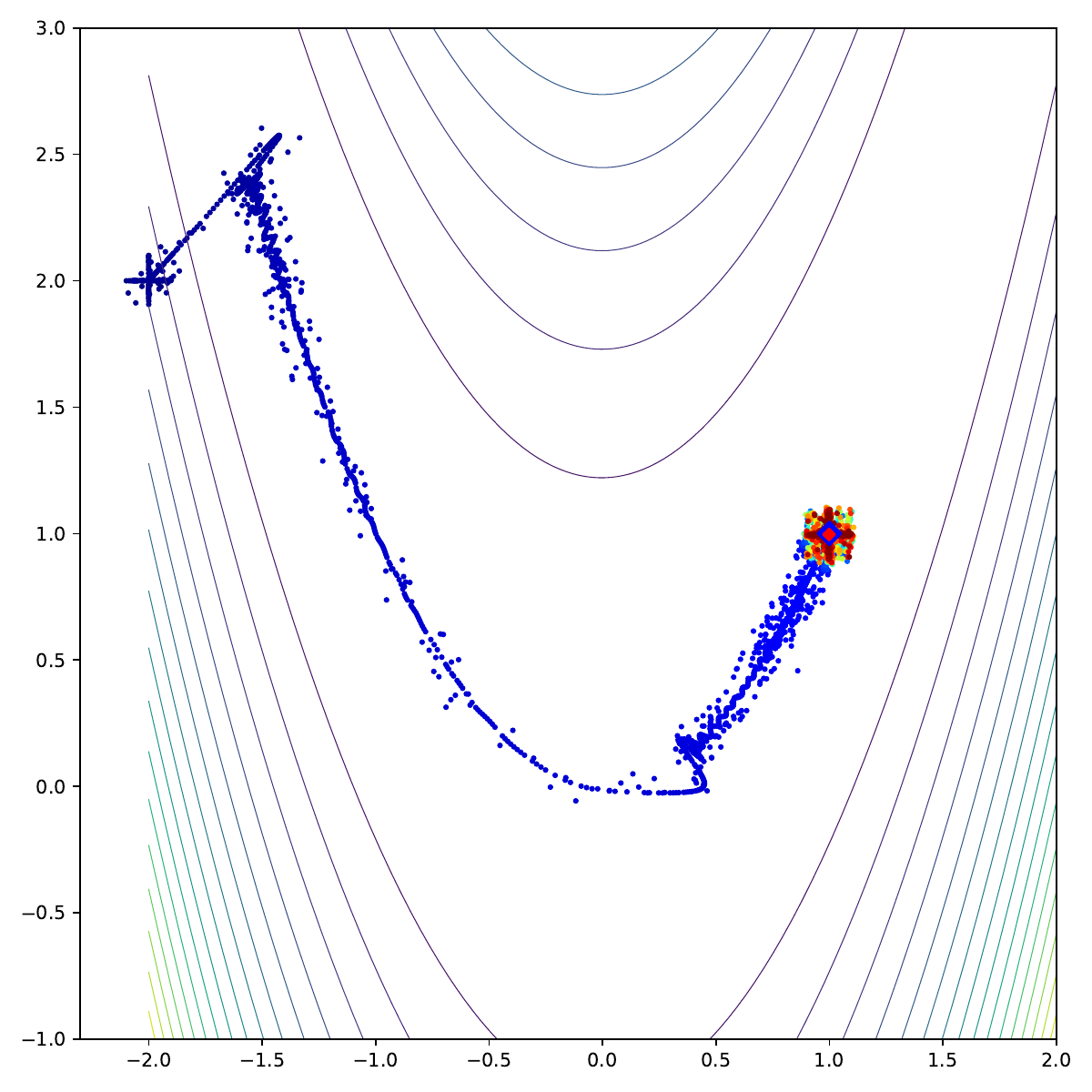}
        \subcaption{AdaTerm at 15\% noise}
        \label{fig:trajectory_rosenbrock_15_adaterm}
    \end{subfigure}
    \begin{subfigure}[b]{0.225\linewidth}
        \centering
        \includegraphics[keepaspectratio=true,width=\linewidth]{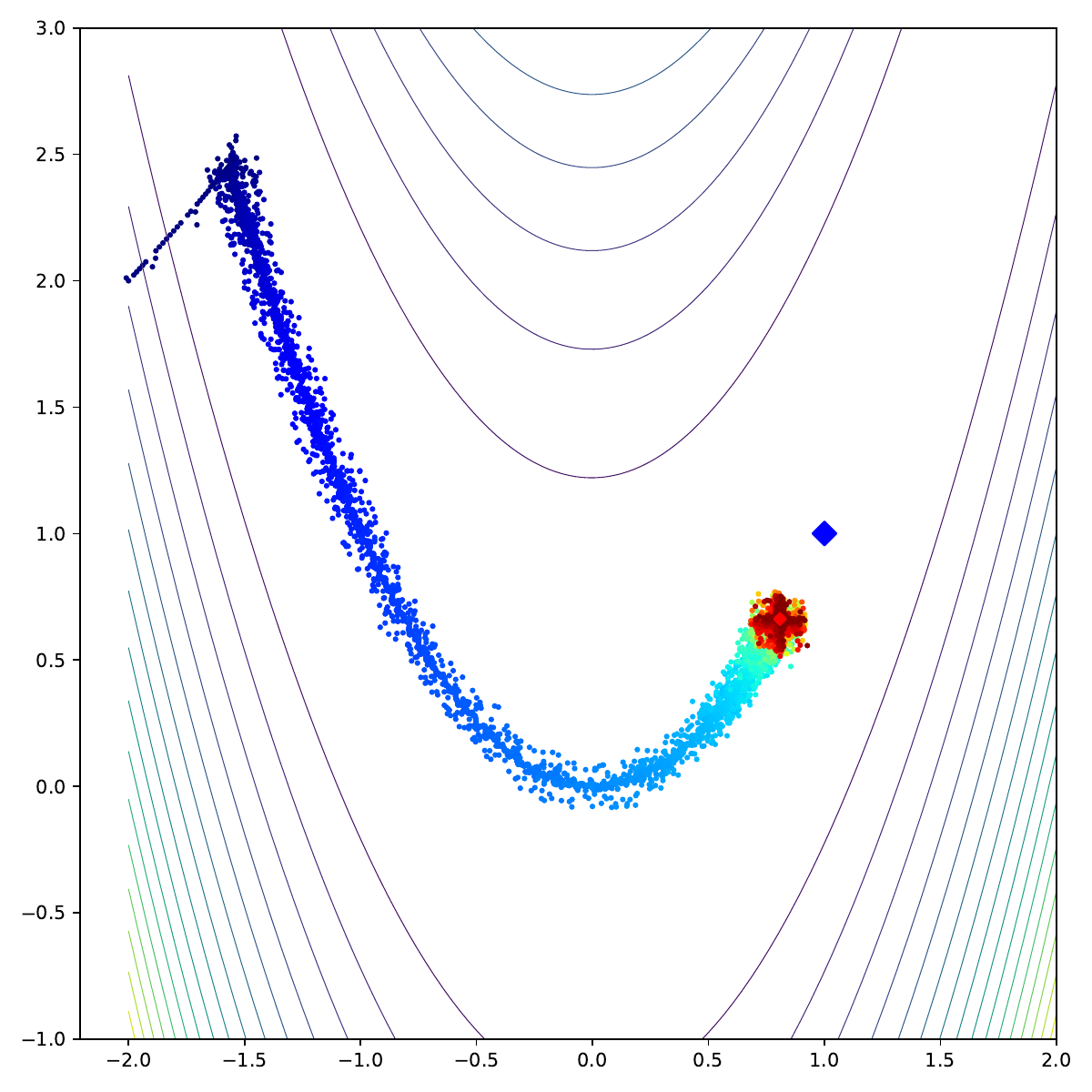}
        \subcaption{AdaBelief at 15\% noise}
        \label{fig:trajectory_rosenbrock_15_adabelief}
    \end{subfigure}
    \caption{Trajectories on Rosenbrock function}
    \label{fig:trajectory_rosenbrock}
\end{figure*}

\begin{figure*}[tb]
    \centering
    \begin{subfigure}[b]{0.225\linewidth}
        \centering
        \includegraphics[keepaspectratio=true,width=\linewidth]{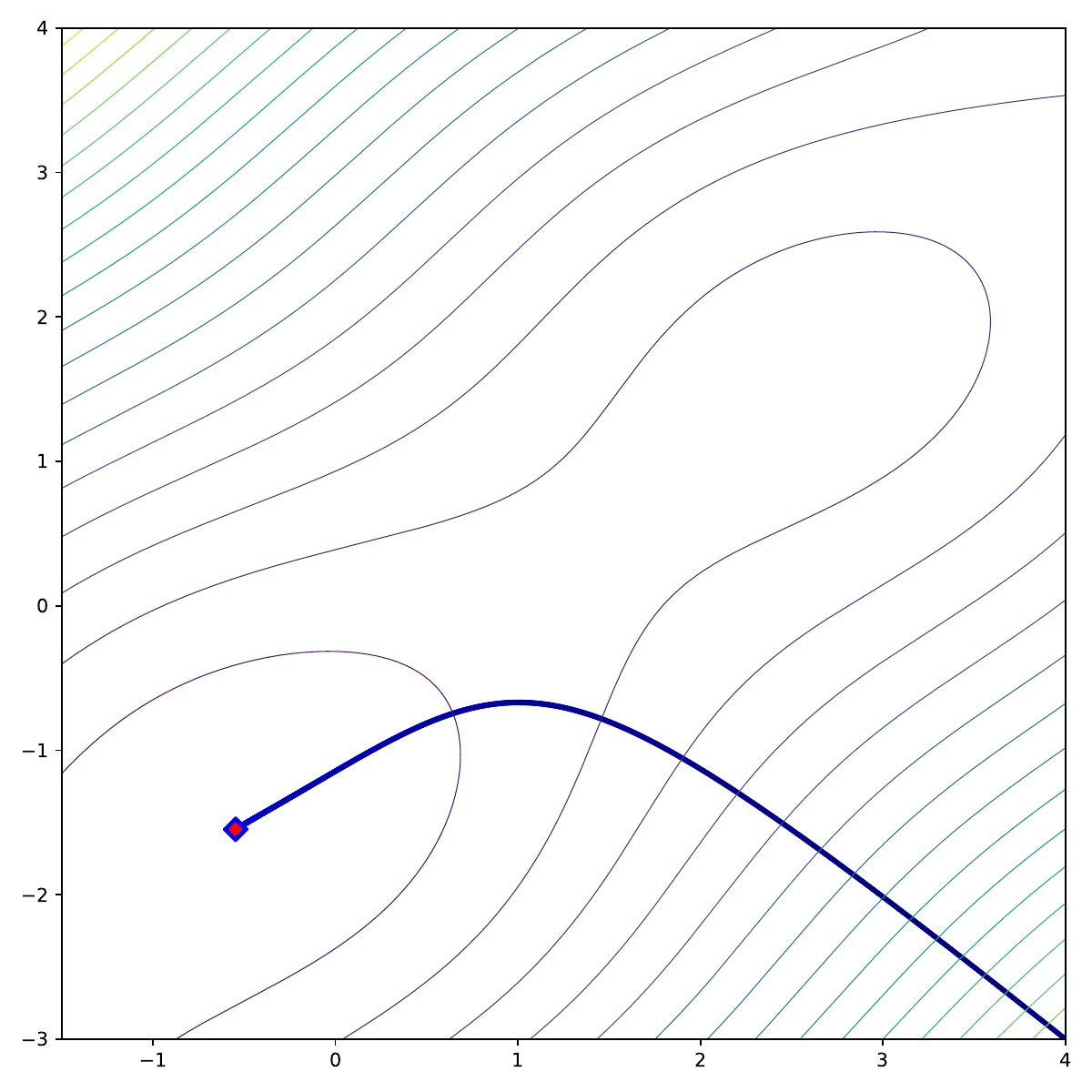}
        \subcaption{Adam at 0\% noise}
        \label{fig:trajectory_mccormick_0_adam}
    \end{subfigure}
    \begin{subfigure}[b]{0.225\linewidth}
        \centering
        \includegraphics[keepaspectratio=true,width=\linewidth]{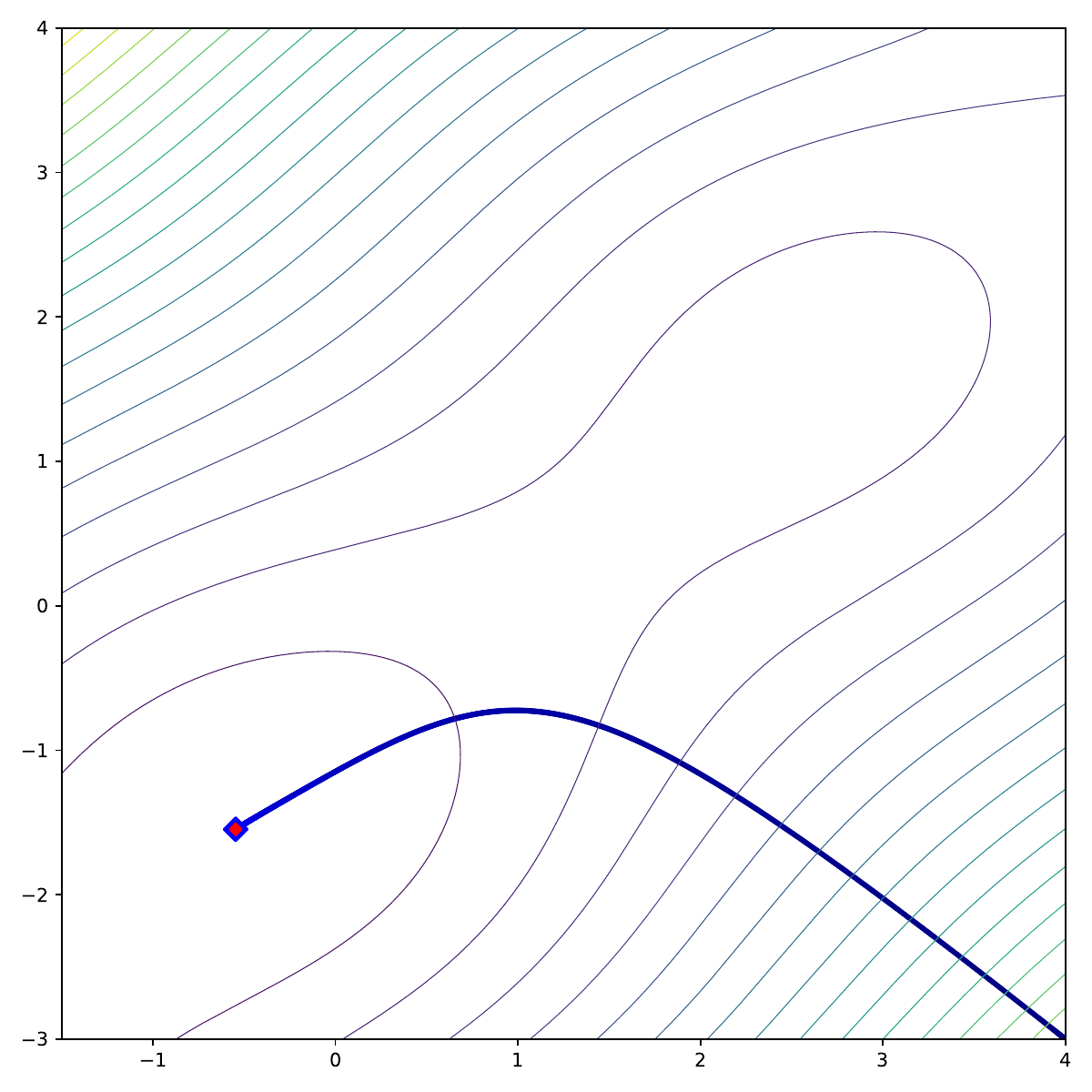}
        \subcaption{At-Adam at 0\% noise}
        \label{fig:trajectory_mccormick_0_atadam}
    \end{subfigure}
    \begin{subfigure}[b]{0.225\linewidth}
        \centering
        \includegraphics[keepaspectratio=true,width=\linewidth]{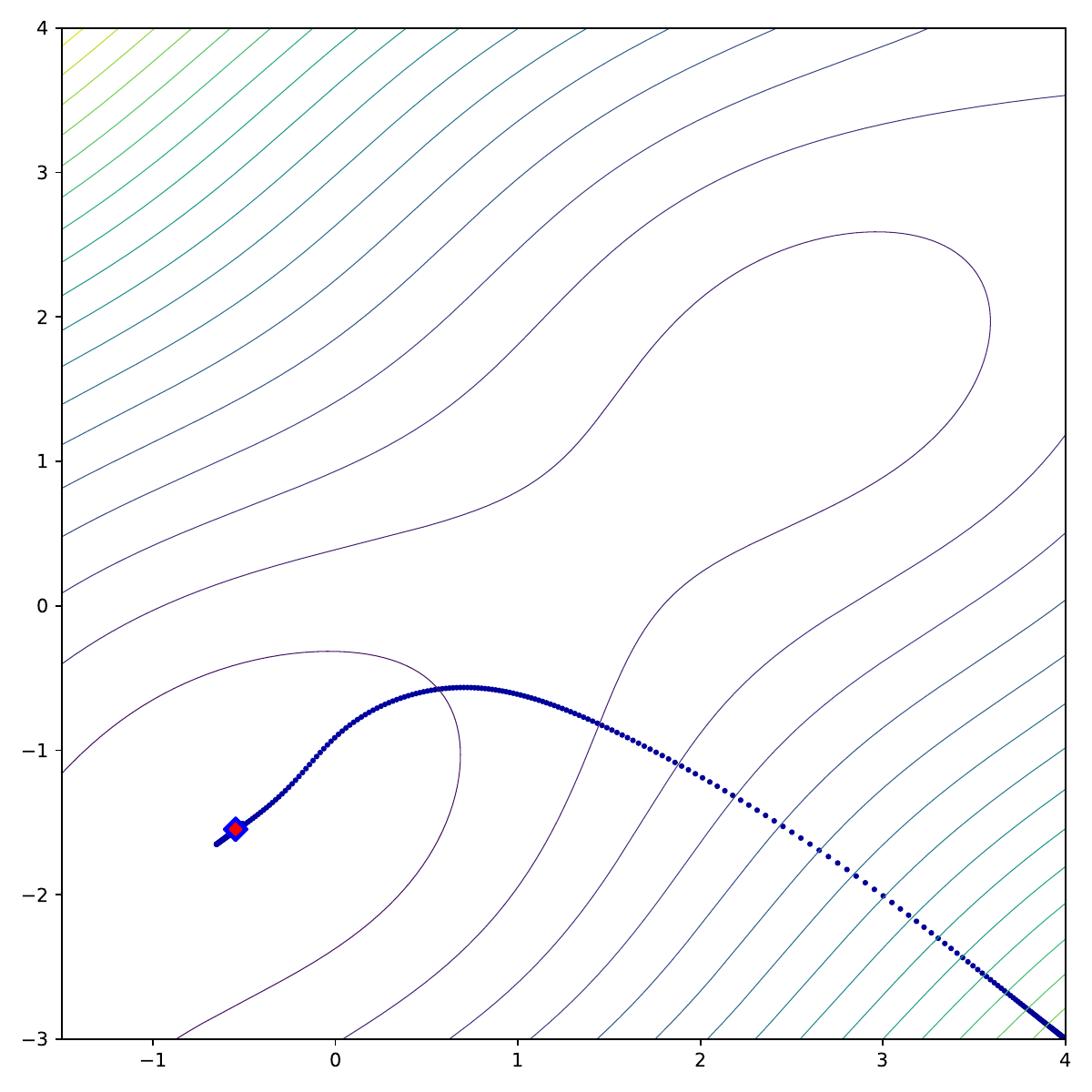}
        \subcaption{AdaTerm at 0\% noise}
        \label{fig:trajectory_mccormick_0_adaterm}
    \end{subfigure}
    \begin{subfigure}[b]{0.225\linewidth}
        \centering
        \includegraphics[keepaspectratio=true,width=\linewidth]{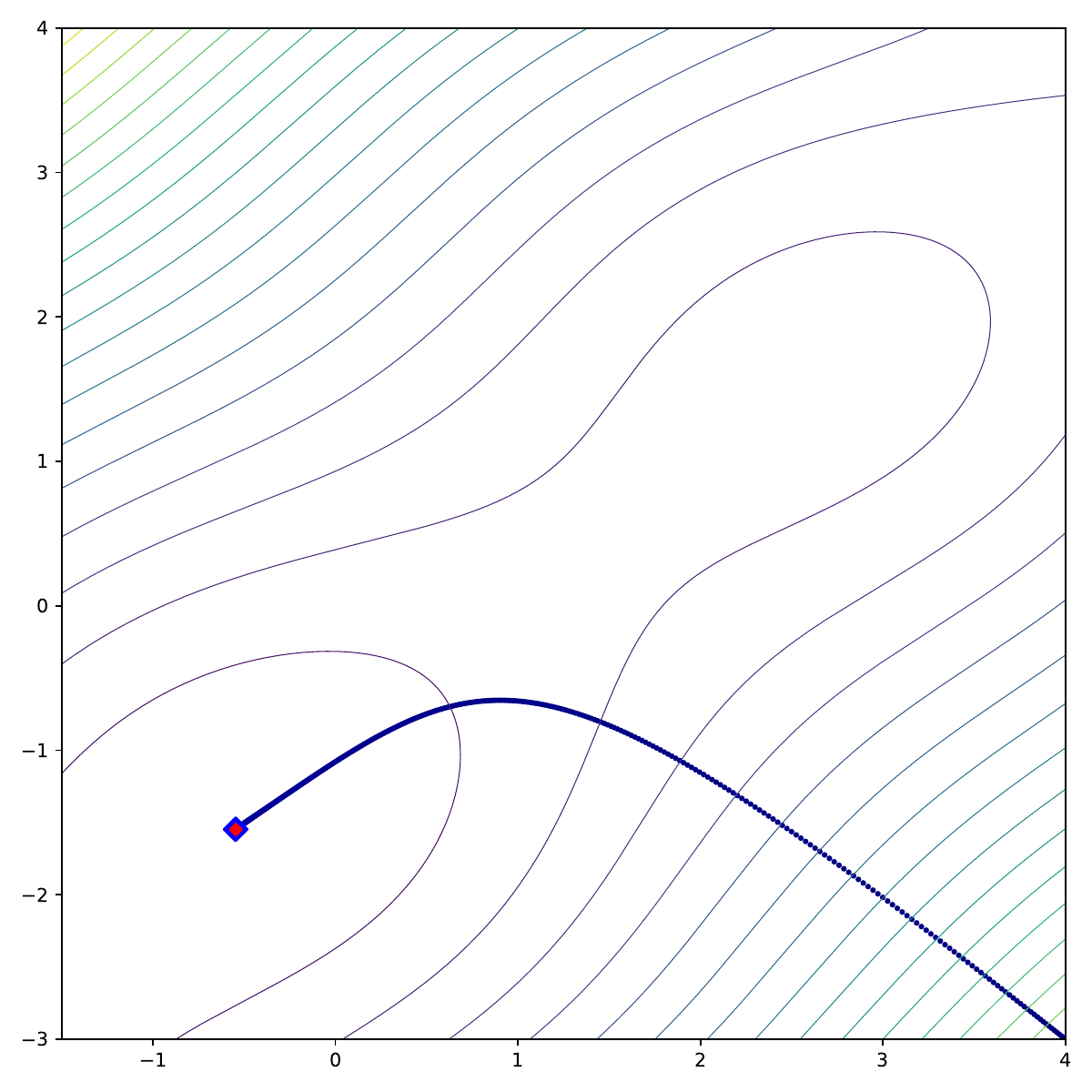}
        \subcaption{AdaBelief at 0\% noise}
        \label{fig:trajectory_mccormick_0_adabelief}
    \end{subfigure}
    \\
    \begin{subfigure}[b]{0.225\linewidth}
        \centering
        \includegraphics[keepaspectratio=true,width=\linewidth]{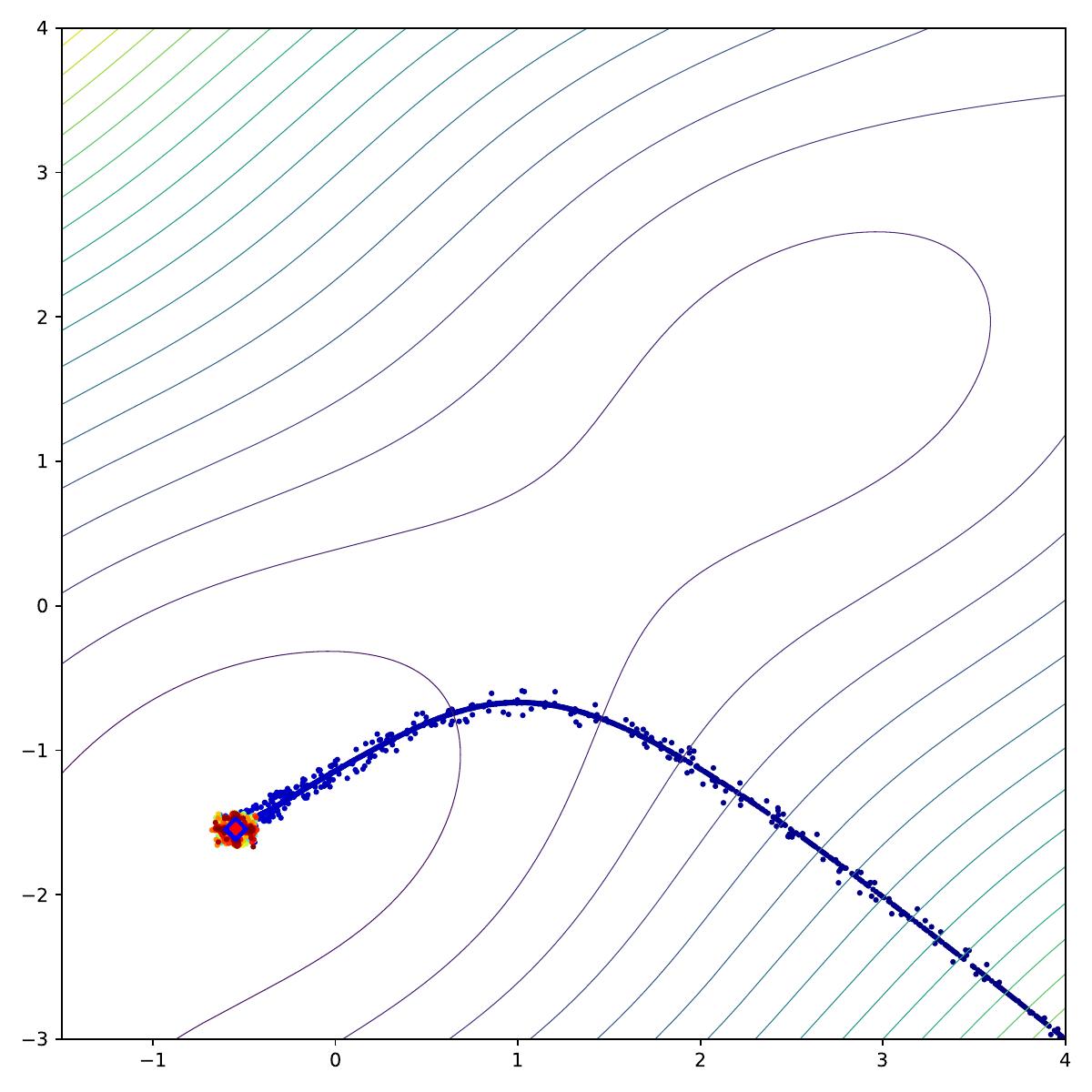}
        \subcaption{Adam at 15\% noise}
        \label{fig:trajectory_mccormick_15_adam}
    \end{subfigure}
    \begin{subfigure}[b]{0.225\linewidth}
        \centering
        \includegraphics[keepaspectratio=true,width=\linewidth]{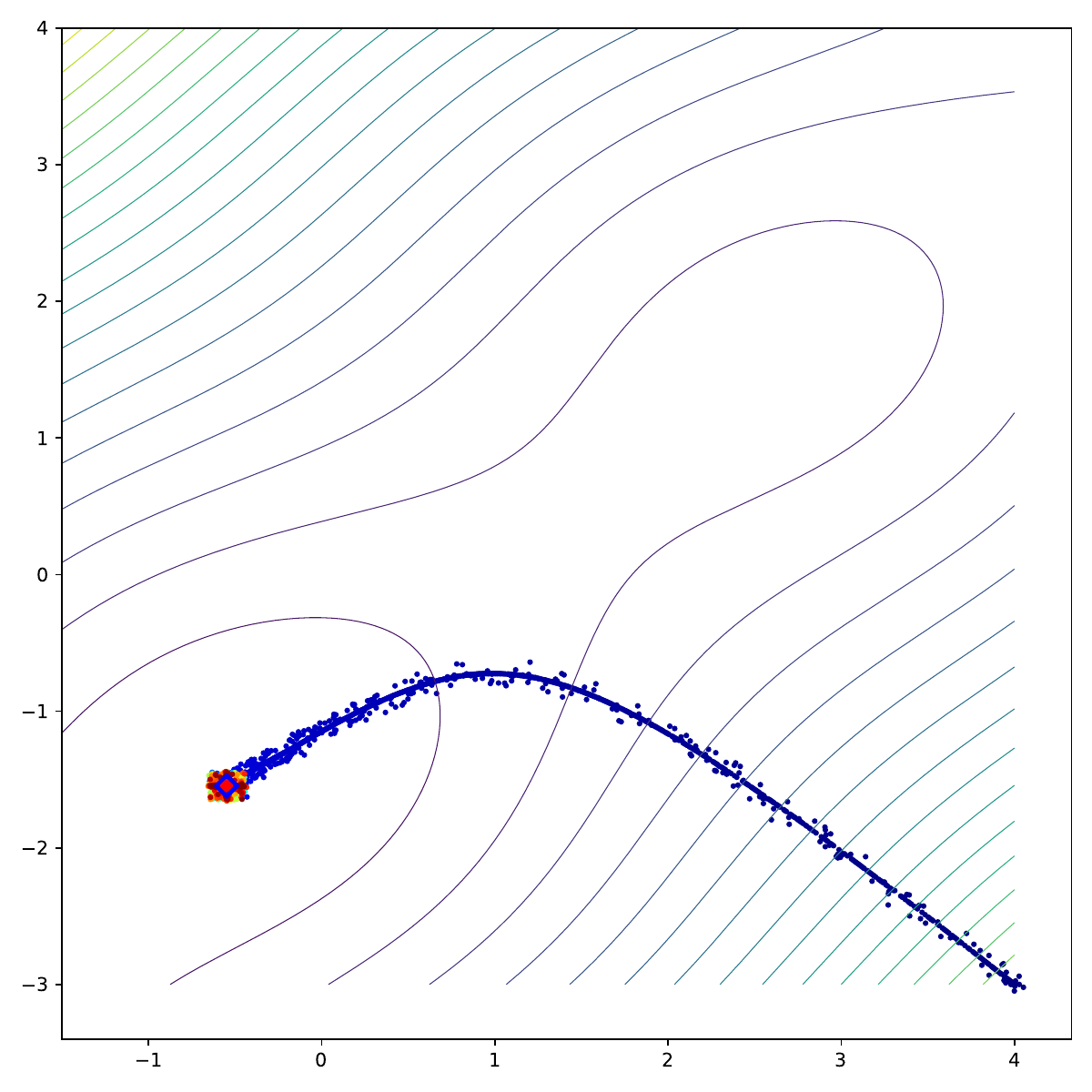}
        \subcaption{At-Adam at 15\% noise}
        \label{fig:trajectory_mccormick_15_atadam}
    \end{subfigure}
    \begin{subfigure}[b]{0.225\linewidth}
        \centering
        \includegraphics[keepaspectratio=true,width=\linewidth]{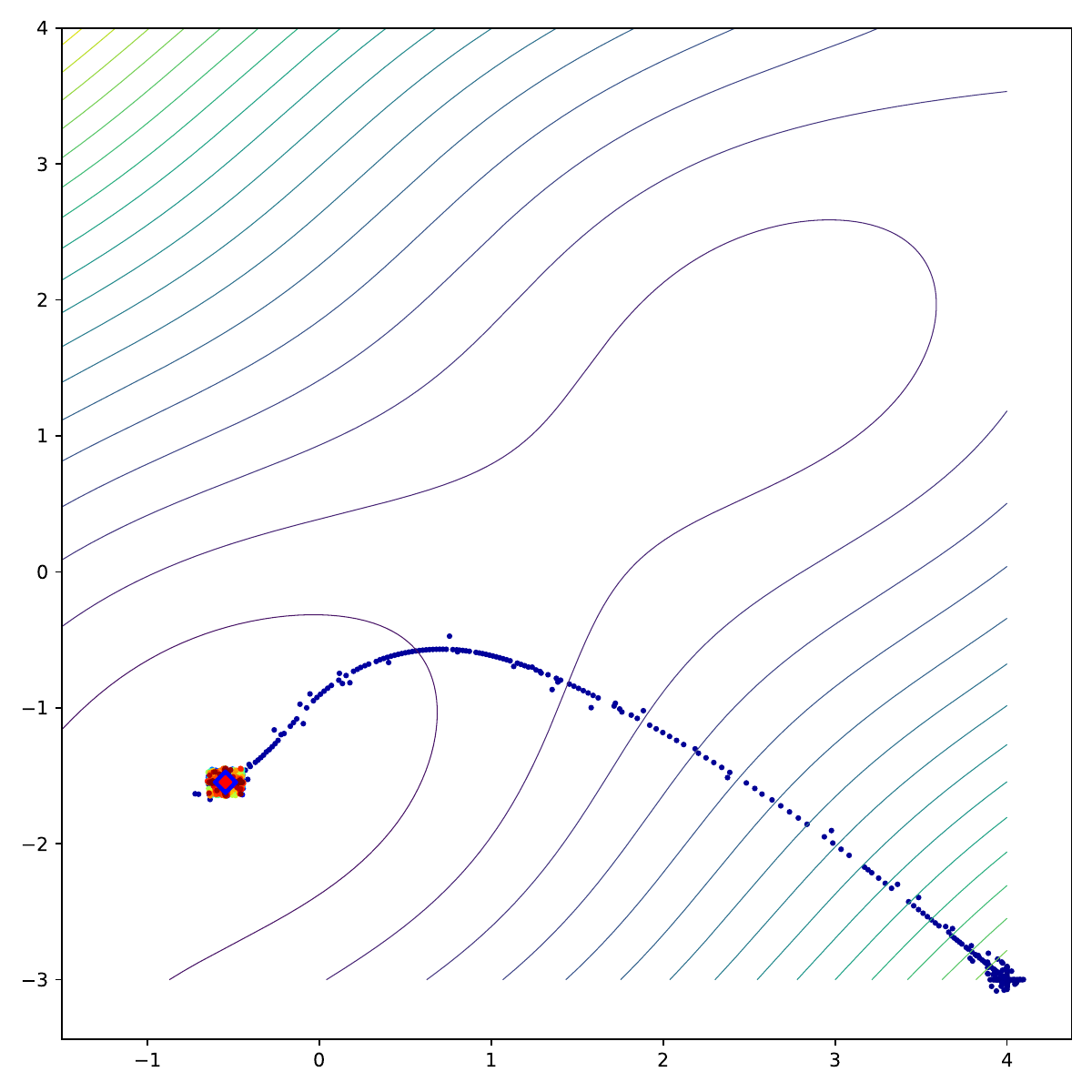}
        \subcaption{AdaTerm at 15\% noise}
        \label{fig:trajectory_mccormick_15_adaterm}
    \end{subfigure}
    \begin{subfigure}[b]{0.225\linewidth}
        \centering
        \includegraphics[keepaspectratio=true,width=\linewidth]{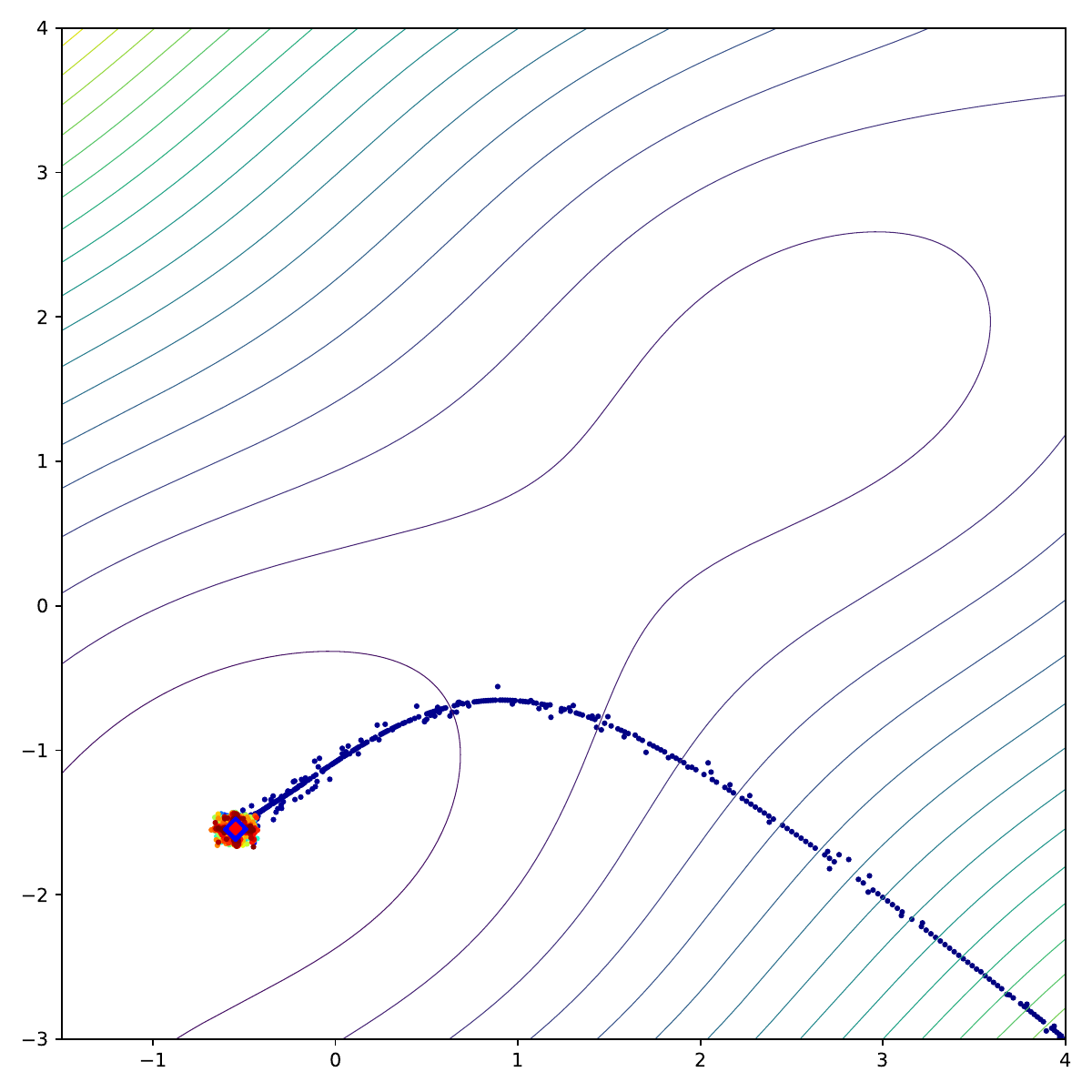}
        \subcaption{AdaBelief at 15\% noise}
        \label{fig:trajectory_mccormick_15_adabelief}
    \end{subfigure}
    \caption{Trajectories on McCormick function}
    \label{fig:trajectory_mccormick}
\end{figure*}

\begin{figure*}[tb]
    \centering
    \begin{subfigure}[b]{0.225\linewidth}
        \centering
        \includegraphics[keepaspectratio=true,width=\linewidth]{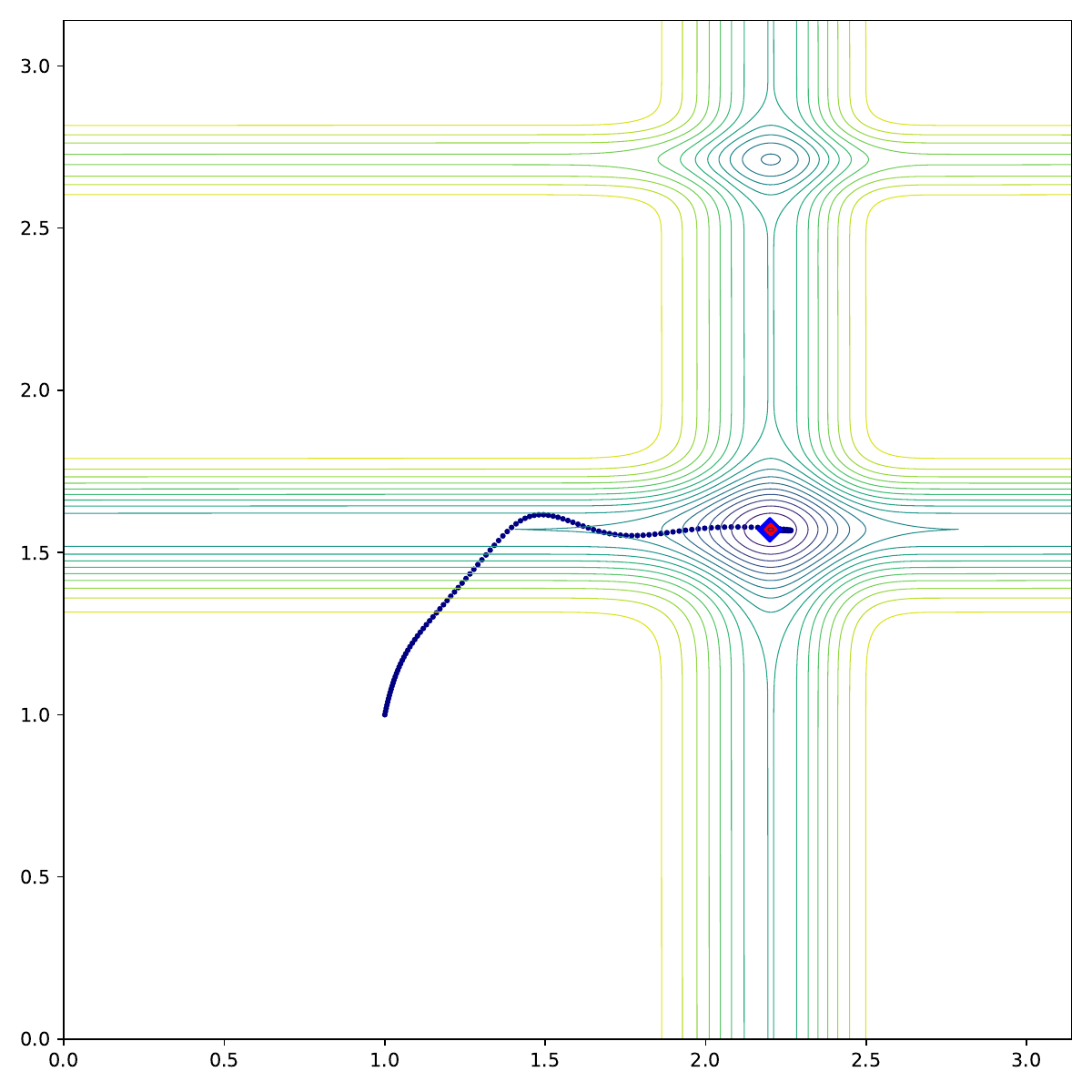}
        \subcaption{Adam at 0\% noise}
        \label{fig:trajectory_michalewicz_0_adam}
    \end{subfigure}
    \begin{subfigure}[b]{0.225\linewidth}
        \centering
        \includegraphics[keepaspectratio=true,width=\linewidth]{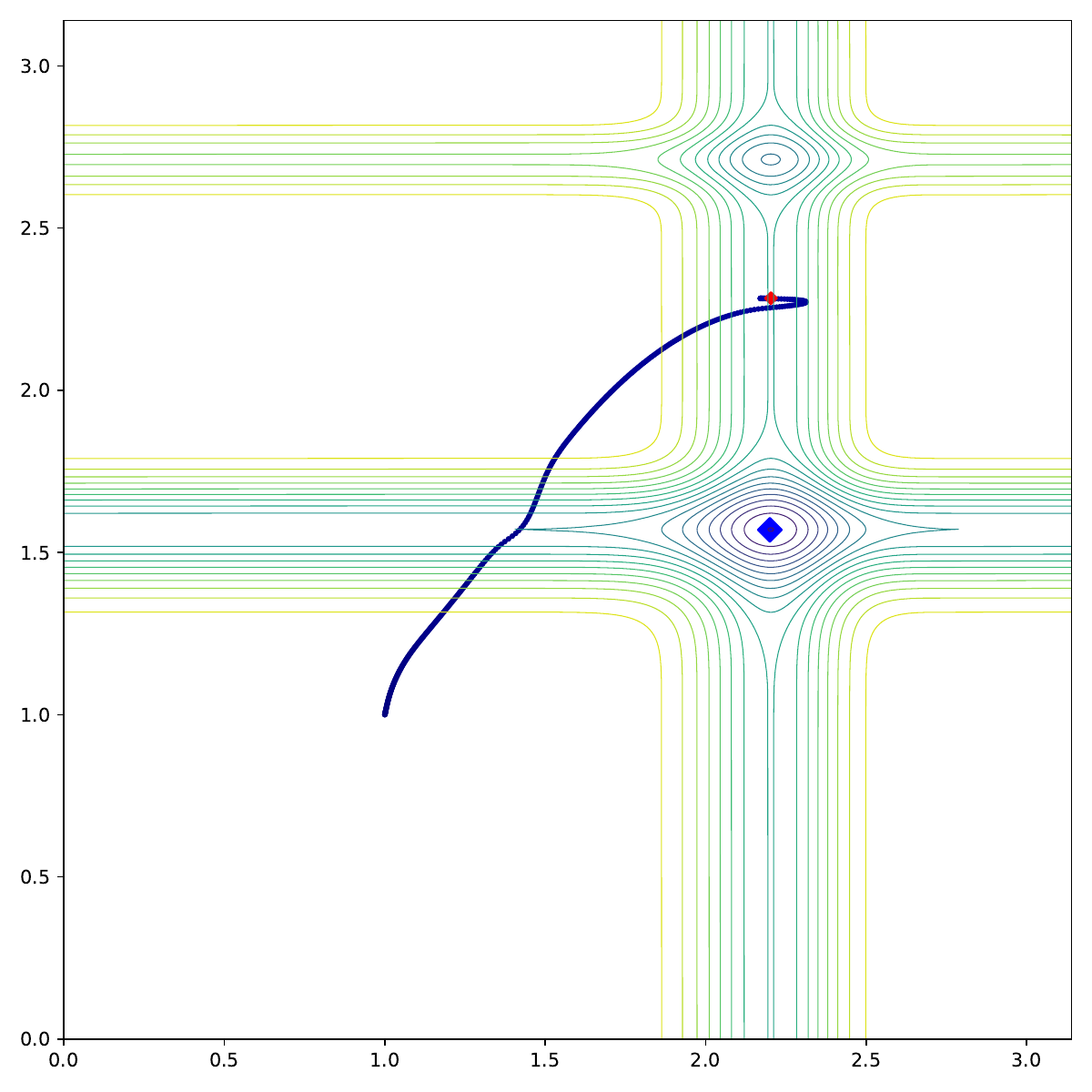}
        \subcaption{At-Adam at 0\% noise}
        \label{fig:trajectory_michalewicz_0_atadam}
    \end{subfigure}
    \begin{subfigure}[b]{0.225\linewidth}
        \centering
        \includegraphics[keepaspectratio=true,width=\linewidth]{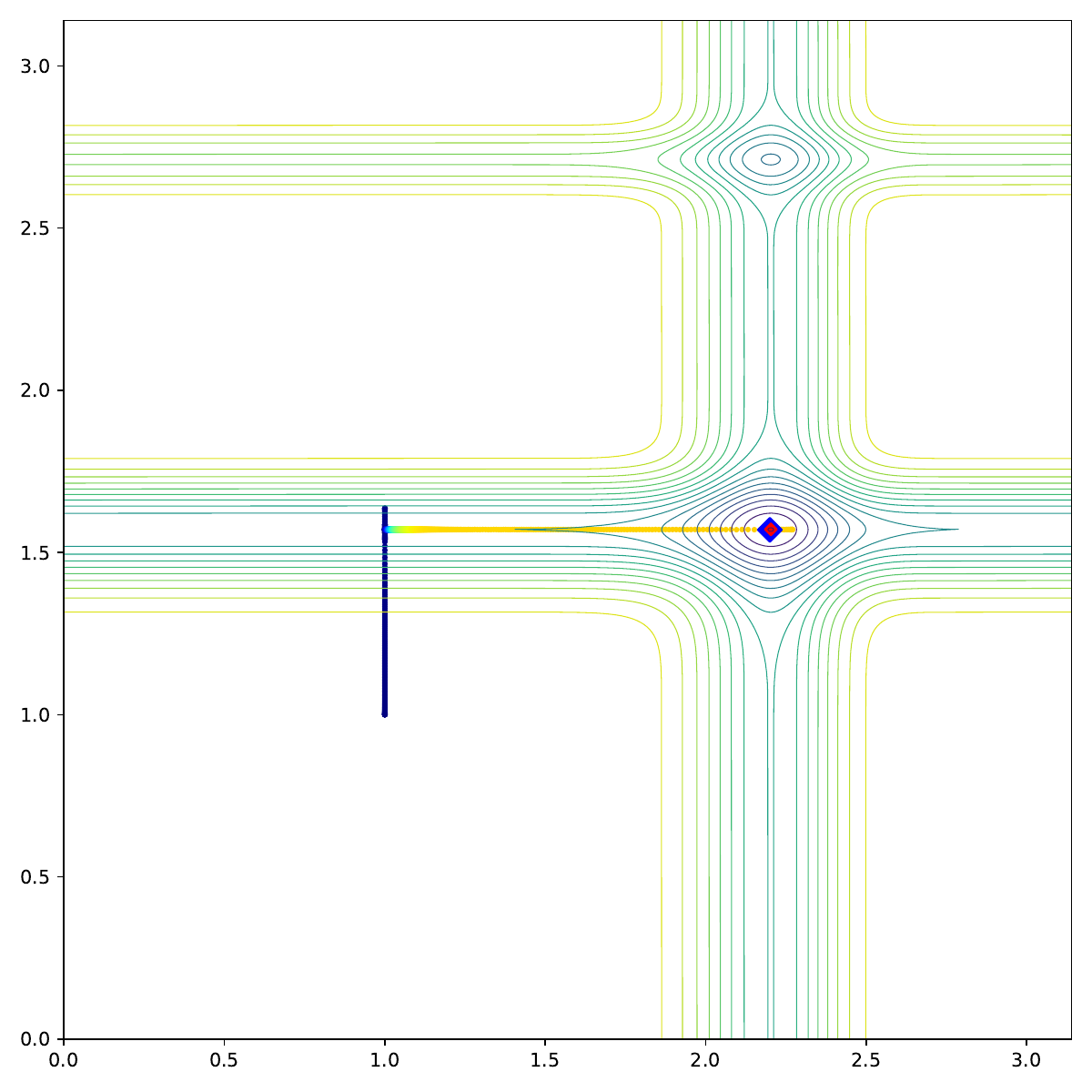}
        \subcaption{AdaTerm at 0\% noise}
        \label{fig:trajectory_michalewicz_0_adaterm}
    \end{subfigure}
    \begin{subfigure}[b]{0.225\linewidth}
        \centering
        \includegraphics[keepaspectratio=true,width=\linewidth]{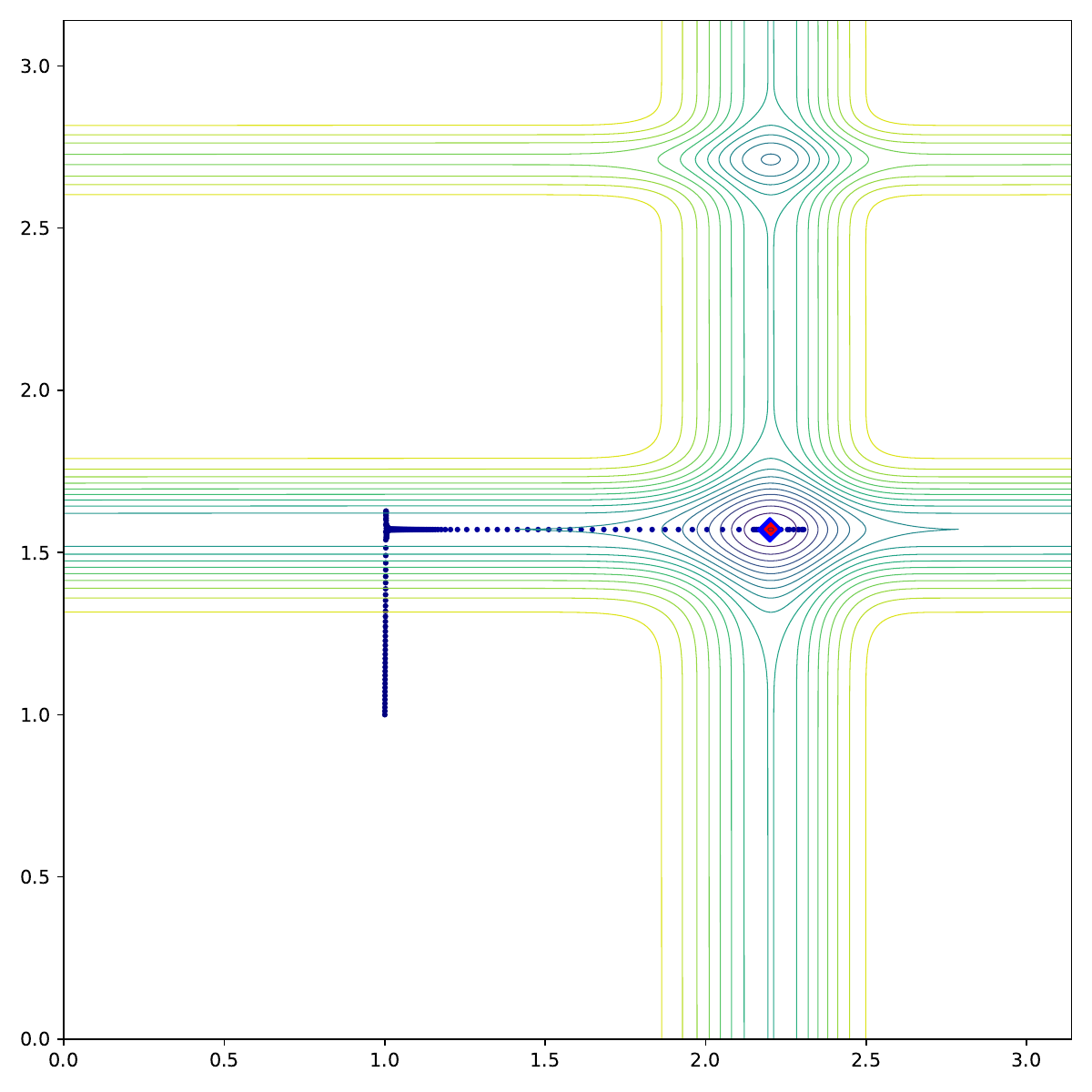}
        \subcaption{AdaBelief at 0\% noise}
        \label{fig:trajectory_michalewicz_0_adabelief}
    \end{subfigure}
    \\
    \begin{subfigure}[b]{0.225\linewidth}
        \centering
        \includegraphics[keepaspectratio=true,width=\linewidth]{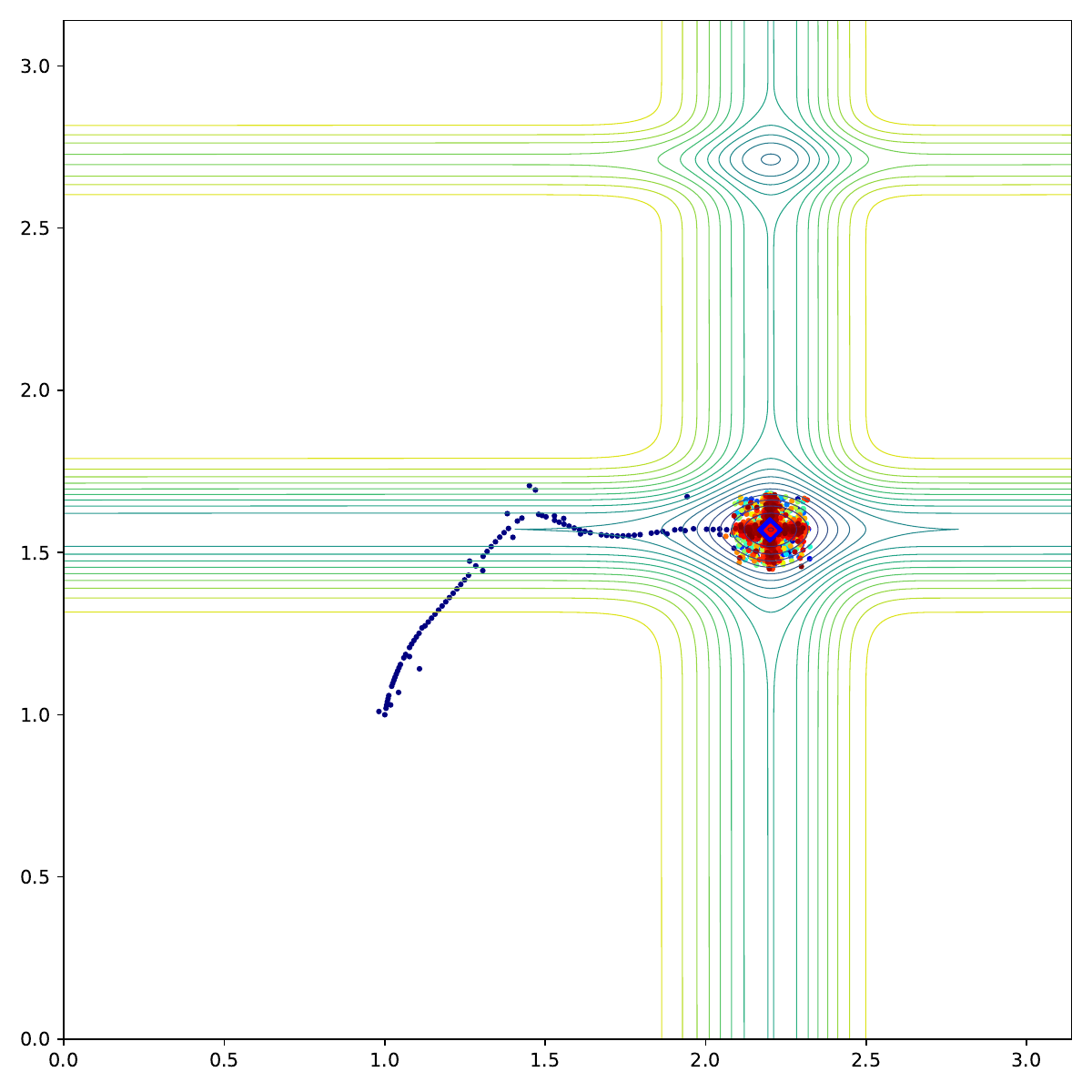}
        \subcaption{Adam at 15\% noise}
        \label{fig:trajectory_michalewicz_15_adam}
    \end{subfigure}
    \begin{subfigure}[b]{0.225\linewidth}
        \centering
        \includegraphics[keepaspectratio=true,width=\linewidth]{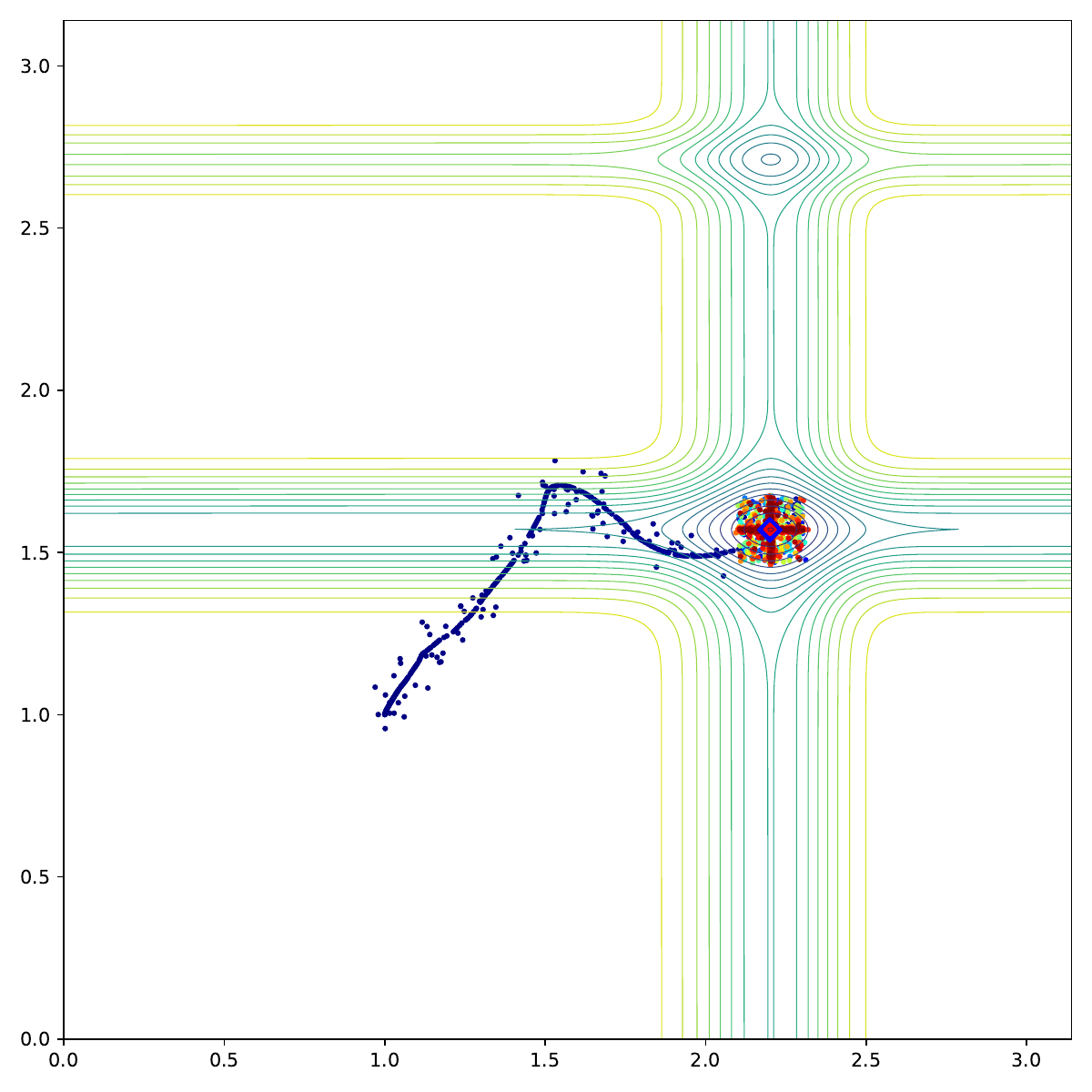}
        \subcaption{At-Adam at 15\% noise}
        \label{fig:trajectory_michalewicz_15_atadam}
    \end{subfigure}
    \begin{subfigure}[b]{0.225\linewidth}
        \centering
        \includegraphics[keepaspectratio=true,width=\linewidth]{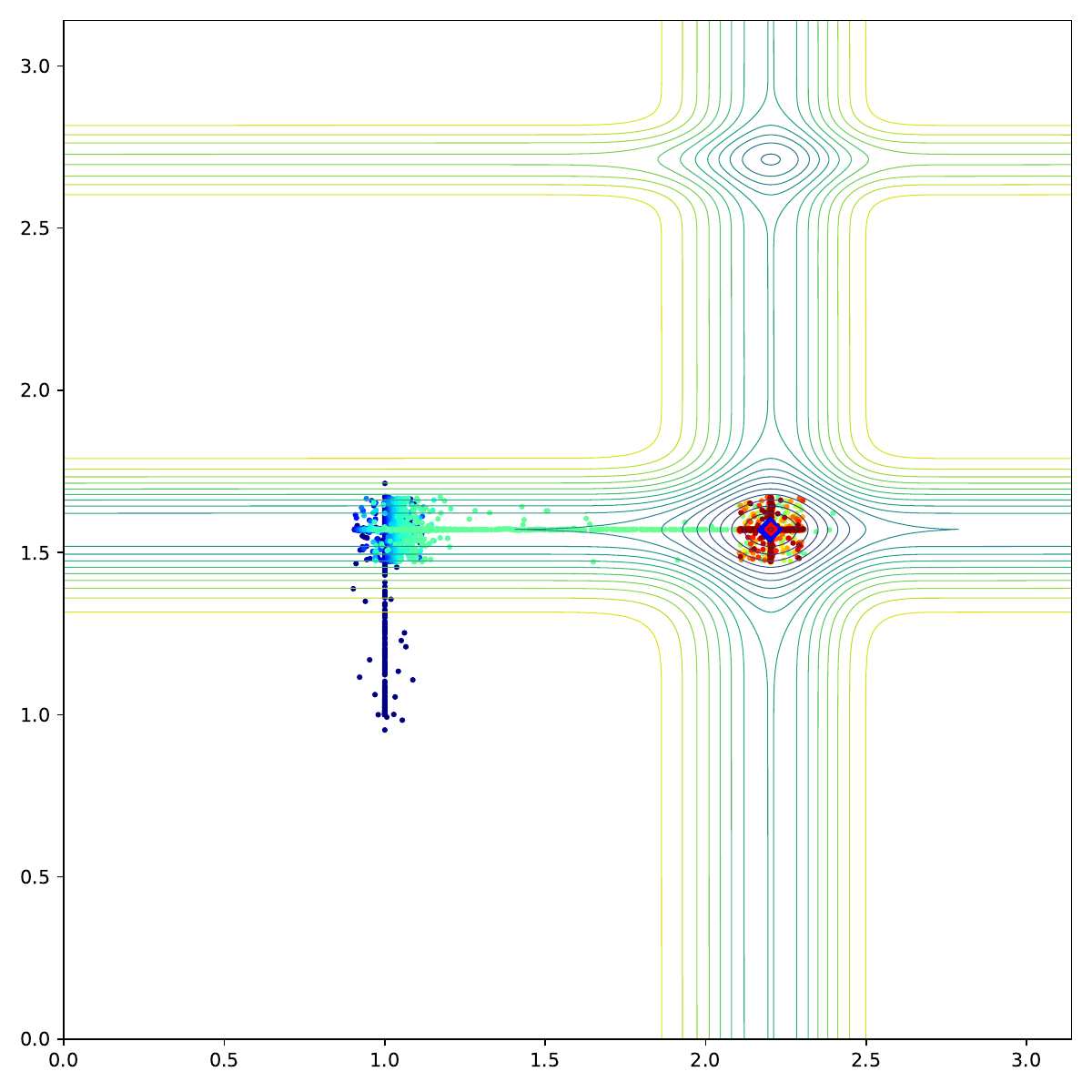}
        \subcaption{AdaTerm at 15\% noise}
        \label{fig:trajectory_michalewicz_15_adaterm}
    \end{subfigure}
    \begin{subfigure}[b]{0.225\linewidth}
        \centering
        \includegraphics[keepaspectratio=true,width=\linewidth]{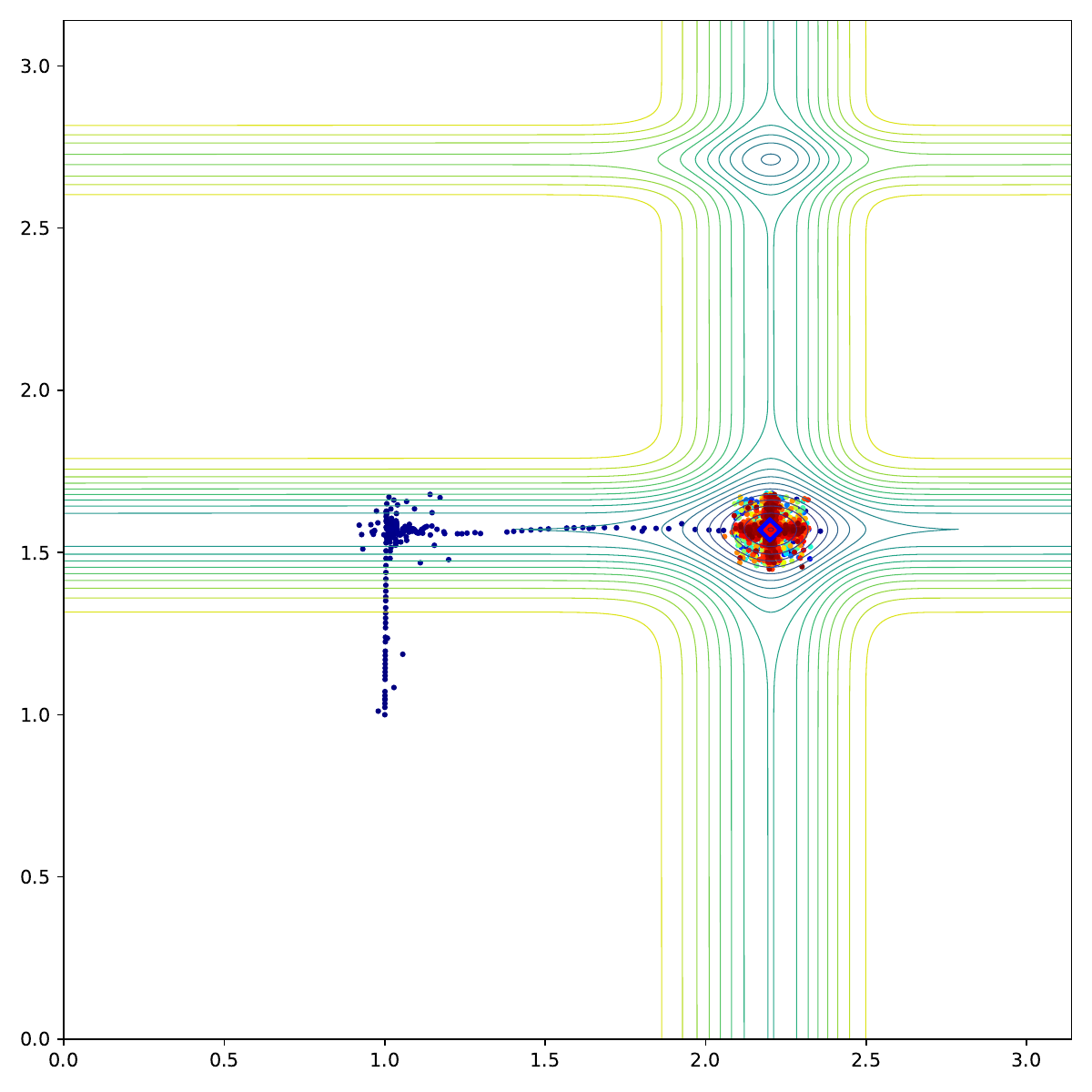}
        \subcaption{AdaBelief at 15\% noise}
        \label{fig:trajectory_michalewicz_15_adabelief}
    \end{subfigure}
    \caption{Trajectories on Michalewicz function}
    \label{fig:trajectory_michalewicz}
\end{figure*}

Trajectories of the convergence process are illustrated in Fig.~\ref{fig:trajectory_rosenbrock}--\ref{fig:trajectory_michalewicz}.
Note that the color of each point indicates the elapsed time, changing from blue to red.
AdaTerm was less likely to update in the incorrect direction caused by noise than the others, indicating stable updating. 
In addition, we observe higher update speed of AdaTerm compared to others.
This is probably because $\beta < \beta_2$ could rapidly adapt to small gradients in the saddle area, although this caused an overshoot around the optimal point in Fig.~\ref{fig:trajectory_mccormick_0_adaterm}, resulting in the larger error norm than the others in Fig.~\ref{fig:comp_norm}.

A remarkable result was obtained for Michalewicz function (see Fig.~\ref{fig:trajectory_michalewicz_0_adaterm}).
Specifically, AdaTerm initially moved only in the $y$-direction, as in AdaBelief, and subsequently paused at steep gradients in the $x$-direction.
This is because the steep gradients toward the optimum were considered as noise in certain cases.
In fact, with the elapse of time, AdaTerm judged the steep gradients as non-noise.
Gradually, as $\tilde{\nu}$ increased, the update was resumed, finally reaching the optimal point.
Furthermore, At-Adam was unable to adapt $\nu$ to the steep gradients, and the point moved to avoid the optimal point (see Fig.~\ref{fig:trajectory_michalewicz_0_atadam}).
However, when noise was added, the steep gradients started to be utilized with the aid of the noise, and the optimal point was finally reached.

\section{Learning setups}
\label{apdx:learning}

The models for learning the four benchmark problems are implemented using PyTorch~\cite{paszke2017automatic} with the respective setups, as summarized in Table~\ref{tab:config}.
The parameters that are not listed in the table are basically set at the recommended default values.
Note that the learning rate for RL was smaller than the ones for the other problems to avoid being trapped in local solutions before effective samples are collected.
In addition, the batch size was set to 32, such that the noise is more likely to affect the gradients. However, for the classification problem only, the batch size was increased to 256, which is within the general range, in response to the large learning time cost of this task.

For the classification problem, ResNet18~\cite{he2016deep} was employed as the network model.
The official PyTorch implementation was employed; however, to account for the size difference of the input image, the first layer of convolution was changed to 1 with a kernel size of 3 and a stride of 1; furthermore, the subsequent MaxPool layer was excluded.

For the prediction problem, a gated recurrent unit (GRU)~\cite{chung2014empirical} was employed for constructing a network model consisting of two serial hidden layers, with each including 128 GRU units.
The output layer generates the means and standard deviations of the multivariate diagonal Gaussian distribution used to sample the predicted state.
Note that, since extending the number of prediction steps delays the learning process, we set the number of epochs to 100 for single-step prediction and 300 for 30-step prediction.

For the RL problem, five fully-connected layers with 100 neurons for each were implemented as hidden layers.
Each activation function was an unlearned LayerNorm~\cite{ba2016layer,xu2019understanding} and Swish function~\cite{elfwing2018sigmoid}.
The output is a multivariate diagonal Student's t-distribution used to improve the efficiency of the exploration~\cite{kobayashi2019student}.

For the policy distillation problem, two fully-connected layers with only 32 neurons for each were implemented as the hidden layers.
This structure is clearly smaller than that for the RL problem described above.
The only activation function used for the different layers is the Tanh function, with reference to the fact that this design has been reported to have sufficient expressive capability~\cite{de2021approximation}.
The output layer generates the means and standard deviations of the multivariate diagonal Gaussian distribution used to capture the stochastic teacher policy.

\begin{table}[ht]
    \caption{Learning setups}
    \label{tab:config}
    \centering
    \begin{tabular}{l cccc}
        \hline\hline
        Parameter & Classification & Prediction & RL & Distillation
        \\
        \hline
        Learning rate
        & 1e-3
        & 1e-3
        & 1e-4
        & 1e-3
        \\
        Batch size
        & 256
        & 32
        & 32
        & 32
        \\
        \#Epoch
        & 100
        & 100/300
        & 2000
        & 200
        \\
        Label smoothing
        & 0.2
        & -
        & -
        & -
        \\
        Truncation
        & -
        & 30
        & -
        & -
        \\
        \#Batch/\#Replay buffer
        & -
        & -
        & 128/1e+4
        & -
        \\
        Weight decay
        & -
        & -
        & -
        & 1e-4
        \\
        \hline\hline
    \end{tabular}
\end{table}

\section{Ablation study}
\label{apdx:ablation}

For the ablation test of AdaTerm, we test the following three conditions:
$\Delta \tilde{\nu} = 0$, called no adaptiveness;
$\underline{\tilde{\nu}} = \infty$, called no robustness;
and $\tilde{\nu}_0 = 100$, called large init.
All other conditions are the same as the experiments in the main text.

The test results are summarized in Table~\ref{tab:ablation}.
As expected, the cases with no adaptiveness outperformed the cases with no robustness for the problems with high noise effects, and vice versa.
As observed from the results of the 30-step prediction and the policy distillation, the optimal solution may lie somewhere in the middle rather than at the two extremes, and the normal AdaTerm has been successful in finding it.
In the case where $\tilde{\nu}_0$ is increased and the noise robustness is inferior at the beginning of training, the performance was worse than the normal case where $\tilde{\nu}_0 = \underline{\tilde{\nu}} + \epsilon$ and the noise robustness is maximized at the beginning, except for the classification problem.
This tendency suggests that even if $\tilde{\nu}$ is adjusted to gain the appropriate noise robustness after optimization without considering the effects of noise, the performance would be prone to local solution traps.
For the classification problem only, the size of the network architecture was larger than that for the other problems, and its redundancy allowed the classifier to escape from the local solutions.
In such a case, using more gradients from the beginning resulted in performance improvement.

\begin{table}[ht]
    \caption{Ablation results}
    \label{tab:ablation}
    \centering
    \begin{tabular}{l cc cc cc cc}
        \hline\hline
         & \multicolumn{2}{c}{Classification}
         & \multicolumn{2}{c}{Prediction}
         & \multicolumn{2}{c}{RL}
         & \multicolumn{2}{c}{Distillation}
         \\
         & \multicolumn{2}{c}{Accuracy}
         & \multicolumn{2}{c}{MSE at final prediction}
         & \multicolumn{2}{c}{The sum of rewards}
         & \multicolumn{2}{c}{The sum of rewards}
         \\
        Method & 0~\% & 10~\% & 1 step & 30 steps & Hopper & Ant & w/o amateur & w/ amateur
        \\
        \hline
        AdaTerm
        & 0.7315
        & 0.6815
        & 0.0335
        & \textcolor{orange}{\textbf{1.0016}}
        & 1550.25
        & 2021.37
        & \textcolor{orange}{\textbf{1770.17}}
        & 1411.02
        \\

        & (3.66e-3)
        & (4.46e-3)
        & (3.09e-4)
        & (2.31e-1)
        & (5.88e+2)
        & (3.87e+2)
        & (2.17e+2)
        & (1.92e+2)
        \\
        \hline
        No adaptiveness
        & 0.7330
        & \textcolor{orange}{\textbf{0.6840}}
        & 0.0336
        & 1.0718
        & 496.75
        & \textcolor{orange}{\textbf{2192.21}}
        & 1701.63
        & 1437.18
        \\

        & (3.20e-3)
        & (4.20e-3)
        & (4.30e-4)
        & (2.75e-1)
        & (6.19e+2)
        & (4.18e+2)
        & (2.32e+2)
        & (2.26e+2)
        \\
        No robustness
        & 0.7330
        & 0.6810
        & \textcolor{orange}{\textbf{0.0328}}
        & 1.2553
        & 1666.30
        & 1412.93
        & 1625.05
        & 1403.92
        \\

        & (3.20e-3)
        & (3.80e-3)
        & (4.14e-4)
        & (2.09e-1)
        & (6.57e+2)
        & (6.52e+2)
        & (2.61e+2)
        & (2.56e+2)
        \\
        Large init
        & \textcolor{orange}{\textbf{0.7350}}
        & 0.6830
        & 0.0332
        & 1.2966
        & \textcolor{orange}{\textbf{1750.63}}
        & 1902.63
        & 1655.40
        & \textcolor{orange}{\textbf{1442.24}}
        \\

        & (3.50e-3)
        & (3.90e-3)
        & (4.47e-4)
        & (2.36e-1)
        & (5.47e+2)
        & (3.87e+2)
        & (1.71e+2)
        & (2.84e+2)
        \\
        \hline\hline
        t-AdaBelief
        & 0.7231
        & 0.6827
        & \textcolor{blue}{\textbf{0.0318}}
        & 1.1517
        & 1104.87
        & 2185.45
        & 1639.93
        & 1303.19
        \\

        & (4.23e-3)
        & (4.02e-3)
        & (4.42e-4)
        & (1.17e-1)
        & (4.48e+2)
        & (4.83e+2)
        & (3.19e+2)
        & (2.58e+2)
        \\
        \hline
        At-AdaBelief
        & 0.7223
        & 0.6815
        & \textcolor{blue}{\textbf{0.0318}}
        & 1.1306
        & 1214.47
        & \textcolor{blue}{\textbf{2196.84}}
        & 1671.65
        & 1325.70
        \\

        & (4.50e-3)
        & (3.46e-3)
        & (5.02e-4)
        & (1.17e-1)
        & (3.77e+2)
        & (2.29e+2)
        & (3.32e+2)
        & (2.72e+2)
        \\
        \hline\hline
    \end{tabular}
\end{table}

\section{Analysis of the batch size's effect on robustness}
\label{apdx:batch_size_effect}

As previously proven~\cite{lee2018directional}, the batch size has an impact on the stochastic gradient's noise strength.
Furthermore, larger batch sizes would aid mitigating the effect of corrupted data points by drowning them in a majority of good points inside the mini batch.
However, in typical machine learning tasks, given a certain dataset, an increase in the batch size also implies a decrease in the number of updates (and therefore also a decrease in the number of gradients observed).
It has also been reported that the large batch size degrades generalization performance~\cite{keskar2016large}.
To replicate such trade-off effect and to study its influence on robustness, we use the regression task where the number of samples is fixed at $40000$ and subsequently split into batches of different sizes $\in \left\{ 8, 16, 32, 64 \right\}$.

The results show that for all optimizers in noiseless scenarios, the performance degrades when increasing the batch size, consistent with the results of previous studies, although this is more pronounced for AdaTerm and (A)t-Adam.
Specifically, the performance of the robust optimizers degraded as the batch size increased, as expected, since the detection of the effect of aberrant points inside a majority of good points is difficult. 
Nevertheless, this majority of good points could not mitigate the effect of noise, as can be confirmed in the fact that the non-robust Adam and AdaBelief remained sensitive to noise.

\begin{figure*}[ht]
    \centering
    \begin{subfigure}[b]{0.23\linewidth}
        \centering
        \includegraphics[keepaspectratio=true,width=\linewidth]{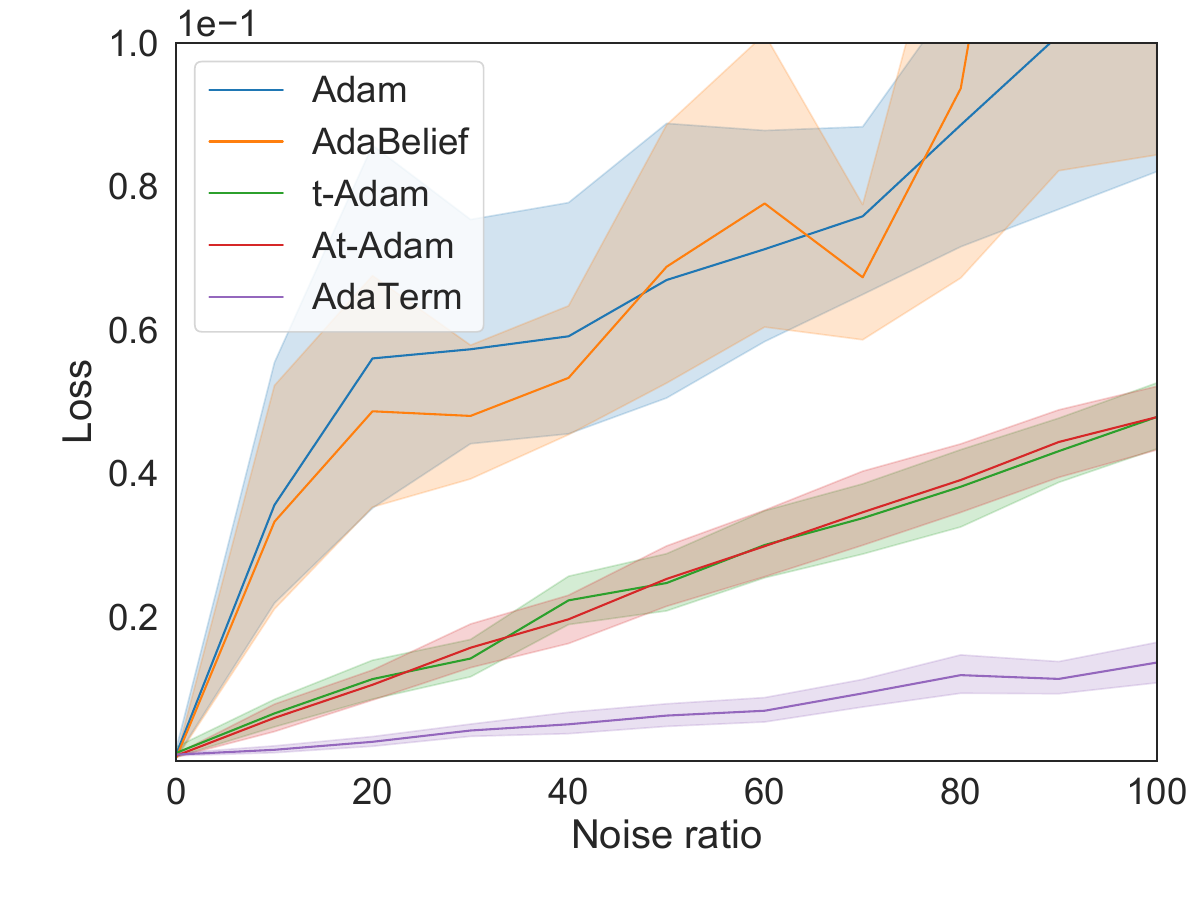}
        \subcaption{Batch size $= 8$}
        \label{fig:loss_nbatch8}
    \end{subfigure}
    \begin{subfigure}[b]{0.23\linewidth}
        \centering
        \includegraphics[keepaspectratio=true,width=\linewidth]{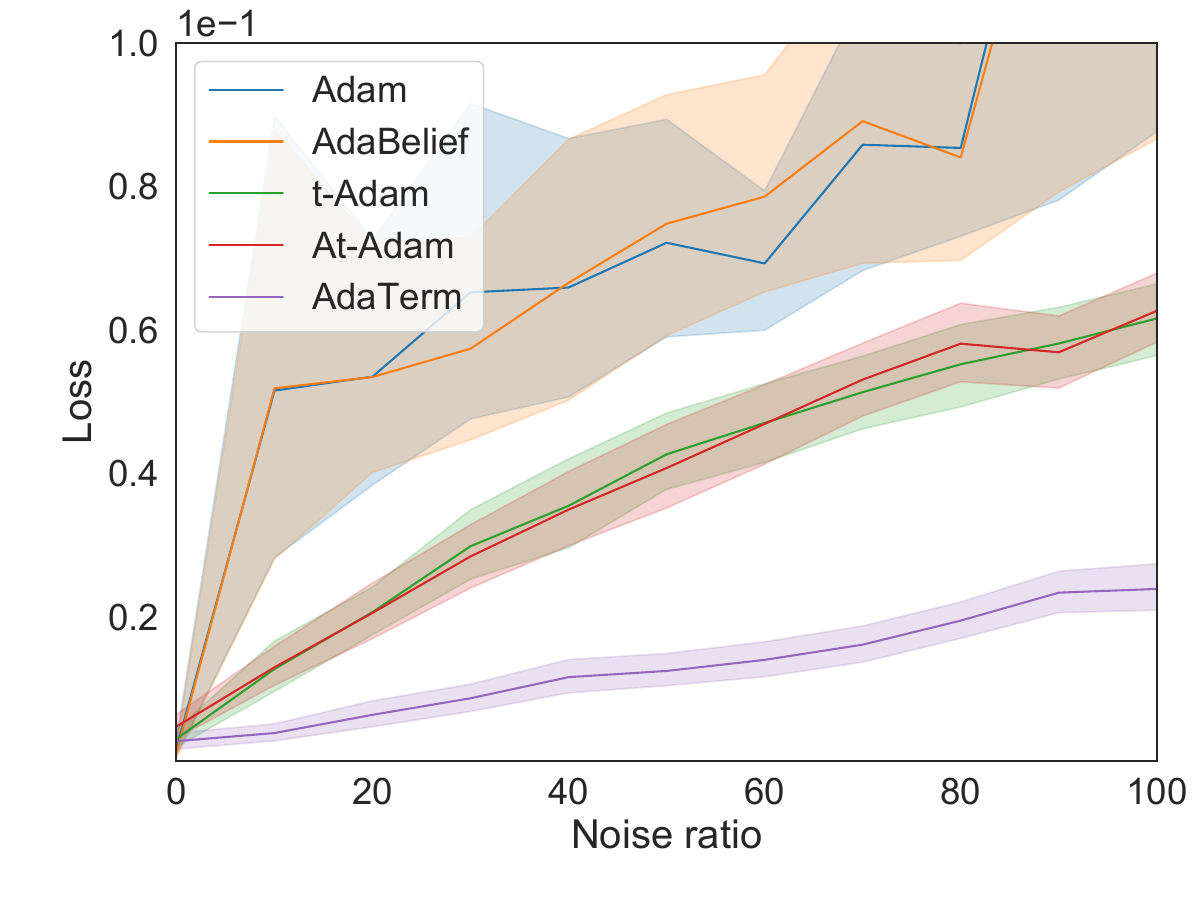}
        \subcaption{Batch size $= 16$}
        \label{fig:loss_nbatch16}
    \end{subfigure}
    \begin{subfigure}[b]{0.23\linewidth}
        \centering
        \includegraphics[keepaspectratio=true,width=\linewidth]{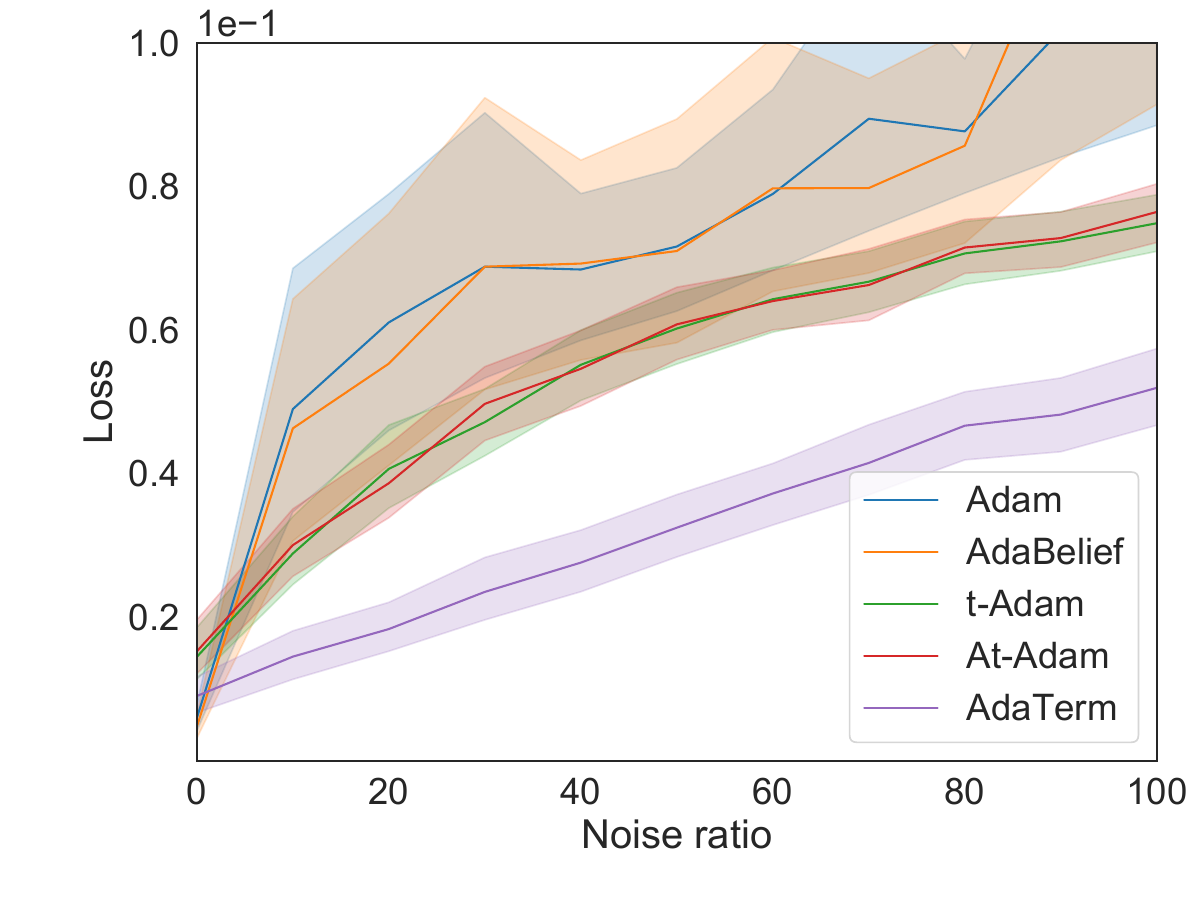}
        \subcaption{Batch size $= 32$}
        \label{fig:loss_nbatch32}
    \end{subfigure}
    \begin{subfigure}[b]{0.23\linewidth}
        \centering
        \includegraphics[keepaspectratio=true,width=\linewidth]{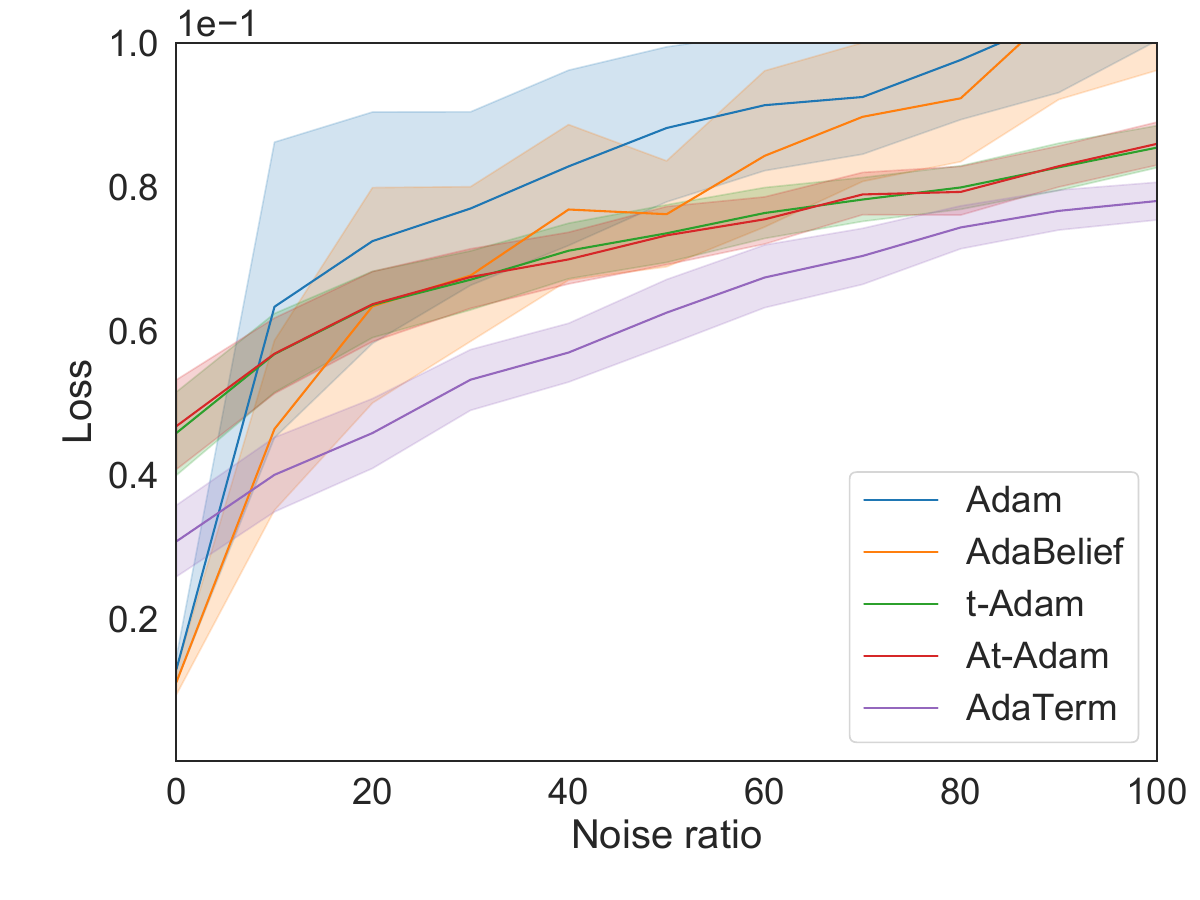}
        \subcaption{Batch size $= 64$}
        \label{fig:loss_nbatch64}
    \end{subfigure}
\end{figure*}

\section{AdaTerm variants}
\label{apdx:variants}

Our main proposed algorithm is given in Alg.~\ref{alg:adaterm}.
However, we consider and study two alternative variants as described below.

\subsection{Updates with uncentered second-order moment}

The derived AdaTerm equations use the estimate of the centered second-order moment, i.e., $\mathbb{E}[(g - m)^2]$, in the update based on the same logic outlined in~\cite{zhuang2020adabelief}. However, the Adam algorithm employs an estimate of the uncentered second-order moment, i.e., $\mathbb{E}[g^2]$. Therefore, we also consider a gradient descent update based on the uncentered estimated moment.
\begin{align}
    \eta^\mathrm{AdaTerm\_Uncentered}(g_t) = \cfrac{m_t (1 - \beta^t)^{-1}}{\sqrt{(v_t + m_t^2) (1 - \beta^t)^{-1}}}
\end{align}
Here, we approximately use the relationship, $\mathbb{E}[(g - m)^2] = \mathbb{E}[g^2] - \mathbb{E}[g]^2$.
Note that, in that case, $\eta^\mathrm{AdaTerm\_Uncentered} < 1$ is always satisfied, which would yield more conservative learning.

\subsection{Adaptive bias correction term}

The bias correction term has its origin in the first proposal of Adam~\cite{kingma2014adam},
where the authors derive the bias term as follows:
\begin{align}
    (1-\beta) \sum\limits_{i=1}^{t} \beta^{t-i} &= 1 - \beta^{t} \label{eq:adam_bias_cor}
\end{align}
and they interpret it as the term needed to correct the bias caused by initializing the EMA with zeros.

 Another interpretation (that does not require considering the expectation of the EMA) is to view this term as the normalization term in a weighted average.
 Indeed, given $\left\{ g_1, \cdots, g_t \right\}$, the weighted sum of $g$ with respect to the weights $\left\{ w_1, \cdots, w_t \right\}$ is expressed as follows:
\begin{align*}
    \sum\limits_{i=1}^{t} w_i g_i
\end{align*}
 If $\sum\limits_{i=1}^{t} w_i \neq 1$, then to obtain the proper weighted average, a correction is required as follows:
\begin{align*}
    \mathrm{ave}(g) &= \left( \sum\limits_{i=1}^{t} w_i \right)^{-1} \sum\limits_{i=1}^{t} w_i g_i
\end{align*}
In the case of the EMA, the weights $w_i$ are given by:
\begin{align}
    m_t &= \sum\limits_{i=1}^{t} (1-\beta) \beta^{t-i} g_i \;\;\Leftrightarrow\;\; w_i = (1-\beta) \beta^{t-i}
    \nonumber \\
    \implies &\sum\limits_{i=1}^{t} w_i = \sum\limits_{i=1}^{t} (1-\beta) \beta^{t-i} = 1 - \beta^{t}
\end{align}
The bias correction term of Adam can therefore be seen as a correction required for a proper weighted average.
We observe that the bias correction term at time step $t$, $c_t$, corresponds to the EMA of $1$:
\begin{align*}
    c_t = \sum\limits_{i=1}^{t} (1-\beta) \beta^{t-i} & = 1 - \beta^{t} = \mathrm{EMA}\left( x_i = 1 \right)
    \\
    \mathrm{such\;that\;}& c_t = \beta c_{t-1} + (1-\beta),\;\;\;\mathrm{with}\;c_{0} = 0
\end{align*}
$c_{0} = 0$ corresponds to the fact that EMA is initialized at $0$.
If instead, $m_{0} \neq 0$, then $c_{0} = 1$.

In AdaTerm, the EMA is expressed as follows:
\begin{align}
    m_t &= (1-\tau_{mv,t}) m_{t-1} + \tau_{mv,t} g_t = \sum\limits_{i=1}^{t} \tau_{mv, i} \prod\limits_{j=1}^{t-i} (1-\tau_{mv, j}) g_i
    \nonumber \\
    &\implies w_i = \tau_{mv, i} \prod\limits_{j=1}^{t-i} (1-\tau_{mv, j})
\end{align}
Since $(1-\beta)$ linearly enters in the expression of $\tau_{mv}$, the bias correction of Adam can also be used in AdaTerm, as expressed in eq.~\ref{eq:adaterm_update_rule}.
However, following the weighted average interpretation, a different correction term can be obtained as follows:
\begin{align}
    c_t &= \sum\limits_{i=1}^{t} \tau_{mv, i} \prod\limits_{j=1}^{t-i} (1-\tau_{mv, j})
    \nonumber \\
    &\implies c_t = \tau_{mv} c_{t-1} + (1-\tau_{mv}),\;\;\;\mathrm{with}\;c_{0} = 0
\end{align}
This correction term is adaptive, and in the Gaussian limit, it will be the same as the Adam/Adabelief term in eq.~\eqref{eq:adam_bias_cor}.
Using this adaptive correction term, a slightly modified AdaTerm is used:
\begin{align}
    \eta^\mathrm{AdaTerm\_AdaBias}(g_t) = \cfrac{m_t c_t^{-1}}{\sqrt{v_t c_t^{-1}}}
\end{align}

\subsection{Comparison}

As can be observed in Table~\ref{tab:comp_results}, the adaptive bias correction is a complete downgrade compared with the vanilla bias correction. However, the uncentered version of AdaTerm outperforms the main algorithm in certain cases and proves to have a certain merit that deserves to be further investigated, as mentioned in the conclusion section above.

\begin{table*}[ht]
    \caption{Results for AdaTerm's variants
    }
    \label{tab:comp_results}
    \centering
    \begin{tabular}{l cc cc cc cc}
        \hline\hline
         & \multicolumn{2}{c}{Classification}
         & \multicolumn{2}{c}{Prediction}
         & \multicolumn{2}{c}{RL}
         & \multicolumn{2}{c}{Distillation}
         \\
         & \multicolumn{2}{c}{Accuracy}
         & \multicolumn{2}{c}{MSE at final prediction}
         & \multicolumn{2}{c}{The sum of rewards}
         & \multicolumn{2}{c}{The sum of rewards}
         \\
        Method & 0~\% & 10~\% & 1 step & 30 steps & Hopper & Ant & w/o amateur & w/ amateur
        \\
        \hline
        AdaTerm
        & \textcolor{orange}{\textbf{0.7315}}
        & \textcolor{orange}{\textbf{0.6815}}
        & \textcolor{orange}{\textbf{0.0335}}
        & 1.0016
        & 1550.25
        & 2021.37
        & \textcolor{orange}{\textbf{1770.17}}
        & \textcolor{orange}{\textbf{1411.02}}
        \\
        (Ours)
        & (3.66e-3)
        & (4.46e-3)
        & (3.09e-4)
        & (2.31e-1)
        & (5.88e+2)
        & (3.87e+2)
        & (2.17e+2)
        & (1.92e+2)
        \\
        \hline\hline
        AdaTerm
        & 0.7309
        & \textcolor{orange}{\textbf{0.6815}}
        & 0.0338
        & \textcolor{orange}{\textbf{0.9085}}
        & \textcolor{orange}{\textbf{1688.59}}
        & \textcolor{orange}{\textbf{2113.81}}
        & 1649.53
        & 1372.08
        \\
        (Uncentered)
        & (2.60e-3)
        & (4.77e-3)
        & (5.23e-4)
        & (1.21e-1)
        & (5.42e+2)
        & (4.01e+2)
        & (2.48e+2)
        & (2.48e+2)
        \\
        \hline
        AdaTerm
        & 0.7294
        & 0.6810
        & 0.0336
        & 1.0342
        & 1595.10
        & 2017.69
        & 1672.71
        & 1400.88
        \\
        (AdaBias)
        & (4.26e-3)
        & (3.13e-3)
        & (4.65e-4)
        & (2.53e-1)
        & (6.45e+2)
        & (3.21e+2)
        & (2.50e+2)
        & (2.45e+2)
        \\
        \hline
        AdaTerm
        & 0.7299
        & 0.6800
        & 0.0336
        & 1.1310
        & 1552.83
        & 1960.78
        & 1624.56
        & 1302.55
        \\
        (Uncentered + AdaBias)
        & (3.70e-3)
        & (3.76e-3)
        & (4.31e-4)
        & (6.43e-1)
        & (6.72e+2)
        & (3.74e+2)
        & (2.21e+2)
        & (2.47e+2)
        \\
        \hline\hline
    \end{tabular}
\end{table*}

\section{Alternative $v$ update rule}
\label{apdx:alt_v}

The gradient ascent update rule for the scale parameter $v$ is expressed as follows:
\begin{align}
v_t &= v_{t-1} + \kappa_v g_v = v_{t-1} + \kappa_v \frac{w_{mv} \tilde{\nu}}{2v_{t-1}^{2} (\tilde{\nu} + 1)} \left[ (s_{t} + \Delta s) - v_{t-1} \right] = v_{t-1} + \kappa_v \frac{w_{mv} \tilde{\nu}}{2v_{t-1}^{2} (\tilde{\nu} + 1)} \left[ (s_{t} + (s_{t} - D_{t}v_{t-1}) \tilde{\nu}^{-1}) - v_{t-1} \right] \\
    &= v_{t-1} + \kappa_v \frac{w_{mv} \tilde{\nu}}{2v_{t-1}^{2} (\tilde{\nu} + 1)} \left[ \left( \frac{\tilde{\nu} + 1}{\tilde{\nu}} \right) s_{t} - \left( \frac{\tilde{\nu} + D_{t}}{\tilde{\nu}} \right) v_{t-1} \right] = v_{t-1} + \frac{\kappa_v}{2v_{t-1}^{2}} \left[ w_{mv} s_{t} - v_{t-1} \right] = v_{t-1} + \frac{\kappa_v}{2v_{t-1}^{2}} \left[ s_{t} + \epsilon_{t} - v_{t-1} \right] \\
    &= v_{t-1} + \tau_v \left[ s_{t} + \epsilon_{t} - v_{t-1} \right]
\end{align}

where $\kappa_v$ is the update step size, and $\epsilon_{t} = (w_{mv} - 1) s_{t}$. To preserve a limiting Gaussian model, we require that for $\tilde{\nu} \to \infty$, $\tau_v \to (1 - \beta)$, since we already have $\epsilon_{t} \to 0$.
Therefore, we have the following:
\begin{align}
\tau_v &= \frac{\kappa_v}{2v_{t-1}^{2}} \implies 0 < \kappa_v < 2v_{t-1}^{2} \implies \kappa_v = 2kv_{t-1}^{2},\; 0 < k < 1 \implies \tau_v = k
\end{align}

Since $s_{t} + \epsilon_{t} = w_{mv}s_{t}$, it is no longer necessary to have $k$ as a function of $w_{mv}$. In particular, we can directly set $k = (1 - \beta)$. However, in practice, this can lead to instabilities and \textit{NaN} values. Therefore, we instead use $k = (1 - \beta)(\overline{w}_{mv})^{-1}$, resulting in the following update rule:
\begin{align}
\kappa_v &= 2v_{t-1}^2 (1 - \beta),\; \tau_v = (1 - \beta)(\overline{w}_{mv})^{-1} \\
v_t &= v_{t-1} + \kappa_v g_v = v_{t-1} + \frac{1 - \beta}{\overline{w}_{mv}} \left[ (s_t + \epsilon_{t}) - v_{t-1} \right] = (1 - \tau_v) v_{t-1} + \tau_v (w_{mv} s_{t}) \nonumber \\
v_t &= \left(1 - \frac{1 - \beta}{\overline{w}_{mv}} \right) v_{t-1} + \frac{1 - \beta}{\overline{w}_{mv}} (w_{mv} s_{t} + \epsilon^{2})
\end{align}


where $\epsilon^{2}$ is added to avoid the zero value problem. Since $w_{mv}$ and $s_{t}$ are both positive, the gradient ascent rule never violates the constraint.

This version of AdaTerm is called \textit{AdaTerm2} in the results below, where we can observe that on the simple regression task, it performs on par with our main algorithm but performs less efficiently on the prediction task. Notably, we can see signs of its instability in the $30$-step prediction task around epoch $100$.
\begin{figure*}[ht]
    \centering
    \begin{subfigure}[b]{0.3\linewidth}
        \centering
        \includegraphics[keepaspectratio=true,width=\linewidth]{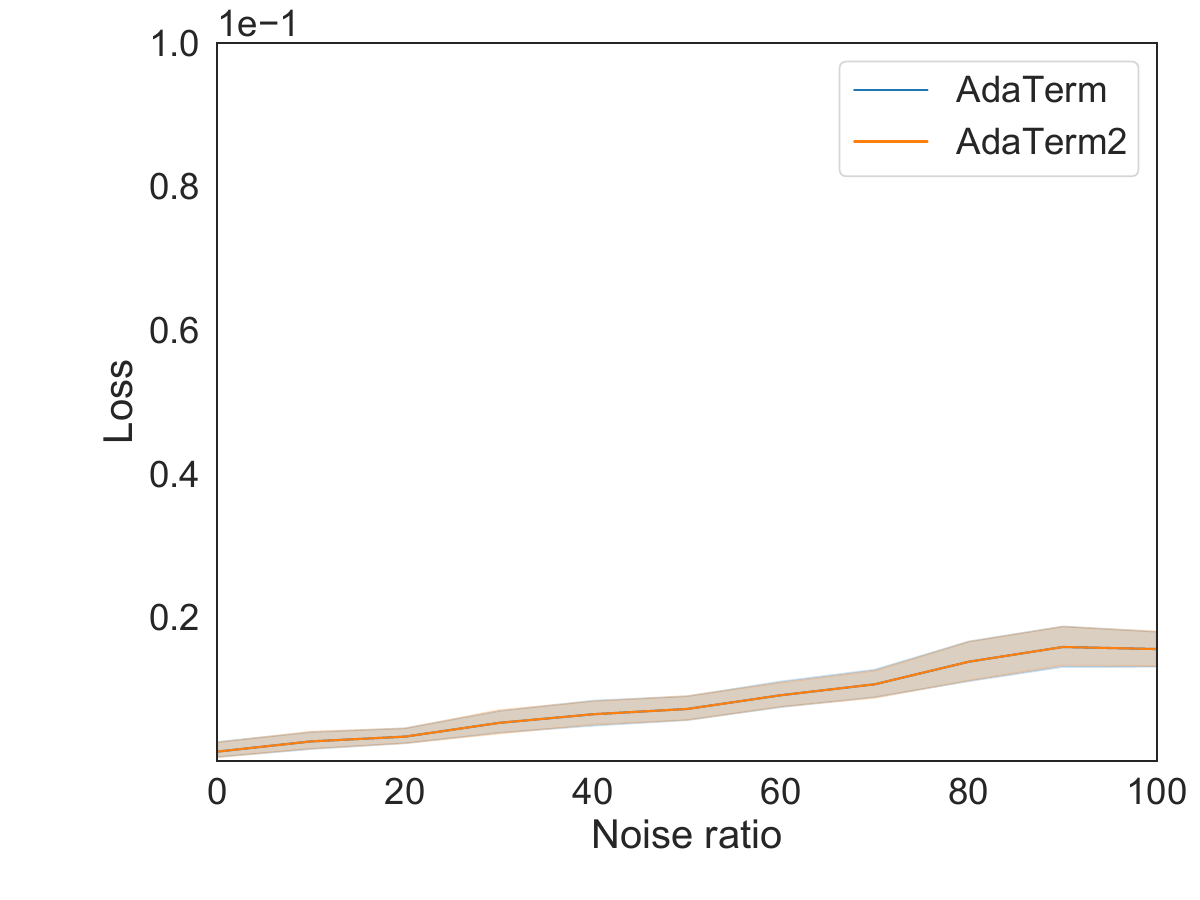}
        \subcaption{Test loss on Regression task}
        \label{fig:loss_adaterm2}
    \end{subfigure}
    \begin{subfigure}[b]{0.3\linewidth}
        \centering
        \includegraphics[keepaspectratio=true,width=\linewidth]{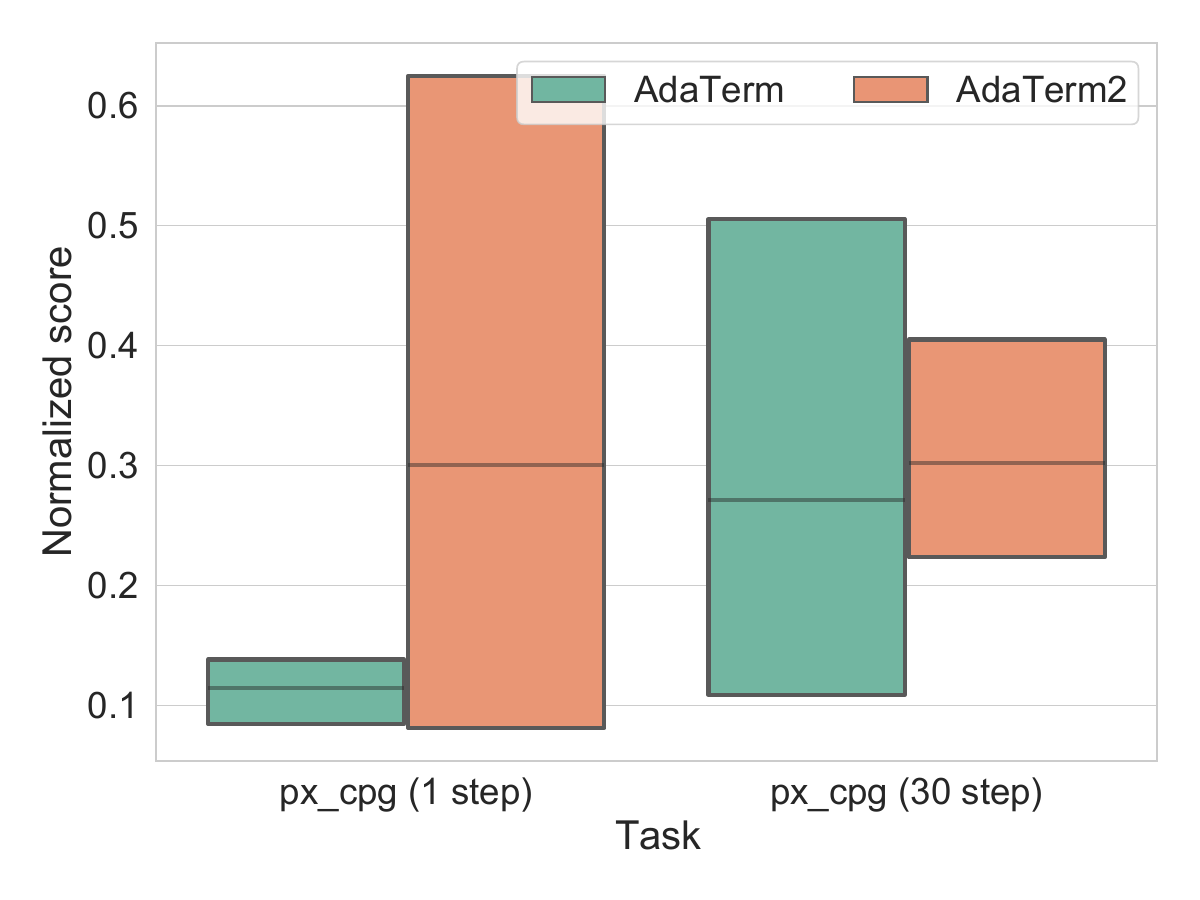}
        \subcaption{Prediction task test error}
        \label{fig:test_prediction_adaterm2}
    \end{subfigure}
\end{figure*}
\begin{figure*}[ht]
    \centering
    \begin{subfigure}[b]{0.23\linewidth}
        \centering
        \includegraphics[keepaspectratio=true,width=\linewidth]{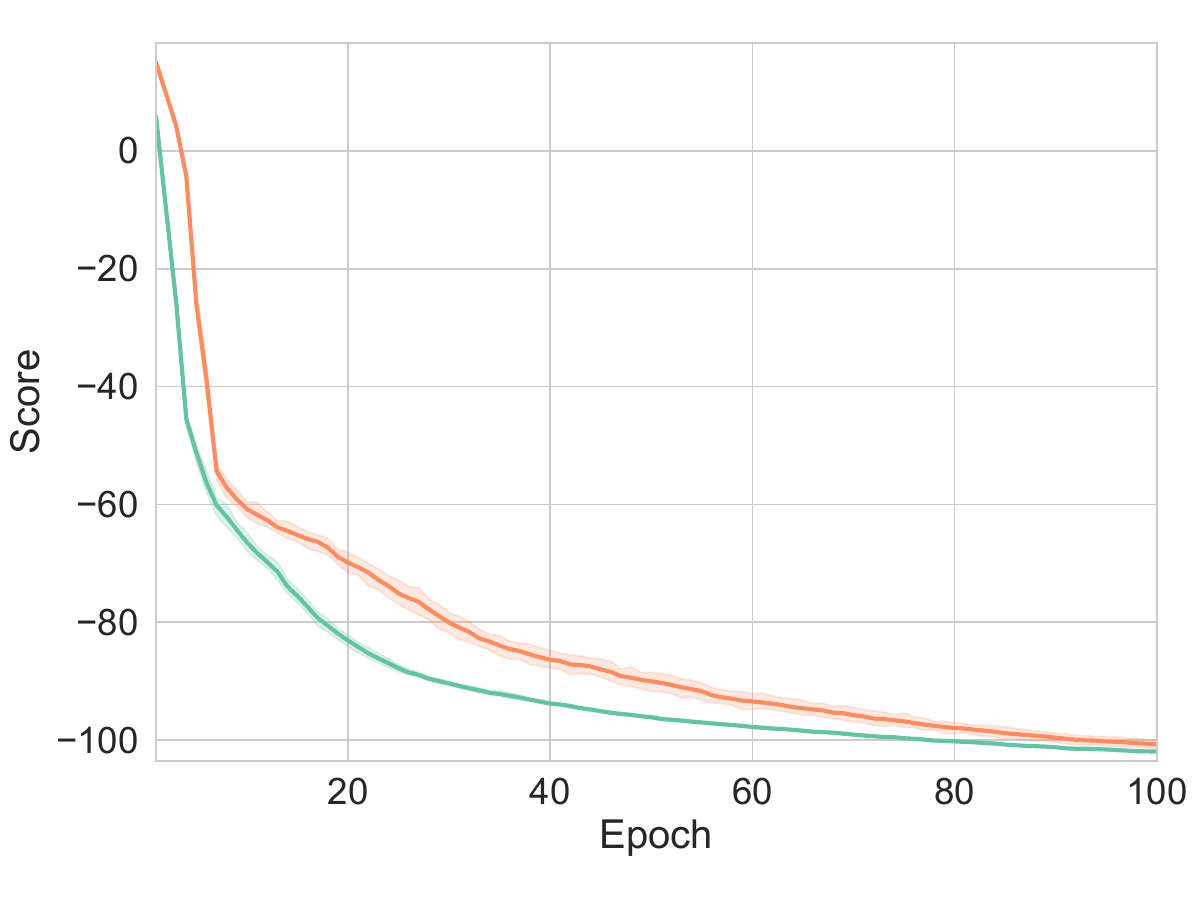}
        \subcaption{Training loss (1 step)}
        \label{fig:train_loss_1step_adaterm2}
    \end{subfigure}
    \begin{subfigure}[b]{0.23\linewidth}
        \centering
        \includegraphics[keepaspectratio=true,width=\linewidth]{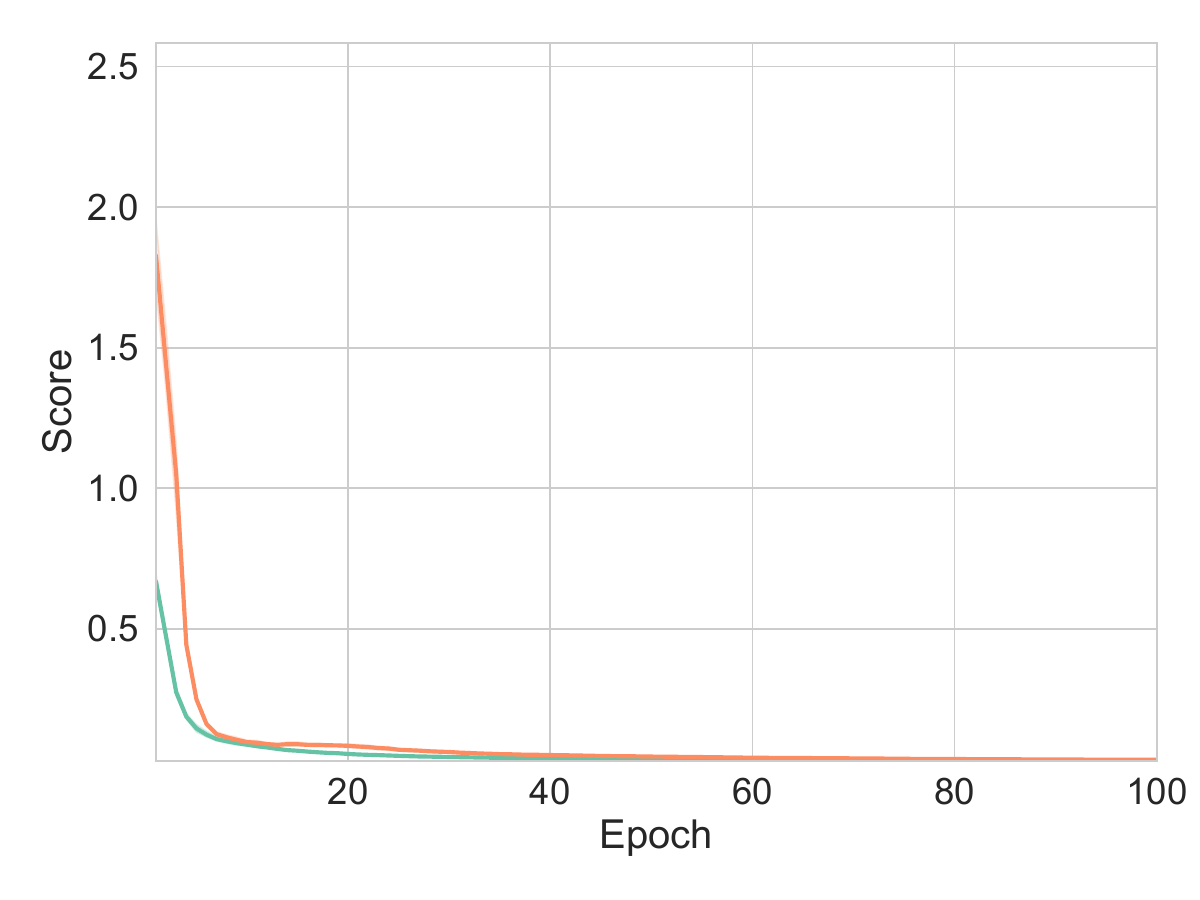}
        \subcaption{Eval. prediction error (1 step)}
        \label{fig:prediction_error_1step_adaterm2}
    \end{subfigure}
    \begin{subfigure}[b]{0.23\linewidth}
        \centering
        \includegraphics[keepaspectratio=true,width=\linewidth]{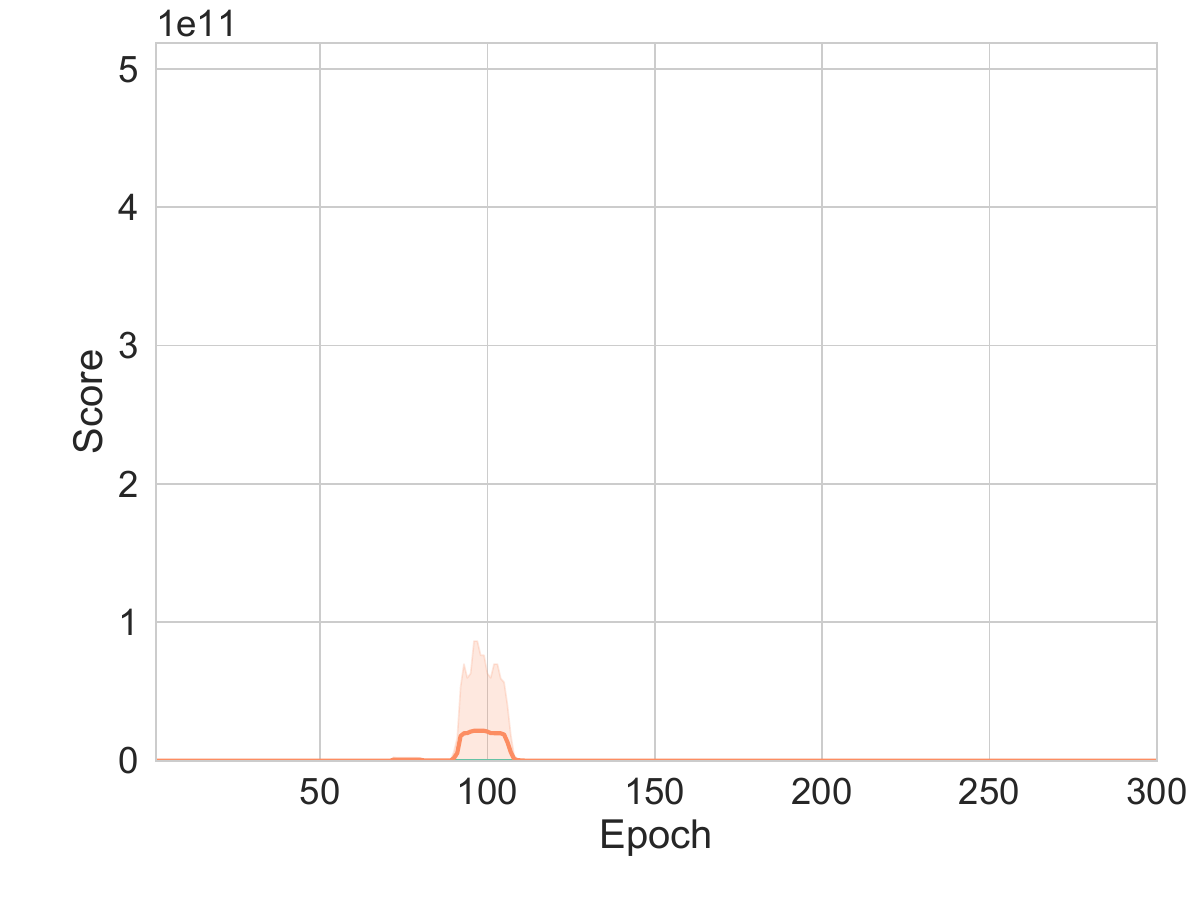}
        \subcaption{Training loss (30 steps)}
        \label{fig:train_loss_30step_adaterm2}
    \end{subfigure}
    \begin{subfigure}[b]{0.23\linewidth}
        \centering
        \includegraphics[keepaspectratio=true,width=\linewidth]{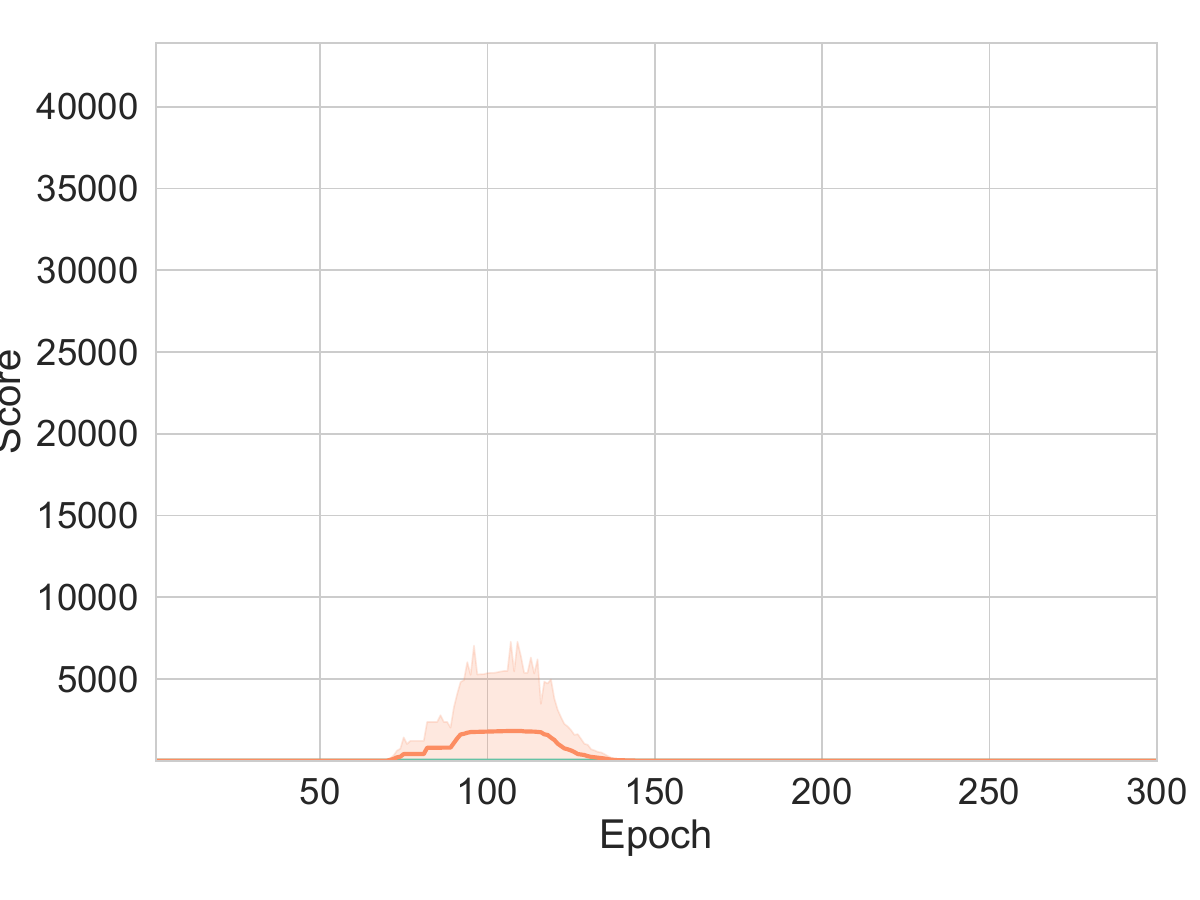}
        \subcaption{Eval. prediction error (30 steps)}
        \label{fig:prediction_error_30step_adaterm2}
    \end{subfigure}
\end{figure*}

\end{document}